\documentclass[preprint,12pt]{elsarticle}

\usepackage{amssymb}
\usepackage{amsmath}

\usepackage[autostyle]{csquotes}
\usepackage{graphicx}
\usepackage{fontawesome5}
\usepackage{tabularx}
\usepackage{caption}
\usepackage{xspace}
\usepackage{amsmath}
\usepackage{mathtools}
\usepackage{enumitem}
\usepackage{booktabs}
\usepackage{longtable}
\usepackage{multirow}
\usepackage{tikz}
\usetikzlibrary{positioning,calc,fit,shapes,arrows.meta,backgrounds,decorations.pathreplacing,calligraphy,backgrounds}
\usepackage{subcaption}
\usepackage{comment}
\usepackage{hyperref}
\usepackage[table]{xcolor}
\usepackage{colortbl}
\usepackage{rotating}

\usepackage{listings}
\lstdefinestyle{prompt}{
    basicstyle=\fontfamily{fvm}\fontsize{8}{10}\selectfont, 
    columns=fullflexible,
    keepspaces=true,
    breaklines=true,
    breakatwhitespace=true,
    postbreak=\mbox{\textcolor{gray}{$\hookrightarrow$}\space},
    showstringspaces=false,
    frame=single, rulecolor=\color{gray!60},
    framesep=4pt, xleftmargin=4pt, xrightmargin=4pt,
    aboveskip=6pt, belowskip=6pt,
    alsoletter={:},
    morekeywords={system:,human:,assistant:},
    keywordstyle=\bfseries,
    extendedchars=true,
    literate={ä}{{\"a}}1 {ö}{{\"o}}1 {ü}{{\"u}}1
             {Ä}{{\"A}}1 {Ö}{{\"O}}1 {Ü}{{\"U}}1 {ß}{{\ss}}1
             {-}{{-}}1
             {–}{{--}}1 {„}{{\quotedblbase}}1 {“}{{\textquotedblleft}}1
             {→}{{$\rightarrow$}}1,
  }

\usepackage{numprint}
\npthousandsep{,}
\npdecimalsign{.}

\newcommand{\TaskEcoNU}[1]{\texttt{Nudging}\xspace}
\newcommand{\TaskEcoNUShort}[1]{\texttt{NU}\xspace}

\newcommand{\TaskEcoSU}[1]{\texttt{Startup}\xspace}
\newcommand{\TaskEcoSUShort}[1]{\texttt{SU}\xspace}

\newcommand{\TaskEcoPS}[1]{\texttt{Pilot strike}\xspace}
\newcommand{\TaskEcoPSShort}[1]{\texttt{PS}\xspace}

\newcommand{\TaskEcoWP}[1]{\texttt{Wind park}\xspace}
\newcommand{\TaskEcoWPShort}[1]{\texttt{WP}\xspace}

\newcommand{\TaskMedAT}[1]{\texttt{Breathing}\xspace}
\newcommand{\TaskMedATShort}[1]{\texttt{BR}\xspace}

\newcommand{\TaskMedAU}[1]{\texttt{Eye}\xspace}
\newcommand{\TaskMedAUShort}[1]{\texttt{EY}\xspace}

\newcommand{\TaskMedKL}[1]{\texttt{Circulation}\xspace}
\newcommand{\TaskMedKLShort}[1]{\texttt{CI}\xspace}

\newcommand{\TaskMedMO}[1]{\texttt{Middle ear}\xspace}
\newcommand{\TaskMedMOShort}[1]{\texttt{ME}\xspace}

\newcommand{\CritFilter}[1]{\texttt{FIL\-TER}\xspace}
\newcommand{\CritCoherence}[1]{\texttt{CO\-HE\-RENCE}\xspace}
\newcommand{\CritArgumentsCount}[1]{\texttt{AR\-GU\-MENTS\_\-COUNT}\xspace}
\newcommand{\CritArgumentation}[1]{\texttt{AR\-GU\-MEN\-TA\-TION}\xspace}
\newcommand{\CritConclusion}[1]{\texttt{CON\-CLU\-SION}\xspace}
\newcommand{\CritReference}[1]{\texttt{RE\-FE\-RENCE}\xspace}
\newcommand{\CritSpecificArguments}[1]{\texttt{SPE\-CI\-FIC\_\-AR\-GU\-MENTS}\xspace}
\newcommand{\CritSumscore}[1]{\texttt{SUM\_SCORE}\xspace}

\newcommand{\CritMed}[1]{\texttt{#1}\xspace}

\newcommand{\ExpNu}[1]{the \texttt{Nudging-only} study\xspace}
\newcommand{\ExpAll}[1]{the \texttt{main} study\xspace}

\begin{document}

\begin{frontmatter}




\title{Rubric-Based AES with LLM Ensembles: Scoring of Student Responses in Critical Online Reasoning Using Ensembles of LLMs}





\author[1]{Daniel Baumartz}
\ead{baumartz@em.uni-frankfurt.de}

\author[1]{Alexander Mehler}
\ead{mehler@em.uni-frankfurt.de}

\author[1]{Patrick Schrottenbacher}
\ead{schrottenbacher@em.uni-frankfurt.de}

\author[1]{Leon Hammerla}
\ead{hammerla@em.uni-frankfurt.de}

\author[2]{Sebastian Gombert}
\ead{s.gombert@dipf.de}


\affiliation[1]{
    organization={Goethe University Frankfurt},
    country={Germany}
}

\affiliation[2]{
    organization={DIPF Leibniz Institute for Research and Information in Education},
    country={Germany}
}


\begin{abstract}

Grading student responses is costly and time-consuming. Moreover, human annotators are prone to bias, fatigue, and rater drift, which further limits grading reliability.
Various methods have been employed to develop automated essay scoring (AES) systems to reduce these costs and address bias-related issues. These methods range from early text-feature-based systems to current machine-learning and transformer-based models.
Specifically, large language models (LLMs) offer the potential to streamline the development of AES systems. However, they are subject to well-known limitations, such as hallucinations and prompt sensitivity. 
This creates a need to explore the potential of a mixed-methods approach that leverages the synergies among individual LLMs through ensemble formation.

We evaluate the potential of ensembles of LLMs to match human-rater agreement when grading student responses in economics and medical education across 103 criteria.
We use 631 German-language responses and corresponding human ratings by 2 to 5 annotators from a previously conducted assessment.
We experiment with 523 rubric-based prompt configurations across 15 open-weight LLMs and assess their consistency with human raters using Krippendorff's alpha for inter-annotator agreement (IAA) and the \textit{Alternative Annotator Test} (alt-test).

Our results show that replacing one human rater with the best-performing LLM--prompt configuration maintains or improves IAA for 91 of the 103 criteria and that LLMs pass the alt-test for 90 criteria.
Our findings suggest that current state-of-the-art LLMs can support AES of German responses in economics and medical education, reducing human workload and enabling faster processing, while validation against human ratings remains necessary.

\end{abstract}




\end{frontmatter}



\section{Introduction}
\label{sec:introduction}

The grading or scoring of human-written texts, such as responses to tasks in higher education, is a central component of educational research, but it is also a lengthy and costly process~\citep{Rudner:Gagne:2001,Livingston:2009,NASEM:2022,Firoozi:et:al:2022}.
Human raters exhibit biases such as halo effects, fatigue, and inconsistency, especially when evaluation rubrics become more fine-grained and large-scale assessment is required~\citep{Eckes:2005,Leckie:et:al:2011,Mahshanian:Shahnazari:2020,Jin:Chiu:2022}.
Consequently, the need for automated essay scoring (AES) systems in education has long been recognised~\citep{Page:1966,Hahn:et:al:2021,Link:Koltovskaia:2023}.
This is supported by the fact that automation and machine learning (ML) have been used in education for a variety of tasks~\citep{Korkmaz:Correia:2019}.
Such methods not only lower costs but also enable applications such as real-time feedback during the writing process~\citep{Cotos:2023,Liu:2024}.
Moreover, the success of large language models (LLMs), driven by their strong performance and ease of use through natural language, has led to extensive research and practical adoption of machine-learning methods~\citep{Ouyang:et:al:2022,Zhao:et:al:2025}.

This raises the question of whether LLM-based approaches can perform scoring tasks: while they may reduce human biases, they introduce biases of their own stemming from the LLM and underlying training data, thereby potentially compromising scoring results~\citep{Amorim:et:al:2018,Weidinger:et:al:2021}.
Research, particularly beyond the field of education, highlights these limitations: 
\citet{Zhou:et:al:2024,Wang:et:al:2025} evaluate LLMs for paper reviewing and participant simulation, reporting problems with long texts and harmful misrepresentation when humans are replaced by LLMs.
In this vein, \citet{Ma:2025} shows that \enquote{assigning a human-like identity to LLMs does not produce human-like behaviors}, while \citet{Piot:et:al:2025} suggest that, although LLMs are no substitute for humans, they can produce similar results in terms of relative rankings.
Nevertheless, LLMs are widely used for annotation and assessment tasks, that is, as \enquote*{LLMs as judges}; see, for example, LLM-Eval by \citet{Lin:Chen:2023}, MT-Bench, Chatbot Arena~\citep{Zheng:et:al:2023}, and the surveys by \citet{Gu:et:al:2026,Li:et:al:2024}, which provide overviews of relevant fields and applications.
Research more often highlights the opportunities and strong performance of LLMs.
They span a wide range of domains, both in education~\citep{Jin:et:al:2025,Deroy:et:al:2025} and beyond, for example in medicine and healthcare~\citep{Bafna:et:al:2025,Luo:et:al:2025,Li:et:al:2025}, law~\citep{Carnat:et:al:2024}, and finance~\citep{Schopf:et:al:2025}.

Consequently, we face a dilemma: although LLMs continue to evolve and are increasingly studied for their potential to complement or replace human linguistic abilities, their limitations are also becoming clearer, calling their broad applicability into question.
Despite their widespread use and considerable potential, further research is needed to disentangle their strengths and limitations, especially in the assessment of human-produced work. 
This is exactly the task of the present article, with a focus on AES.

More specifically, we examine LLMs as means for rubric-based scoring of responses in the context of critical online reasoning (COR).
In COR research, students are assessed on their ability to gather online information to complete a task by writing a response to a question, with COR examined across three phases or facets:
(1) how they select sources and by which methods (OIA, online information acquisition), (2) how well they assess source reliability (CIE, critical information evaluation), and (3) their reasoning skills, as reflected in their response (REAS, reasoning with evidence, argumentation, and synthesis)~\citep{Molerov:et:al:2020}.
With a focus on the third facet, REAS, we study the potential of using LLMs for the automated scoring of responses using previously collected responses and ratings from assessments in German economics and medical education.
We provide empirical evidence on whether LLMs can replace individual human annotators while matching their level of agreement, i.e., without lowering the agreement of the rater group.
Thus, our construct is COR performance, while we examine student responses as items and the features attributed to them by LLMs as observable indicators.

In this study, we address the following research questions:
\begin{itemize}
    \item[RQ1] To what extent can LLM-based AES systems replace individual human raters in grading student responses while maintaining the level of agreement among the raters?
    \item[RQ2] How robust are LLM-based AES systems across eight tasks spanning the domains of economics and medicine?
    \item[RQ3] Which prompt- and model-based features have the greatest influence on the performance of LLM-based AES systems?
\end{itemize}

Since our investigation focuses on LLMs as a potential replacement for human annotators of responses to COR tasks in the context of AES, \autoref{sec:related_work} is divided into four parts covering background on AES, related work on LLM-based annotation, LLM-based AES, and rater evaluation.

\section{Related Work}\label{sec:related_work}
\subsection{Background on AES}\label{sec:AES}

Automated essay scoring (AES) is the task of automatically grading written texts according to one or more assessment criteria.
See \citet{Shermis:Burstein:2013,Lim:et:al:2021} for overviews of AES systems and methods.
A closely related task is automatic short-answer grading (ASAG); see \citet{Burrows:et:al:2015,Haller:et:al:2022} for overviews.
Based on the criteria for short answers~\citep{Burrows:et:al:2015}, the statements we analyse in this study share characteristics of short answers, as they require external knowledge, are written in natural language, and are assessed mostly on content, but they are less restricted in length and openness.
We therefore consider relevant research from both areas.
Starting with early AES systems such as Project Essay Grade (PEG)~\citep{Page:1968}, researchers first studied feature-engineering-based approaches~\citep{Burstein:et:al:1998,Rudner:Liang:2002,Yannakoudakis:et:al:2011,Chen:et:al:2013,Cozma:et:al:2018} before increasingly relying on neural networks that perform feature engineering as part of the training process~\citep{Taghipour:et:al:2016,Alikaniotis:et:al:2016,Dong:et:al:2017,Wang:et:al:2018,Xie:et:al:2022}.
Rather than relying solely on the latest developments in LLMs, several approaches use a wider range of machine-learning methods to perform AES.
\citet{Wilson:et:al:2024} recently studied the automatic scoring of students' scientific argumentation skills without the use of LLMs, relying on an ensemble of machine-learning algorithms.
While they do not compare their results with LLM-based approaches, their approach achieves strong performance while avoiding LLM-specific limitations such as hallucinations.
However, it still exhibits bias, for example, by assigning lower ratings to English learners, and requires more effort and expert knowledge to prepare.

With the use of LLMs, the preparation and development of such tools can potentially be shifted to prompt engineering, while natural language as an interface makes these systems more accessible to non-experts.
The ability to instruct LLMs in natural language and obtain explanations is also identified as an advantage by \citet{Gao:et:al:2025}.
They caution against focusing on English, especially in light of the limited amount of training data available for low-resource languages in LLMs.
\citet{Emirtekin:2025} presents a review of LLM use for automated assessment, giving an overview of domains, datasets and languages.
At the same time, newer reasoning models can reduce the need for specialised fine-tuning~\citep{Brown:et:al:2020,Wei:et:al:2022,Kojima:et:al:2022,Guo:et:al:2025}.
Current advances in multimodal~\citep{Su:et:al:2025}, agentic~\citep{Li:2025}, and workflow-based AI~\citep{Xiao:et:al:2024,Zhang:et:al:2025} make LLMs increasingly capable of scoring not only text data, but also image data (see vision language models, and, more generally, multimodal LLMs \citep{Caffagni:et:al:2024,Yin:et:al:2024}) and process data such as visited websites and keystrokes~\citep{He:et:al:2025}.
Although these developments are outside the scope of the present study, they indicate how LLM-based assessment may develop beyond written statements alone.

In light of these developments, the prospect of AES increasingly based on LLMs arises. 
However, fundamental problems such as hallucinations remain unresolved, especially for LLMs. 
This raises the question of whether combining LLMs with prompting strategies can produce scoring results suitable for supporting or even replacing human scorers. 
In other words, can LLMs serve as scorers or annotators alongside humans?
This question is the central focus of this article.

\subsection{LLM-based Annotation}
\label{sec:related_work_llm_annotation}

The potential of using LLMs such as ChatGPT for annotation tasks has been demonstrated across different domains and tasks, although performance relative to human annotators varies:
\citet{Gilardi:et:al:2023} show that ChatGPT-3.5 outperforms crowdworkers, but not trained annotators, whereas \citet{Tornberg:2023} reports ChatGPT-4 outperforming both.
However, \citet{Tavakoli:Zamani:2025} identify calibration issues and prompt sensitivity as persistent problems even in recent models such as ChatGPT-4o and Claude~3.
They therefore propose a human-in-the-loop workflow to improve the reliability of LLM-based annotation.
Similarly, \citet{Wang:et:al:2024} use human-rated gold data to verify LLM-generated annotations and report improved accuracy after re-annotating failed annotations by human raters.

\citet{Borse:et:al:2025} investigate the replacement of a human rater.
They compare ChatGPT-4o and ChatGPT-4.5 with the combined ratings of two human raters and report reliability in terms of Cohen's kappa.
They conclude that LLMs can replace human raters in the qualitative analysis of transcripts, but \enquote{have limitations in coding themes that are not domain-specific}.
Other studies caution against replacing humans with LLMs, including \citet{Clarke:Dietz:2025} and \citet{Shen:et:al:2023}.
The latter conclude that, while models like ChatGPT and Llama can outperform other automated systems, they cannot fully replace humans, and report problems with LLMs' ability to distinguish between results with similar performance, model bias, and decreasing correlation with human judgments for higher-quality summaries.

Model bias in automated annotation is not unique to LLM-based methods.
\citet{Amorim:et:al:2018} show that feature-based machine-learning models trained on data from which human-rater bias has been removed perform better.
For LLM-based annotation, \citet{Weidmann:et:al:2025} find that Llama~3.1 and ChatGPT-4o are able to perform annotation and approximate human judgments, but that the models' bias is clearly noticeable and affects the annotation.
They show that using an ensemble method is not sufficient on its own, but that combining two LLMs with opposing biases improves the results.
The responsible use of LLMs, and AI more generally, is studied by \citet{Johnson:Zhang:2024}, who observe that the average score assigned by the LLM is lower than that assigned by the human annotators.
The authors therefore conclude that evaluation should include fairness and explainability.
Similarly, \citet{Chen:et:al:2024} show that both humans and LLMs produce significant biases, although these biases differ and vary in degree.

Taken together, these findings strongly suggest the need for human-based validation of LLM-based annotation, even though human judgement is itself not free from biases and should be taken into account when evaluating LLM-based approaches.
The findings motivate the approach in our study to compare multiple LLMs and prompts with individual human ratings.
We discuss the potential and use of LLM-based tools and methods for AES specifically in the next section.

\subsection{LLM-based AES}
\label{sec:related_work_llm_aes}

Research on the assessment of written responses in education in both AES and ASAG also shows ambivalent results.
In 2024, \citet{Chamieh:et:al:2024} analyse the use of LLMs to automatically score short-answer responses.
They use existing datasets such as ASAP in their evaluation and test GPT-3.5, GPT-4, and Llama.
The authors report low performance, which they were able to improve by fine-tuning the LLMs.
They conclude that there \enquote{currently seems to be no basis for applying LLMs for short-answer scoring}, 
especially when tasks require domain knowledge or complex reasoning.
Using Llama~2 and ChatGPT-3.5, \citet{Mansour:et:al:2024} evaluate LLM performance in automated essay scoring on the ASAP dataset and compare it with conventional non-LLM systems.
They show the importance of prompt engineering for each task and LLM separately and observe differences in prompt performance between ChatGPT and Llama.
Additionally, Llama's performance fluctuates much more than ChatGPT's, further highlighting the differences between more open and more closed LLMs.
Compared with non-LLM systems, they report much lower performance for both LLMs, highlighting the limitations of models from just a few years ago.
Conversely, in a different assessment setting, \citet{Latif:Zhai:2024} report an average performance increase of 9.1\% when comparing a fine-tuned GPT-3.5 model with a BERT-based approach in an assessment task in science education.

Since these studies, multiple LLM generations have significantly improved models' reasoning abilities.
Reasoning, as manifested, for example, in chain-of-thought (CoT) prompting \citep{Wei:Wang:2022,Kojima:et:al:2022}, concerns the models' ability to break a task down into subtasks and elaborate on them in order to arrive at an overall solution, thereby giving the models greater control over response generation.
Regardless of whether reasoning is explicitly elicited by prompting, newer models are trained to generate reasoning traces or argumentation chains to produce better answers; the approaches further discussed in this section focus on such models.

\citet{Yancey:et:al:2023} compare ChatGPT-3.5 and ChatGPT-4 with human annotators using essays written by English learners and report that ChatGPT-4 performs on a par with conventional models.
Very recently, \citet{Cong:et:al:2026} also report performance matching or exceeding that of BERT-based models.

\citet{Lee:et:al:2024} compare GPT-3.5 and GPT-4 in scoring science assessments completed by U.S. middle school students and graded by human raters.
They observe that there was no single best-performing prompt across all tasks and that LLMs are capable of generating explainable and interpretable ratings, while the use of CoT prompting and few-shot examples improves accuracy.
Similarly, using student-writing assessment data from young students in the U.S., \citet{Huang:et:al:2025} evaluate GPT-4o, Gemini~2.5 Flash, and Claude Sonnet~4 and compare them with ratings produced by trained professionals.
Their findings confirm that structured prompts and few-shot examples improve Quadratic Weighted Kappa (QWK) alignment~ \citep{Doewes:et:al:2023,DeRaadt:et:al:2021} and can achieve performance comparable to feature-based systems, which they report remain \enquote{the most reliable}.

The effects of rubric design are demonstrated by \citet{Frohn:et:al:2025}, who compare human ratings with ratings generated by 11~LLMs, including Claude, Gemini, ChatGPT, and Llama variants, on short-answer questions in mathematics and English.
They score answers as correct or incorrect and evaluate three prompts that provide (1) only the question, (2) the question plus rating criteria, and (3) examples.
Although the authors investigate ASAG, their comparison of multiple LLMs and different rubric information is directly relevant to our multi-model rubric-based experiment setting.
They observe that the group of LLMs showed higher agreement reliability among themselves than the human raters.
Results are reported using Fleiss' kappa and show that LLM performance depends on rubric design: \textit{the more information provided in the prompt, the higher the reliability}.

This phenomenon is also investigated in the AES field by \citet{Yoshida:2025}, who use the TOEFL\-11 dataset, Claude~3.5 Haiku, ChatGPT-4o-mini, Gemini~1.5 Flash, and Llama~3, along with three rubric variants.
Except for Gemini, they report higher QWK scores with more informative prompts; however, the differences are small and not significant.
They conclude that Gemini's lower performance may result from the longer prompt.
While prompt differences were small, absolute scores reached only around 0.6, leading them to conclude that AES using LLMs \enquote{is not yet ready for practical use}.

\citet{Kortemeyer:2023} further compares the general-purpose LLM ChatGPT-4 with specialised or fine-tuned models for automated short-answer grading.
They show that ChatGPT does not match the performance of specialised models; however, they note the advantage of not having to train the LLM.
Additionally, they test two variants, one of which provides the LLM with a reference answer and yields comparable performance.

\citet{Shibata:Miyamura:2025} study automated essay scoring by asking LLMs to rank essay pairs rather than assign scores directly.
The final score is then computed by training RankNet~\citep{Burges:et:al:2005} on multiple comparison pairs and linearly mapping its output to the rubric value range.
The method is evaluated on two datasets using five LLMs: ChatGPT-4o and 4o-mini, Llama~3.1 and~3.2, and Mistral. It is found to outperform other prompting-based methods in most cases.

Working with student data requires great care with regard to data protection and privacy.
\citet{Stasuik:2025} compares locally hosted LLMs (Llama~3.1 and~3.2) with closed-weight commercial models (ChatGPT-4o-mini, o1, and o3-mini) for scoring essays from English courses.
Using mean absolute grade difference, \citeauthor{Stasuik:2025} shows that none of the tested models can replace human assessment and that LLMs often assign lower scores than human raters.
In that study, newer LLMs with larger context windows performed better.
Locally hosted models performed much worse; however, the newer Llama version showed substantial improvement, suggesting that future models may become more capable.

\citet{Emirtekin:Ozarslan:2026} study agreement between human annotators and four closed- and open-weight LLMs (ChatGPT-4o, Gemini~2 Flash, DeepSeek~V3, and Llama~3.3) using university students' short-answer texts.
They report moderate agreement in terms of the QWK metric, with much higher agreement among the LLMs themselves.
Further, they show that LLMs can efficiently grade lower-order cognitive skills, with criteria targeting factual and definitional aspects showing higher agreement than those requiring inference or contextual awareness.
However, \citeauthor{Emirtekin:Ozarslan:2026} conclude that fully automated assessment is limited when higher-order skills are considered and that LLMs cannot replace human raters in this case.
In this way, \citet{Tate:et:al:2024} report positive results for holistic AES using ChatGPT-3.5 and~4 and find no statistically significant difference between humans and ChatGPT.

In medical education, \citet{Quah:et:al:2024} highlight the importance of rubric design and note that ChatGPT was unable to identify incorrect statements or irrelevant content.
They compare human and ChatGPT-4 performance using IAA and correlation measures.
Their results vary by assessed question but still show potential for LLM-based automated scoring.
Similarly, \citet{Deng:et:al:2025} show that alignment with human ratings decreases as criteria become more complex, from binary to more fine-grained criteria.

This research often demonstrates a limited assessment potential of LLMs. 
That potential depends on various factors, including model maturity and size. 
However, the use of large commercial models raises significant data-protection concerns. 
Overall, LLMs show greater potential for assessing lower-order than higher-order cognitive skills. 
Based on the current state of research, LLM-based AES systems cannot be recommended without reservation for the latter.

This interim finding raises several questions that we address in this article:
since LLM-based scoring cannot currently replace human scoring and remains limited in the range of cognitive skills it can assess, the question arises whether LLMs are better suited to supplement human raters. 
This would require adapting the models to specific scoring criteria, either through fine-tuning or specialised prompting techniques, provided that these criteria do not merely target low-level skills. 
However, for data-protection reasons, only non-commercial models that are available locally and do not transmit any data to external providers can be considered for this purpose.
Consequently, the next section reviews related work on rater evaluation: given a low IAA among raters, supplementing human raters becomes a challenging task.

\subsection{Rater Evaluation}
\label{sec:related_work_evaluation}

A general overview of rater evaluation through the measurement of agreement among annotators is presented by \citet{Artstein:Poesio:2008}, covering, e.g., Krippendorff's alpha or Cohen's kappa.
A more recent line of research examines how annotator agreement and reliability can be measured and interpreted, especially when using inexperienced raters or crowdworkers.
In such cases, reliance on absolute inter-rater metrics or thresholds, such as Krippendorff's alpha, is considered problematic by \citet{Wong:et:al:2021}.
They present a novel measure based on Cohen's kappa, named cross-replication reliability (xRR).
In the same way, \citet{Andersen:et:al:2021} compare a set of measures in the context of evaluating AES systems and propose an evaluation method that considers acceptable ranges when multiple annotations are available.
These findings show that measures and thresholds should be contextualised, e.g., through replication, rater experience, and the purpose of comparison.
In our study, we use rater agreement on human- and LLM- assigned ratings.
For this reason, we rely on the \textit{Alternative Annotator Test}, which is presented below, as a complementary measure.

When no human-generated gold data are available, for example because of cost, another method is to use an LLM to judge the outputs of multiple LLMs and generate a final result~\citep{Guerdan:et:al:2025}.
These judges, however, still have to be evaluated against human annotations to determine their performance.
\citet{Guerdan:et:al:2025} study such judge systems under the assumption that there is not always a single correct gold annotation, or that one annotator may regard multiple ratings as correct.
They find that such systems differ both from one another and from the human annotators.

\citet{Wang:et:al:2025b} present an evaluation framework based on Generalizability Theory (G-theory) and Many-Faceted Rasch Measurement (MFRM).
Notably, the authors report results comparing Claude (3.5 Haiku and Sonnet) with the widely used ChatGPT models (4o and 4o-mini).
Claude outperforms the ChatGPT models and is reported to perform on a par with fine-tuned ChatGPT variants, further highlighting the rapid pace of model development.
Using the ASAP datasets, the authors show that LLMs perform well in relative ranking, but are limited in absolute scoring.
Compared with humans, who perform strongly on holistic scoring, LLMs significantly outperform them on analytic scoring; however, both are affected by biases such as halo effects.

The \textit{Alternative Annotator Test} (alt-test), which we use for comparison in our evaluation, is presented by \citet{Calderon:et:al:2025}.
It is a statistical approach for evaluating whether human annotators can be replaced by LLMs and whether an LLM \enquote{offers a better alternative to recruiting human annotators}.
\citet{Calderon:et:al:2025} position the alt-test as a more rigorous metric for comparing an LLM with a group of human annotators than measures such as IAA, which assess the group as a whole.
Like IAA, the alt-test assesses alignment with the human group rather than correctness.
\citet{Calderon:et:al:2025} apply the alt-test to six LLMs (ChatGPT-4o and 4o-mini, Gemini~1.5 Flash and Pro, Llama~3.1, and Mistral) and human annotators with different skill levels.
Based on the test, they show that at least one LLM can replace human annotators in nine of the ten datasets, and that closed-weight models outperform locally hosted ones.
This insight is particularly relevant in our context, as it highlights the challenges associated with our focus on local models, which are largely indispensable for data-protection reasons. 
In other words, achieving satisfactory alt-test results with local open-weight models would demonstrate the effectiveness of our approach in this setting.
Because the alt-test is a recent method with limited independent validation, we use it as a complementary evaluation measure in combination with IAA for evaluating replaceability.

\subsection{Summary}\label{sec:Summary}

Overall, the research shows the potential of LLMs for AES and highlights the need for extensive model testing and prompt engineering.
It also cautions against overreliance on LLM capabilities, suggesting that any fully automated approach must be validated by human raters in their intended assessment setting for each combination of model, task, rubric or criterion, and prompt.

In this article, we study how a range of open, locally hosted LLMs from different families rate student answers and examine, among other things, whether they can supplement or even replace human annotators using inter-annotator agreements and the alt-test.
Compared with other studies, we evaluate previously collected, real-world responses in German, a less-studied language, and economics, a less-studied domain.
That is, we explicitly investigate the possibility of replacing human raters with local models in this specific application area. 
We show that, under these conditions, individual human raters can indeed be replaced without lowering the agreement among the raters, even though previous research has generally reported negative results for local models. 
This provides valuable insight into the potential of rapidly advancing research on automating human capabilities in areas heavily affected by data-privacy concerns.

\section{Methods}
\label{sec:methods}

\begin{figure}
    \centering
    \resizebox{\textwidth}{!}{%
\begin{tikzpicture}[
    font=\sffamily\small,
    >=Latex,
    node distance=0.95cm and 0.9cm,
    box/.style={
        draw,
        rounded corners=2pt,
        align=center,
        minimum width=5.6cm,
        minimum height=0.95cm,
        text width=5.25cm,
        fill=white
    },
    header/.style={
        draw,
        rounded corners=2pt,
        align=center,
        minimum width=5.6cm,
        minimum height=0.65cm,
        text width=5.25cm,
        fill=gray!10,
        font=\sffamily\small\bfseries
    },
    compactbox/.style={
        draw,
        rounded corners=2pt,
        align=center,
        minimum width=4.65cm,
        minimum height=0.95cm,
        text width=4.35cm,
        fill=white
    },
    group/.style={
        draw,
        dashed,
        rounded corners=4pt,
        inner sep=7pt,
        fill=gray!5
    },
    arrow/.style={
        ->,
        thick
    },
    note/.style={
        font=\sffamily\scriptsize,
        align=center,
        text=gray!70
    }
]


\node[box] (responses) {
    \textbf{Student Responses}\\
    \scriptsize Online assessment\\
    \scriptsize 8 tasks across 2 domains
};

\node[box, below=0.55cm of responses] (humans) {
    \textbf{Human Ratings}\\
    \scriptsize 2--5 human raters per response\\
    \scriptsize with rubric of 103 criteria including 8 derived holistic scores
};

\begin{scope}[on background layer]
    \node[group, fit=(responses)(humans)] (prior) {};
\end{scope}


\node[box, below=1.2cm of humans] (llm) {
    \textbf{LLM-based AES}\\
    \scriptsize Student responses graded using different\\
    \scriptsize prompt and model configurations
};


\node[box, below left=1.15cm and 0.5cm of llm] (iaa) {
    \textbf{IAA Change}\\
    \scriptsize Stable IAA when replacing one\\
    \scriptsize  human rater with an LLM
};

\node[box, below right=1.15cm and 0.5cm of llm] (alt) {
    \textbf{Alternative Annotator Test}\\
    \scriptsize Can the LLM replace\\
    \scriptsize $>= 50\%$ of the raters
};

\node[box, below=3.6cm of llm] (outcomes) {
    \textbf{Criterion is Gradable}\\
    \scriptsize Non-negative IAA change\\
    \scriptsize or passing alt-test
};

\begin{scope}[on background layer]
    \node[group, fit=(iaa)(alt)(outcomes)] (grading) {};
\end{scope}


\node[box, below=1.15cm of outcomes] (features) {
    \textbf{Feature Extraction}\\
    \scriptsize Model, prompt, token,\\
    \scriptsize task, and domain features
};



\node[box, below=1.2cm of features] (proxy) {
    \textbf{Train Proxy Models}\\
    \scriptsize Predict LLM performance\\
    \scriptsize from extracted features
};

\node[box, below left=1.15cm and 0.65cm of proxy] (regression) {
    \textbf{Regression}\\
    \scriptsize Predict exact IAA
};

\node[box, below right=1.15cm and 0.65cm of proxy] (classification) {
    \textbf{Classification}\\
    \scriptsize Predict non-negative change
};

\begin{scope}[on background layer]
    \node[group, fit=(proxy)(regression)(classification)] (model) {};
\end{scope}

\node[box, below=3.6cm of proxy] (shap) {
    \textbf{SHAP Analysis}\\
    \scriptsize Explain feature importance
};


\coordinate (evalsplit) at ($(llm.south)+(0,-0.45)$);
\coordinate (evalmerge) at ($(outcomes.north)+(0,0.45)$);

\coordinate (modelsplit) at ($(proxy.south)+(0,-0.45)$);
\coordinate (modelmerge) at ($(shap.north)+(0,0.45)$);


\draw[arrow] (responses.south) -- (humans.north);
\draw[arrow] (humans.south) -- (llm.north);

\draw[arrow] (llm.south) -- (evalsplit);
\draw[arrow] (evalsplit) -| (iaa.north);
\draw[arrow] (evalsplit) -| (alt.north);

\draw[arrow] (iaa.south) |- (evalmerge) -- (outcomes.north);
\draw[arrow] (alt.south) |- (evalmerge) -- (outcomes.north);

\draw[arrow] (outcomes.south) -- (features.north);
\draw[arrow] (features.south) -- (proxy.north);

\draw[arrow] (proxy.south) -- (modelsplit);
\draw[arrow] (modelsplit) -| (regression.north);
\draw[arrow] (modelsplit) -| (classification.north);

\draw[arrow] (regression.south) |- (modelmerge) -- (shap.north);
\draw[arrow] (classification.south) |- (modelmerge) -- (shap.north);

\end{tikzpicture}
    }
    \caption{Flowchart of the study design and evaluation process.}
    \label{fig:flowchart}
\end{figure}
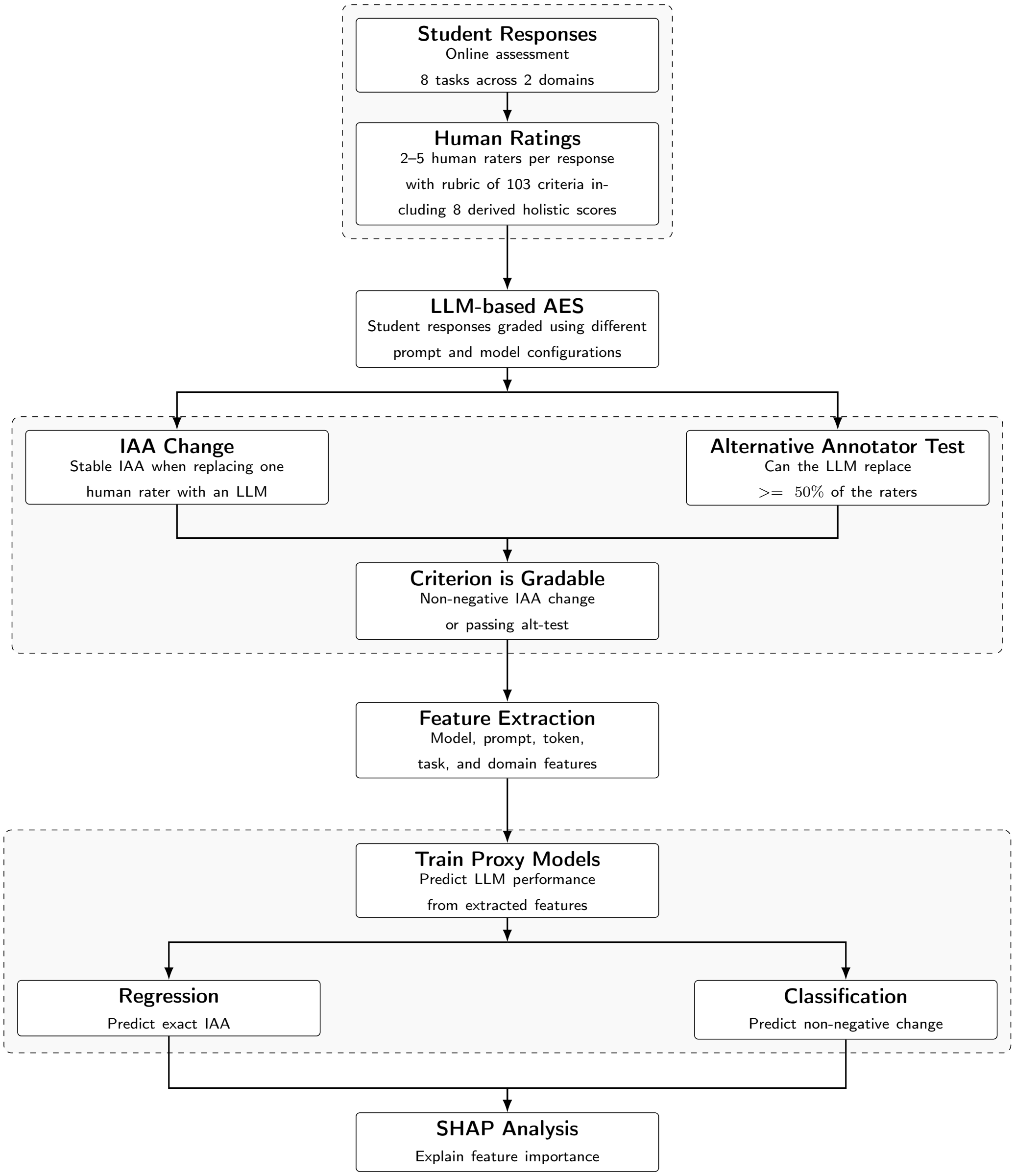

We analyse the potential to replace human annotators by prompting LLMs to automatically rate students' statements against a set of predefined criteria.
Our approach combines a series of local models with various criterion-specific prompting strategies. 
This allows us to achieve the contextual sensitivity essential to our approach. 
We first present the application domain and the data analysed in it (\autoref{sec:methods_data}). 
We then describe the LLMs we evaluate and the prompting strategies we partly developed and tested (\autoref{sec:methods_llms}). 
\autoref{fig:flowchart} summarises our approach graphically.

\subsection{Data and Rubrics}
\label{sec:methods_data}

We consider real-world statements as written responses to questions posed to economics and social-science students (\enquote{economics} group), as well as medicine and physics (\enquote{medicine} group) students, during an online assessment of online reasoning in higher education in the \textit{Critical Online Reasoning in Higher Education} (CORE) project~\citep{Nagel:2025,Hartig:et:al:2026}.
The assessment was conducted independently of this study.
As part of the original assessment, students were asked to search online for sources to support a written response, which was then graded by a subset of two to five trained raters.
Searching and evaluating online sources is part of the assessed construct~\citep{Molerov:et:al:2020,Nagel:et:al:2022}; the students were not supplied with any curated reading materials beyond the task and scenario.
The questions are divided into two domains, economics and medicine, each containing four tasks.
We study the responses in the \textit{practical} context (also referred to as \textit{clinical reasoning} in medicine) in the CORE assessment.
The four economics tasks require students to answer questions that ask for arguments both for and against a concept or topic introduced in a scenario text:
\begin{itemize}
    \item \TaskEcoNU{} (\TaskEcoNUShort{}) asks for arguments regarding the use of personalised nudging to save energy.
    \item \TaskEcoPS{} (\TaskEcoPSShort{}) asks for arguments regarding the issue of pay gaps and different pay scales for pilots.
    \item \TaskEcoSU{} (\TaskEcoSUShort{}) asks for arguments regarding the key economic criteria for selecting a location when relocating a startup.
    \item \TaskEcoWP{} (\TaskEcoWPShort{}) asks for arguments regarding the economic considerations in choosing between a municipal wind park business model or no wind park development.
\end{itemize}
The medical tasks ask for diagnoses, symptoms, treatments and therapies in the following scenarios:
\begin{itemize} 
    \item \TaskMedAT{} (\TaskMedATShort{}) addresses breathing problems during the ascent of Mount Everest.
    \item \TaskMedAU{} (\TaskMedAUShort{}) addresses declining vision in the context of the mother having glaucoma.
    \item \TaskMedKL{} (\TaskMedKLShort{}) addresses circulatory problems resulting from a spider bite.
    \item \TaskMedMO{} (\TaskMedMOShort{}) addresses an ear infection in a toddler.
\end{itemize}

\begin{table}[!htbp]
    \centering
    \begin{tabular}{l|r|r|r|r}
       & \TaskEcoNU{} & \TaskEcoPS{} & \TaskEcoSU{} & \TaskEcoWP{} \\
       \toprule
       \#Responses & \numprint{122} & \numprint{81} & \numprint{81} & \numprint{86} \\
       \midrule
       Characters \\
       - Mean & \numprint{1125.55} & \numprint{1089.77} & \numprint{1181.99} & \numprint{1063.70} \\
       - SD & \numprint{509.14} & \numprint{507.29} & \numprint{505.85} & \numprint{496.25} \\
       - Min & \numprint{0} & \numprint{0} & \numprint{0} & \numprint{0} \\
       - Max & \numprint{2364} & \numprint{2284} & \numprint{2402} & \numprint{2367} \\
       \midrule
       Sentences \\
       - Mean & \numprint{8.24} & \numprint{8.28} & \numprint{9.06} & \numprint{8.42} \\
       - SD & \numprint{4.28} & \numprint{4.24} & \numprint{4.86} & \numprint{5.25} \\
       - Max & \numprint{21} & \numprint{19} & \numprint{25} & \numprint{36} \\
       \midrule
       Tokens/Sent. \\
       - Mean & \numprint{21.08} & \numprint{20.12} & \numprint{18.89} & \numprint{19.24} \\
       - SD & \numprint{10.60} & \numprint{7.66} & \numprint{7.89} & \numprint{7.97} \\
       - Max & \numprint{82.00} & \numprint{48.50} & \numprint{40.00} & \numprint{35.75} \\
    \bottomrule
    \end{tabular}
    \caption{Student response statistics for the economics tasks.}
    \label{tab:responses_statistics_eco}
\end{table}

\begin{table}[!htbp]
    \centering
    \begin{tabular}{l|r|r|r|r|r|r|r|r|}
       & \TaskMedAT{} & \TaskMedAU{} & \TaskMedKL{} & \TaskMedMO{} \\
       \toprule
       \#Responses & \numprint{48} & \numprint{58} & \numprint{55} & \numprint{100} \\
       \midrule
       Characters \\
       - Mean & \numprint{1109.54} & \numprint{859.69} & \numprint{979.24} & \numprint{1108.55} \\
       - SD & \numprint{573.16} & \numprint{488.06} & \numprint{497.13} & \numprint{607.73} \\
       - Min & \numprint{0} & \numprint{0} & \numprint{0} & \numprint{0} \\
       - Max & \numprint{2860} & \numprint{1941} & \numprint{2212} & \numprint{3475} \\
       \midrule
       Sentences \\
       - Mean & \numprint{9.83} & \numprint{7.98} & \numprint{8.15} & \numprint{9.39} \\
       - SD & \numprint{5.10} & \numprint{4.09} & \numprint{4.63} & \numprint{5.29} \\
       - Max & \numprint{25} & \numprint{18} & \numprint{20} & \numprint{28} \\
       \midrule
       Tokens/Sent. \\
       - Mean & \numprint{16.70} & \numprint{15.29} & \numprint{18.59} & \numprint{16.62} \\
       - SD & \numprint{4.97} & \numprint{6.90} & \numprint{7.81} & \numprint{6.96} \\
       - Max & \numprint{26} & \numprint{30} & \numprint{36} & \numprint{37} \\
    \bottomrule
    \end{tabular}
    \caption{Student response statistics for the medicine tasks.}
    \label{tab:responses_statistics_med}
\end{table}

\begin{figure}[!htbp]
    \centering
    \includegraphics[trim=0 0 5cm 0,angle=90,origin=c,width=.4\textheight]{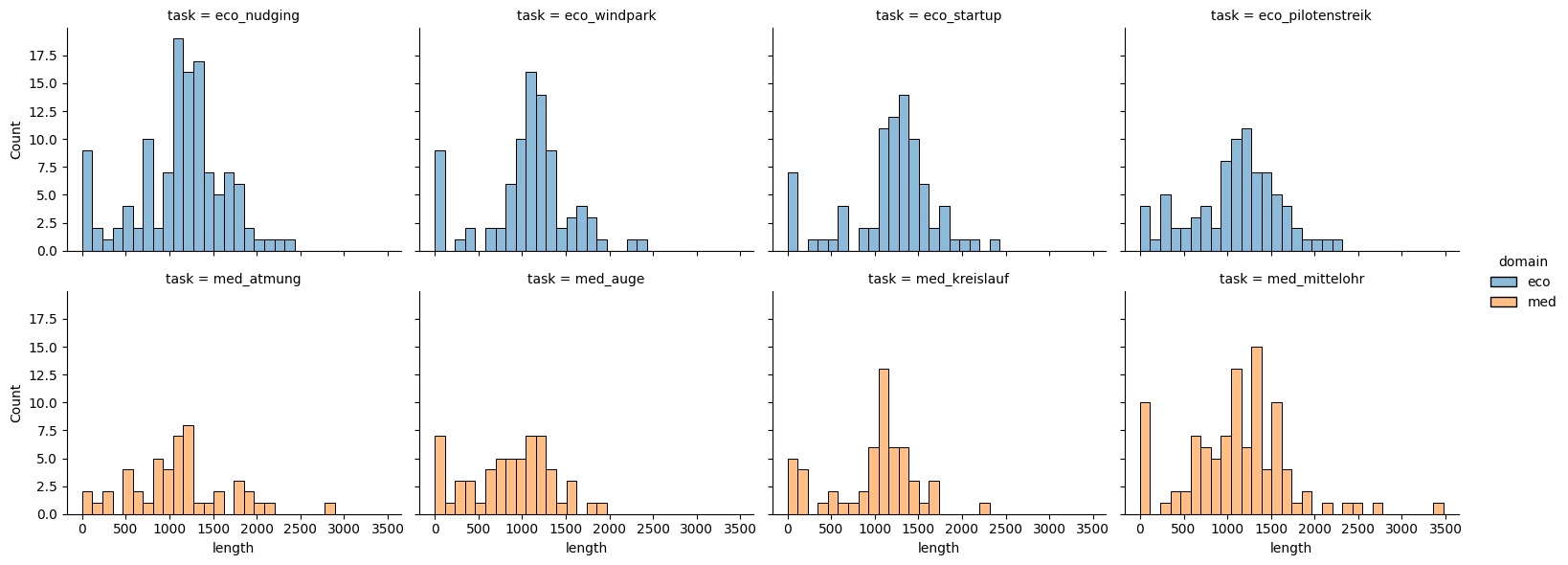}
    \caption{Distribution of student response texts.}
    \label{fig:responses_distribution}
\end{figure}

In total, we consider 631 student-written statements, with an average of 79 responses per task (see Tables~\ref{tab:responses_statistics_eco} and~\ref{tab:responses_statistics_med} for statistics on responses length and number sentences, and \autoref{fig:responses_distribution} for the underlying distribution).
Task design, time limits, and length requirements were defined by the domain experts of the original assessment.
The students were given a limited amount of time to complete the task and were asked to ensure that their responses were between 150 and 250 words for economics and no more than 300 words for medicine.
We notice a few outliers in the medical responses, but the distribution is generally similar across both domains.
These responses were rated by two to four human annotators in economics and three to five human annotators in medicine.
The ratings are based on 115 criteria, comprising two general criteria shared across all eight tasks and multiple task-specific criteria.
The rubric and criteria were developed as part of the original assessment to assess the REAS facet of COR and were based on prior performance-assessment research~\citep{Molerov:et:al:2020,Nagel:et:al:2022}, with task development being based on an Evidence-Centered Design approach~\citep{Mislevy:2018}.
The process for domain-general COR is described by \citet{Hartig:et:al:2026} and the domain-specific assessments in medicine and economics were developed within the same methodological framework.
The students did not have access to the scoring rubric, they were only presented with the task.
We use the original rubric without redesigning its criteria or constructs for the present LLM-based study.
However, the wording and presentation of individual criteria were varied across prompt configurations.
We excluded two criteria relating to consulted sources with and without traceable web references, as assessing them would have required giving the LLMs access to the web. 
Again, this would have raised data-protection concerns by making the students' web searches traceable.
Additionally, the question about the initial working diagnosis is not considered in the medicine tasks, leaving a total of 95 criteria for the LLM-based system.

The grading rubric comprises the following two task-independent criteria:
\begin{enumerate}
    \item The \CritFilter{} criterion assesses whether the student engaged seriously with the task and did not entirely miss the subject matter.
    This Boolean criterion is formulated as a \enquote{\textit{yes}/\textit{no}} question, with \enquote*{no} encoded as 0.
    \item The \CritCoherence{} criterion relates to the assessment of the response's coherence, asking whether the student's argumentation is coherent, consistent, and free of contradictions. 
    It is based on a discrete four-point scale from 0 to 3, where 0 is the lowest rating, that is, the response is completely incoherent.
\end{enumerate}

We further consider the following five economics criteria in our analysis:
\begin{enumerate}
    \item The Boolean \CritReference{} criterion asks whether the student's written statement refers to the problem situation.
    \item The Boolean \CritConclusion{} criterion asks whether the student's written statement reaches a clear conclusion.
    \item The Boolean \CritArgumentation{} criterion asks whether the student's written statement contains coherent economic argumentation.
    Note that this criteria was not rated for the \TaskEcoPS{} task.
    This task therefore only considers four specific criteria.
    \item The \CritArgumentsCount{} criterion asks whether the student's written statement contains comprehensible arguments.
    It counts up to four arguments and is therefore based on a discrete five-point scale from 0 to 4, where 0 is the lowest rating, indicating that no argument is provided.
    \item The Boolean \CritSpecificArguments{} criterion asks whether the student's statement mentions both arguments for and against personalised nudging.
\end{enumerate}
The rubric used to assess the economics tasks contains criteria that can be grouped into the following three categories:
The first category is concerned with \textit{formal} criteria, for which the rater needs to check for the existence of specific elements.
These criteria essentially act as filters at different levels:
(1) Has the student made an actual attempt to work on this task? (\CritFilter{} criterion);
(2) if the student has worked earnestly on the task, does their response consider the stated problem? (\CritReference{}); and
(3) does the student formulate a conclusion, or does the response remain unresolved? (\CritConclusion{}).
The second category is concerned with the \textit{structure} of the response.
This maps to the two non-Boolean criteria in the economics tasks:
the rater needs to consider the text in full and construct some form of argumentative structure rather than only checking for the presence of specific elements.
They have to quantify the coherence (\CritCoherence{} criterion) on a predefined scale and count the number of arguments made (\CritArgumentsCount{}).
Finally, the third category assesses \textit{domain knowledge}.
Unlike the other criteria, which can mostly be graded without specific domain knowledge, raters need to apply their subject knowledge to judge whether students have correctly done so in case of the \CritArgumentation{} and \CritSpecificArguments{} criteria.
Depending on the particular task, 12 to 20 task-specific criteria are rated for the medicine tasks, including the two shared criteria.
All of these criteria are Boolean criteria formulated as \enquote{\textit{yes}/\textit{no}} questions.
\begin{enumerate}
    \item The \CritMed{A1} criterion asks whether the student has identified the most likely differential diagnosis.
    \item The \CritMed{B*} criteria ask for additional mentioned differential diagnoses.
    For the \TaskMedKL{} task, there are up to seven \CritMed{B*} criteria.
    \item The \CritMed{C*} criteria differ by task:
    in task \TaskMedAT{}, they ask whether measures indicated for acute care are listed;
    in \TaskMedAU{}, whether tests to rule out any incorrect differential diagnoses are mentioned;
    in \TaskMedKL{}, whether an explanation of affected organs and systems is given;
    and in \TaskMedMO{}, whether at least one necessary test to confirm the suspected diagnosis is provided in the response.
    \item The \CritMed{D*} criteria ask for tests to rule out or confirm differential diagnoses (\TaskMedAU{} and \TaskMedMO{}), treatment options (\TaskMedAT{}), and organ-specific symptoms (\TaskMedKL{}).
    \item The \CritMed{E*} criteria ask whether therapies or measures are mentioned for all tasks except \TaskMedAU{}.
    \item The following criteria only apply to the \TaskMedMO{} task:
    The \CritMed{F*} criteria ask for a recommendation for action;
    the \CritMed{G*} criteria ask for naming complications;
    and the \CritMed{H*} criteria ask whether at least one plausible sign of complications is mentioned.
\end{enumerate}
Note that the criteria used in the medical assessment, with the exception of \CritFilter{} and \CritCoherence{}, are not as directly comparable as the economics criteria.
Because of the naming scheme, we still present them grouped by name, similar to the economics criteria, and provide a thematic grouping.
In contrast to the economics tasks, all require domain knowledge.
The first category contains criteria that ask for \textit{diagnoses} and diagnostic testing.
It asks what the condition is and how it is confirmed or excluded.
This group includes the criteria \CritMed{A}, \CritMed{B}, \CritMed{C} in tasks \TaskMedAU{} and \TaskMedMO{} and \CritMed{D} in the same two tasks.
The second category includes criteria that are concerned with \textit{symptoms}, complications or clinical features: \CritMed{C} and \CritMed{D} for \TaskMedKL{}, and \CritMed{G} and \CritMed{H}.
The final group asks for \textit{treatments}, or management more generally.
It comprises the remaining criteria: \CritMed{C} (\TaskMedAT{}), \CritMed{D} (\TaskMedAT{}), \CritMed{E} and \CritMed{F}.

Five of the seven criteria for the economics tasks are Boolean (\textit{yes}/\textit{no}), while the other two use discrete scales with four (\CritCoherence{}) and five (\CritArgumentsCount{}) values.
Similarly, except for the task-independent \CritCoherence{} criterion, all criteria are Boolean for the medicine tasks.
In all cases, the lowest, or \enquote{default}, criterion value is 0.

A \CritSumscore{} is provided for each participant based on all criteria except \CritFilter{}.
The \CritSumscore{} is calculated as the average of the remaining criterion values, normalised to a range from 0 to 1.
This composite score can be interpreted as an overall measure of student performance.
This makes the student performance comparable across different tasks and domains, even though the criteria themselves cannot be compared directly in all cases.
Although we do not predict the \CritSumscore{} using the LLM-based AES, we include it in the evaluation, bringing the total number of criteria examined to 103.
The \CritFilter{} criterion is treated as a filter in the assessment and grading of the student response:
If a rater sets this criterion to 0, that is, if the student did not engage seriously with the question, all other criteria are assigned the lowest value.

\begin{table}[!t]
    \centering
    \begin{tabular}{l|r|r|r|r|r|r|r|r}
         & \TaskEcoNUShort{} & \TaskEcoPSShort{} & \TaskEcoSUShort{} & \TaskEcoWPShort{} & \TaskMedATShort{} & \TaskMedAUShort{} & \TaskMedKLShort{} & \TaskMedMOShort{} \\
         \toprule
        \#Ratings & \numprint{2267} & \numprint{1294} & \numprint{2096} & \numprint{1671} & \numprint{2208} & \numprint{2496} & \numprint{3120} & \numprint{8240} \\
        Avg. & \numprint{20.80} & \numprint{17.97} & \numprint{27.58} & \numprint{20.89} & \numprint{48} & \numprint{48} & \numprint{60} & \numprint{90.55} \\
        -- SD & \numprint{1.15} & \numprint{0.17} & \numprint{1.58} & \numprint{0.80} & \numprint{0} & \numprint{0} & \numprint{0} & \numprint{10.47} \\
        -- Min & \numprint{14} & \numprint{17} & \numprint{21} & \numprint{14} & \numprint{48} & \numprint{48} & \numprint{60} & \numprint{60} \\
        -- Max & \numprint{21} & \numprint{18} & \numprint{28} & \numprint{21} & \numprint{48} & \numprint{48} & \numprint{60} & \numprint{100} \\
        \midrule
        Avg. rater & \numprint{2.97} & \numprint{3} & \numprint{3.95} & \numprint{2.99} & \numprint{3} & \numprint{4} & \numprint{3} & \numprint{4.53} \\
        -- SD & \numprint{0.16} & \numprint{0} & \numprint{0.22} & \numprint{0.11} & \numprint{0} & \numprint{0} & \numprint{0} & \numprint{0.52} \\
        -- Min & \numprint{2} & \numprint{3} & \numprint{3} & \numprint{2} & \numprint{3} & \numprint{4} & \numprint{3} & \numprint{3} \\
        -- Max & \numprint{3} & \numprint{3} & \numprint{4} & \numprint{3} & \numprint{3} & \numprint{4} & \numprint{3} & \numprint{5} \\
        \bottomrule
    \end{tabular}
    \caption{Statistics on the human ratings of the student responses to the eight economics and medicine tasks.}
    \label{tab:rater_statistics}
\end{table}

In total, we consider \numprint{7328} human ratings of the student responses for the economics tasks, and \numprint{16064} for the medicine tasks.
See \autoref{tab:rater_statistics} for statistics on the ratings and their distribution across the task.
Each statement was rated by at least two raters in the \TaskEcoNU{} and \TaskEcoWP{} tasks and by up to five raters in the \TaskMedMO{} task.
On average, the number of raters per response is similar across both domains, at 3.2 (economics) and 3.6 (medicine).
Not every rater evaluated every response, which is e.g., visible in case of the \TaskMedMO{} task, where we report between three and five raters per response.
The original assessment aimed for at least two raters per statement; however, the actual number of raters varies between responses and tasks.
See Figures~\ref{fig:humans_llms_rating_distribution_eco} (economics) and~\ref{fig:humans_llms_rating_distribution_med} (medicine) in the appendix for an overview of the distribution of ratings per criterion.
We observe a tendency towards more positive values on the rating scale for most criteria and for raters considering the economy tasks.
Notably, rater B often shows a distribution that differs from those of the others.
Few ratings for the \CritFilter{} and \CritReference{} criteria are assigned negative values.
Compared with the economics criteria, the distribution of the medical criteria shows more \enquote{no} ratings, as these criteria often ask whether a specific symptom is mentioned.
In our real-world scenario, most students produced gradable responses, as reflected in the \CritFilter{} criterion, which was set to 1 in 94\% of cases across all raters and tasks (SD~\numprint{2}, min~\numprint{91.2}, max~\numprint{97.2}).

For reference, \autoref{tab:methods_numbers_summary} summarizes the number of responses, raters, and criteria across all tasks.

\begin{sidewaystable}[!p]
    \centering
    \begin{tabular}{l|r|r|r|r|r|r|r}
      \toprule
      Task & Responses & Raters per Resp. & Original Criteria & Excluded & Scored by LLM & \CritSumscore{} & Evaluated \\
        \midrule
        \TaskEcoNU{} & 122 & 2--3 & 9 & 2 & 7 & 1 & 8 \\
        \TaskEcoPS{} & 81 & 3 & 8 & 2 & 6 & 1 & 7 \\
        \TaskEcoSU{} & 81 & 3--4 & 9 & 2 & 7 & 1 & 8 \\
        \TaskEcoWP{} & 86 & 2--3 & 9 & 2 & 7 & 1 & 8 \\
        \midrule
        \TaskMedAT{} & 48 & 3 & 19 & 3 & 16 & 1 & 17 \\
        \TaskMedAU{} & 58 & 4 & 15 & 3 & 12 & 1 & 13 \\
        \TaskMedKL{} & 55 & 3 & 23 & 3 & 20 & 1 & 21 \\
        \TaskMedMO{} & 100 & 3--5 & 23 & 3 & 20 & 1 & 21 \\
        \midrule
        Total & 631 & 2--5 & 115 & 20 & 95 & 8 & 103 \\
      \bottomrule
    \end{tabular}
    \caption{Summary of the number of responses and human raters per response for each task, the number of criteria in the original rubrics, the criteria excluded from this study, the criteria scored by the LLM-based raters, the derived holistic \CritSumscore{} scores, and the total number of criteria evaluated.
    Totals are summed across tasks.}
    \label{tab:methods_numbers_summary}
\end{sidewaystable}

Student assistants from the economics department and doctoral students in medicine produced the human ratings as part of the original assessment carried out by CORE.
The economic and medicine ratings were conducted independently.
Before the main annotation, the raters performed a training, including explanations of the assessment tasks and criteria, a trial rating, and discussion about the rating results and agreements.
The main ratings that we use in this study, are preliminary and have not been consolidated into a gold standard, i.e., a single rating for each criterion per participant.
For this reason, we compare our LLM-based ratings with those of each human rater separately.

\subsection{LLM-based AES}
\label{sec:methods_llms}

\begin{table}[!htbp]
    \centering
    \begin{tabular}{l|r|l}
    \toprule
        Model & Size & Reference \\
        \midrule
        DeepSeek-R1 & 70b & \citet{DeepSeek:et:al:2025b} \\
        GPT-OSS & 20, 120b  & \citet{OpenAI:et:al:2025} \\
        Llama 3.3 & 70b & \citet{Grattafiori:et:al:2024} \\
        MiniCPM-V & 8b & \citet{Yao:et:al:2024} \\
        Mixtral & 8x7b & \citet{Jiang:et:al:2024} \\
        Nemotron & 70b & \citet{Wang:et:al:2024b} \\
        Qwen 2.5 & 32b & \citet{Yang:et:al:2024} \\
        MedGemma 1.5 & 4b & \citet{Sellergren:et:al:2026} \\
        Plutus & 8b & \citet{Royce:2024} \\
    \bottomrule
    \end{tabular}
    \caption{Overview of the LLMs used in \ExpAll{}.}
    \label{tab:llms_exp2}
\end{table}

\begin{table}[!htbp]
    \centering
    \begin{tabular}{l|r|l}
    \toprule
        Model & Size & Reference \\
        \midrule
        DeepSeek-R1 & 32, 70b & \citet{DeepSeek:et:al:2025b} \\
        DeepSeek v3.1 & 671b & \citet{DeepSeek:et:al:2025a} \\
        Gemma 2 & 27b & \citet{Gemma:et:al:2024} \\
        GPT-OSS & 20, 120b & \citet{OpenAI:et:al:2025} \\
        Llama 3.2 & 3b & \citet{Grattafiori:et:al:2024} \\
        Llama 3.3 & 70b & \citet{Grattafiori:et:al:2024} \\
        MiniCPM-V & 8b & \citet{Yao:et:al:2024} \\
        Mistral & 7b & \citet{Jiang:et:al:2023} \\
        Mixtral & 8x7b & \citet{Jiang:et:al:2024} \\
        Nemotron & 70b & \citet{Wang:et:al:2024b} \\
        Qwen 3 & 0.6, 1.7, 4, 8, 14, 32b &  \citet{Yang:et:al:2025} \\
        VibeThinker & 1.5b & \citet{Xu:et:al:2025} \\
    \bottomrule
    \end{tabular}
    \caption{Overview of the LLMs used in \ExpNu{}.}
    \label{tab:llms}
\end{table}

We employ an LLM-based AES system to rate each student response on the basis of the respective criteria per task.
The LLM receives the student text, the rating criteria, and optional additional information, and is then prompted to assign a value to each criterion.
In total, we experiment with 15 LLMs of various sizes, using \textit{Docker Unified UIMA Interface} (DUUI)~\citep{Leonhardt:et:al:2023} as the NLP framework and LangChain~\citep{Chase:2022} and Ollama~\citep{Ollama:2023} for LLM processing.
All LLMs were queried with the same random seed and a \texttt{temperature} of 0.8.
We fixed the temperature to the default value of LangChain and Ollama across all models to provide a common configuration for evaluation and did not perform model-, prompt-, or task-specific optimization regarding these parameters.
Except for the GPT-OSS models, all models were instructed to return JSON via the \texttt{format} parameter.
Depending on the prompt, we increase the context size (\texttt{num\_ctx}) and leave the number of generated tokens unrestricted (\texttt{num\_predict} = -2) in all cases.
The list of nine studied LLMs in \ExpAll{} is given in \autoref{tab:llms_exp2}, and \autoref{tab:llms} shows the twelve models of \ExpNu{}, six of which also appear in \ExpAll{}.
Together, they include multiple versions and configurations of DeepSeek, GPT-OSS, Llama, and Qwen, as well as MiniCPM-V, Mixtral and Nemotron.
Additionally, we include the MedGemma and Plutus LLMs, which are trained on medical and economic data, respectively.
In \ExpAll{}, we cover all eight tasks using 49 LLM and prompt configurations based on nine LLMs (\autoref{tab:llms_exp2}) for \TaskEcoWPShort{}, \TaskMedAUShort{}, \TaskMedKLShort{}, and \TaskMedMOShort{}; 50 for \TaskMedATShort{}; and 51 for \TaskEcoPSShort{}, \TaskEcoSUShort{}, and \TaskEcoNUShort{}, respectively.
The aim of this first study is to test whether LLM-based ratings can be generalised across the two domains and their eight tasks.
We study 79 additional prompt and LLM configurations based on twelve LLMs (\autoref{tab:llms}) to perform the LLM-based rating of student responses to the economics \TaskEcoNU{} task only.
The prompts included in \ExpNu{} examine two approaches:
first, prompting for all criteria; and second, using prompts dedicated only to the two non-Boolean criteria.
The two studies use partly different sets of LLMs (see Tables~\ref{tab:llms_exp2} and~\ref{tab:llms}).
While \ExpAll{} covers all eight tasks with a fixed set of models, including the domain-specific MedGemma and Plutus, \ExpNu{} analyses a single task in more depth and covers a wider range of model sizes, versions, and families.
Six models are included in both studies.
To assess the consistency of the ratings under the non-deterministic temperature setting, we check whether the selected model configuration results in unstable ratings.
We evaluate this on a subset of LLM--prompt combinations from \ExpNu{}, which we query to rate each response five times using different random seeds, keeping the other parameters fixed.

The prompts are mostly written in German, the language of the statements.
An example template for each prompt variant of \ExpAll{} is shown in \ref{sec:appendix:prompts}.
All prompt templates of both studies are provided as supplementary material, with placeholders for the task texts, scenarios, rubrics, and student data.
In the following, we summarise the prompt variants.
We consider the following prompt variants in \ExpAll{}.
\begin{itemize}
    \item \texttt{SIMPLE SCENARIO} The prompts in this category ask the LLM to produce ratings for all criteria in a single request as a structured, predefined JSON object.
    We provide the same information that was available to human annotators in our annotation tool when they assessed the scoring criteria and their value ranges.
    This includes the relevant scenario text for the task question.
    Additionally, the LLM is informed that the response comes from a student answering a question within a specific domain.
    The prompt assigns the LLM the role of an annotation expert.
    This variant serves as the baseline for the other variants, each of which extends or modifies it.
    \item \texttt{CHAIN NOANSWER SCENARIO}    
    These prompts make multiple requests to the model, rating each criterion separately.
    Criterion definitions are not provided in advance, but only when each criterion is requested.
    However, the model has access to the full message history.
    This decomposition relates to the chain-of-thought prompting~\citep{Wei:et:al:2022}, which has shown to improve the accuracy of LLM-based scoring~\citep{Lee:et:al:2024}.
    \item \texttt{FEATURES SCENARIO}
    This group of prompts builds on \texttt{SIMPLE SCENARIO} with respect to the definition of criteria and JSON output.
    Additionally, we provide the LLM with a range of features derived from the students' answer texts.
    Such text features are the basis of feature-based AES~\citep{Burstein:et:al:1998,Rudner:Liang:2002,Yannakoudakis:et:al:2011,Chen:et:al:2013,Cozma:et:al:2018}, which remains competitive with LLM-based scoring approaches~\citep{Huang:et:al:2025}.
    These features include basic statistics such as token, sentence, unique-token, and lemma counts, as well as hate-speech and toxicity detection, emotion recognition, sarcasm detection, topic classification, sentiment analysis, and readability measures.
    The following natural language processing (NLP) systems are used to calculate these feature groups:
    Statistics are derived from preprocessing the responses with spaCy~\citep{Honnibal:et:al:2020} (efficient model).
    Additional features are generated using implementations in the DUUI NLP framework:
    The readability features comprise nine metrics, including the Flesch-Kincaid grade level and the Automated Readability Index, and are calculated using the implementation by \citet{DiMascio:2020}.
    We use text2ddc~\citep{Uslu:Mehler:Baumartz:2019,Uslu:Mehler:Niekler:Baumartz:2018} for topic classification using the Dewey Decimal Classification scheme, as well as a transformer-based model using the \textit{Media Topic schema} of the IPTC~\citep{Kuzman:Ljubesic:2025a,Kuzman:Ljubesic:2025b}.
    Sentiment analysis is performed using the lexicon- and rule-based tool VADER~\citep{Hutto:Gilbert:2014,Tymann:et:al:2019} and two additional transformer models~\citep{Guhr:et:al:2020,Barbieri:et:al:2022}.
    All further features are based on transformer models for detecting
    hate speech~\citep{Aluru:et:al:2020,deepset:2020}, 
    toxicity~\citep{Stakovskii:2022,AIstrova:2023},
    emotions~\citep{Bianchi:et:al:2022,Bertolini:et:al:2024}, and
    sarcasm~\citep{Helin:2022}.
    All of these text features are provided to the LLM as raw JSON containing the feature group, model name, and values.
    \item \texttt{OPT SCENARIO} We study a prompt that was automatically optimized using Claude Opus~4.7~\citep{Anthropic:2026}.
    Since LLM-based scoring is sensitive to the prompt and no prompt performs best across tasks~\citep{Mansour:et:al:2024,Lee:et:al:2024}, we let the LLM optimize the \texttt{SIMPLE SCENARIO} prompt.
    In a second variant, we instruct the LLM to generate a prompt based on all prompt templates studied in this paper.
    Claude was selected as a state-of-the-art model during prompt development and limited to generating the updated prompt.
    It was used as a complementary proprietary LLM to the locally-hostable models used in our study.
    Following the data protection rules, the proprietary Claude model received only the prompt templates, without any inserted content such as the training transcripts or student data.
    The final prompt was fixed before evaluation and applied to all evaluated models in the study without change.
    \item \texttt{TRAINING SCENARIO} This prompt category extends \texttt{SIMPLE SCENARIO} with additional context from a transcript of a rater meeting comprising a general introduction to the assessment that covers the \CritFilter{} and \CritCoherence{} criteria, the annotation tool, and the rating of student statements.
    The transcript is included in full in the prompt.
    Since prior work reports higher agreement when more detailed rubrics are provided~\citep{Frohn:et:al:2025}, this variant gives the LLM some of the information that the human raters received before rating.
\end{itemize}

In \ExpNu{}, we perform a focused follow-up on the \TaskEcoNU{} task.
We use the same prompt classes as described above and test additional prompts and models.
One important distinction is the exclusion of the scenario description.
The question for the economics task \TaskEcoNU{} is phrased very directly as \textit{key arguments for and against personalized nudging to promote energy conservation}, whereas the other tasks require knowledge from the scenario text.
For this reason, we omit the scenario from the prompts in this experiment.
\begin{itemize}
    \item \texttt{SIMPLE} and \texttt{BASIC}:
    These prompts mirror the \texttt{SIMPLE SCENARIO} prompt from \ExpAll{} but exclude the scenario text.
    We add the \texttt{BASIC} prompt, an English prompt with a question--answer structure, that mainly differs in the question posed to the students:
    The \texttt{SIMPLE} and all other prompts focus on the written statement of arguments for and against, whereas \texttt{BASIC} considers the scenario and the online research needed to produce the answer.
    \item \texttt{CHAIN}: This prompt uses the \texttt{CHAIN NOANSWER SCENARIO} prompt as a basis, again without the scenario text.
    In addition to the criterion definition, each message also contains the full student response, so that the answer is always kept in the most recent message.
    We prompt the LLM for explicit JSON output and provide the criterion definitions accordingly; in an alternative version, we do not specify an output format and instead provide the definitions in plain text.
    \item \texttt{FEATURES}: Building on the \texttt{FEATURES SCENARIO} prompt group from \ExpAll{}, we add three sentence-relation features to the prompt.
    Entailment, justification, and elaboration scores are generated by prompting an LLM (DeepSeek-R1:70b) to assign a confidence score to each sentence pair, for example, by asking whether one sentence entails the other.
    We then use the confidence scores for the first--last and last--first sentence pairs as two features in the criteria-rating prompt.
    The features are provided to the LLM not only as raw JSON but also as summarised text in this experiment.
    To this end, we use different LLMs to generate texts that summarise the most important characteristics of the student response and its text features.
    This summary is then used in the rating prompt instead of the raw feature data.
    \item \texttt{OPT}:
    We optimise the \texttt{SIMPLE} prompt using the \textit{PE2} approach of \citet{Ye:et:al:2024} instead of only relying on the zero-shot capabilities of an LLM for optimisation as in \texttt{OPT SCENARIO}.
    The \textit{PE2} algorithm uses one LLM to generate prompt variants, which are then automatically validated by another LLM on a few examples.
    We run the optimisation for five iterations, generating five new prompts each time (\texttt{T}=5, \texttt{n}=4, \texttt{m}=5, \texttt{k}=5) and rank them by the proportion of correctly predicted criteria.
    In this way, the \texttt{OPT} prompt uses an optimised system prompt based on \texttt{SIMPLE}.
    \item \texttt{TRAINING}:
    We extend the \texttt{TRAINING SCENARIO} prompt by adding a transcript of a task-specific presentation of the economics task.
    In this experiment, we experiment with providing only the general discussion or both parts combined, using either the full transcript or an LLM-generated summary.
\end{itemize}

As in \ExpAll{}, the prompts considered in \ExpNu{} cover all criteria.
Additionally, we analyse the two non-Boolean criteria, \CritCoherence{} and \CritArgumentsCount{}, using dedicated prompts (37 prompts and 8 prompts, respectively) that consider no other criteria.
These are classified as follows:
\begin{itemize}
    \item \texttt{COHERENCE VERY SIMPLE}:
    We employ different prompts for the coherence criterion, all using structured input, that is, a tag-like format enclosing the LLM inputs and contexts.
    This prompt is deliberately simple and asks only for a coherence rating, without further definition or explanation.
    \item \texttt{COHERENCE STRUCT}:
    Three prompts test minor formulation changes, such as capitalisation and the placement of task context for the response being rated.
    \item \texttt{COHERENCE STRUCT FEATURES}:
    As with the all-criteria prompts, we incorporate several text features and instruct the LLM to use them to assess text coherence.
    \item \texttt{COHERENCE STRUCT TRAINING}:
    As in the previous prompts, we experiment with including different variants of the rater-meeting transcript.
    \item \texttt{COHERENCE STRUCT CHAIN FEWSHOT}:
    Finally, we prompt the LLMs for a coherence rating using a few examples in two ways: (1) without specifying an output format, followed by a chained prompt requesting the value in JSON, and (2) by directly requesting structured output for comparison.
    In one variant, we begin with a prompt asking the LLM to analyse the examples and formulate rules for the given ratings.
    \item \texttt{ARGUMENTS COUNT CGPT}:
    Based on \texttt{SIMPLE}, these prompts were optimised by asking ChatGPT~4.5~\citep{OpenAI:2026b} to refine them, yielding two variants: one provides information about the question answered by the student, and the other instructs the LLM more generally to extract arguments from the student response.
    The output is expected to follow a predefined JSON schema consisting of a list of arguments the model can extract from the student's answer.
    This includes the argument itself, a description of it, and a justification for its inclusion.
    ChatGPT received only the prompt templates, without any context such as student data.
    \item \texttt{ARGUMENTS COUNT STRUCT}:
    Prompts in this class build on the \texttt{AR\-GU\-MENTS COUNT CGPT} prom\-pt and use a structured format.
    In one variant, we ask the LLM to merge multiple occurrences of the same argument into a single output instance.
\end{itemize}
For this evaluation, we ignore the \CritFilter{} criterion and ask the model for a rating regardless of the quality threshold imposed by the filter.

In total, counted per task, this amounts to 523 model--prompt configurations:
399 in \ExpAll{} (49--51 per task), 79 in \ExpNu{}, and 45 criterion-specific prompts for \CritCoherence{}~(37) and \CritArgumentsCount{}~(8).

As a result of experimenting with different LLMs and the prompting strategies described above, we consider a total of \numprint{346386} ratings produced by the LLMs for \ExpAll{} and \numprint{65947} for \ExpNu{} (including the criterion-specific prompts).
Figures~\ref{fig:humans_llms_rating_distribution_eco} and~\ref{fig:humans_llms_rating_distribution_med} in the appendix show the distribution of ratings across all considered LLM results for the economics and medicine tasks in comparison to the human ratings.
Similar distributions can be observed for the \CritFilter{} and \CritReference{} economics criteria and, with some exceptions, for \CritSpecificArguments{} and \CritConclusion{} when compared with the human raters’ distributions.
For the medicine tasks, the criteria distributions show less similarity.

In case of the economics \TaskEcoNU{} task, the \CritFilter{} criterion was rated as \texttt{false} in about 4.2\% of the ratings, i.e., the students were assessed by the LLMs as having performed the task inadequately.
In comparison, the other three economics tasks, as well as all four medicine tasks, produced only positive \CritFilter{} ratings.

To account for unbalanced data, that is, not all responses were rated by all human raters and not all models produced results for all responses, we use the following strategies to calculate agreement.
As a baseline, we treat missing LLM ratings as if the corresponding criteria had received the lowest value.
This allows us to handle technical problems that may arise during model querying or output parsing in a predictable manner for this analysis.
This leads to variants for handling the LLM ratings based on the filter rating: (1) \texttt{strict}, which sets all other criteria to the lowest, or default, value when the \CritFilter{} is false, as described above; and (2) \texttt{lenient}, which retains the criterion values produced by the LLM.
\texttt{Lenient} allows the model to recover from an incorrectly rated \CritFilter{} criterion and can simplify and speed up prompting, as a single request covers all criteria, reducing costs and improving performance, for example, in real-time feedback.
In the remainder of this analysis, we only consider the \texttt{strict} \CritFilter{} handling, since the ratings are fully one-sided, i.e., \textit{positive} for all tasks except economics \TaskEcoNU{}.
There, we do not see any differences as the \CritFilter{} criterion is still extremely skewed.

We further evaluate whether a human rater can be replaced by an automatically generated LLM rating by examining inter-annotator agreement using Krippendorff's alpha~\citep{Krippendorff:2011}, which measures how consistently annotators agree beyond chance.
In this way, we assess how strongly a model affects IAA and whether replacing a human rater with an LLM results in an increase or decrease of the agreement in relation to the observed agreement between the human annotators.
A non-negative change in IAA indicates that an LLM can be substituted for a human while preserving the overall group agreement.
However, it does not make any argument regarding the validity of the rating itself, because no consolidated gold rating is available.
We are analysing whether LLMs have the potential to reach a level of agreement comparable to the human annotators under the same rubric-based assessment setting.
Each human rater is replaced in the analysis.
Our objective is not to identify or approximate a specific individual rater using an LLM-generated rating, but to evaluate whether LLMs can substitute raters in the group while preserving overall group agreement.
We select IAA for measuring the results as we do not compare the LLM-generated ratings to a single consolidated gold score, we treat the individual human ratings as equally plausible and evaluate agreement over the group of humans and LLM-based raters.
Krippendorff's alpha is used as the main method because it measures agreement beyond chance, supports more than two raters and handles missing ratings~\citep{Artstein:Poesio:2008}.
These features are relevant to our study because the student statements were rated by different numbers and combinations of human annotators.
In comparison, QWK, a weighted form of Cohen's kappa that is widely used for evaluating AES, supports primarily two raters, and Fleiss's kappa requires equal amounts of ratings per item and assumes randomly sampled annotators; Krippendorff's alpha has therefore been recommended as a general alternative in multi-rater evaluations~\citep{Doewes:et:al:2023,Klie:et:al:2024}.
However, Krippendorff's alpha can be sensitive to imbalanced data and more difficult to interpret.
We therefore report unanimous agreement, Cronbach's alpha and ICC(3,k) as complementary measures of rater consistency~\citep{Albatarni:et:al:2026}.
Additionally, we test whether the LLMs can replace at least 50\% of the human raters at a statistically significant level by employing the \textit{Alternative Annotator Test}.
It can be run on a small sample of 50--100 cases with at least three human annotators, without requiring consolidated gold annotations, and can incorporate annotator costs through a cost-benefit parameter.
The test compares each human annotator's alignment with the rest of the group to the LLM's alignment with that same group, tests the difference for each annotator, and computes the LLM's win rate from the rejected hypotheses.
Following the authors of the alt-test, we use a cost-benefit parameter of 0.15, which they present for skilled workers.
We consider this skill level to be the closest to the human annotators who rated the original student statements and additionally assess the dependence of the parameter choice in a sensitivity analysis.
Finally, we take a further step towards full automation by examining changes in agreement when two human raters are replaced, leaving at least one human rater as the necessary source of verification.
For this analysis, we replace all combinations of two human raters with all unique model--prompt configurations.
In addition to the top-performing configurations, we examine the top 10 configurations to evaluate reliable models and prompts overall.
That is, we select the top 10 configurations based on their best performance per task and criterion, measured by the highest IAA when replacing any of the human raters, including cases in which the change is negative.
We then rank them by how often they appear in the top 10, grouped by domain and criterion and evaluate their Jaccard similarity.
In this way, we address research questions \textit{RQ1} and \textit{RQ2}.

\subsection{SHAP Analysis}
\label{sec:methods_shap}

To study the influence of model and prompt choice, we perform a \textit{SHAP} (SHapley Additive exPlanations)~\citep{Lundberg:Lee:2017,Lundberg:et:al:2020} analysis on a proxy model trained to predict agreement.
For this purpose, we train proxy models that predict the observed IAA based on the experiment, prompt, and model characteristics as input features.
The proxy models are only used as explanatory models for the \textit{SHAP} analysis in this study and are not systems designed for predicting IAA in practical use outside of this study.
We experiment with both regression and classification models, first attempting to predict the actual IAA value and, second, whether a non-negative change can be achieved.
The 33 features include the task and task domain, the LLM and prompt, the LLM family and size, and the prompt length, style, and output formats.
They are grouped as follows:
\begin{enumerate}
    \item The \texttt{model} group contains the three model-related features: model size, family, and full name.
    \item The \texttt{prompt} group covers 17 features, including the prompt name, its class as a categorical feature, and multiple binary features.
    These encode whether the prompt is a chain prompt, contains training material or few-shot examples, is extended with textual features, contains the scenario text, is optimized, uses a structured method to provide the input, and specifies the expected output format.
    \item The domain and task names are used in the \texttt{task} feature group, as well as the criterion itself.
    \item The \texttt{rater} group consists of only one feature related to the human raters and contains the removed rater as a categorical feature.
    \item The \texttt{decoding} group includes the LLM parameters temperature, format, and context size.
    Note that temperature was the same in all experiments.
    \item The \texttt{resource} feature group summarises LLM application values, i.e., the mean, SD, min, and max statistics of the input and output tokens processed.
    These values are calculated as averages per task.
\end{enumerate}
We use \textit{Histogram-based Gradient Boosting} (HGBT) models for both regression and classification.
The targets are the IAA alpha value and Boolean non-negative change, respectively.
We include baseline models using the \textit{mean} strategy for regression and \textit{most frequent} strategy for classification.
For numerical features, we use a \textit{median imputer} and \textit{standard scaler}; for categorical features, we use a \textit{most frequent imputer} and \textit{one-hot encoder}.
The models are trained using the scikit-learn~\citep{Pedregosa:et:al:2011} implementation with default parameters and all results are obtained using 5-fold cross-validation, stratified for the classification model.
We then perform a \textit{SHAP} analysis using its \texttt{TreeExplainer} on the proxy models to determine which experimental characteristics contribute to the LLM performance.
Additionally, we examine the correlation between IAA and the model size as well as the number of processed tokens using Pearson and Spearman correlation coefficients calculated over the mean sizes per LLM--prompt configuration.
This analysis provides insights into research question \textit{RQ3}.

\section{Results}
\label{sec:results}

We evaluate the inter-annotator agreement among the human raters to determine the target quality that the LLMs must match to be able to replace humans in this rating task (\autoref{sec:results_human_iaa}).
Using Krippendorff's alpha, we first compare the LLMs' ratings with the human ratings and then proceed by replacing one human rater with LLMs (\autoref{sec:results_llm_replace}).
We confirm our results by using the \textit{Alternative Annotator Test} as an additional metric (\autoref{sec:results_alt_test}) and then replace two human raters (\autoref{sec:results_replacing_two}).
We summarise the results for RQ1 and RQ2 in \autoref{sec:results_iaa_alt}.
Finally, we present the results of the \textit{SHAP} analysis for RQ3 (\autoref{sec:results_shap}) and the consistency of the LLM-based ratings (\autoref{sec:results_consistency}).

\newpage
\begin{longtable}{l|l|l|l|r|r|r|r}
        \toprule
        Task & Criteria & Range & Group & K$\alpha$ & C$\alpha$ & ICC & UA \\
        \midrule
        \endfirsthead
        
        \toprule
        Task & Criteria & Range & Group & K$\alpha$ & C$\alpha$ & ICC & UA \\
        \midrule
        \endhead
        
        \midrule
        \multicolumn{7}{r}{Continued on next page}\\
        \endfoot

        \endlastfoot
\TaskEcoNUShort{} & filter & Boolean & formal & 0.680 & 0.873 & 0.945 & 0.963 \\
\TaskEcoNUShort{} & coherence & 0--3 & structure & 0.302 & 0.754 & 0.849 & 0.165 \\
\TaskEcoNUShort{} & argumentation & Boolean & domain & 0.173 & 0.492 & 0.661 & 0.468 \\
\TaskEcoNUShort{} & arguments\_count & 0--4 & structure & 0.311 & 0.782 & 0.861 & 0.174 \\
\TaskEcoNUShort{} & conclusion & Boolean & formal & 0.035 & 0.495 & 0.586 & 0.294 \\
\TaskEcoNUShort{} & reference & Boolean & formal & 0.648 & 0.853 & 0.925 & 0.945 \\
\TaskEcoNUShort{} & specific\_arguments & Boolean & domain & 0.404 & 0.691 & 0.766 & 0.615 \\
\TaskEcoNUShort{} & sum\_score & 0..1 & holistic & 0.653 & 0.875 & 0.897 &  \\
\midrule
\TaskEcoPSShort{} & filter & Boolean & formal & 0.147 & 0.350 & 0.935 & 0.931 \\
\TaskEcoPSShort{} & coherence & 0--3 & structure & 0.639 & 0.847 & 0.906 & 0.486 \\
\TaskEcoPSShort{} & arguments\_count & 0--4 & structure & 0.392 & 0.716 & 0.837 & 0.111 \\
\TaskEcoPSShort{} & specific\_arguments & Boolean & domain & 0.376 & 0.670 & 0.721 & 0.542 \\
\TaskEcoPSShort{} & conclusion & Boolean & formal & 0.321 & 0.668 & 0.698 & 0.514 \\
\TaskEcoPSShort{} & reference & Boolean & formal & 0.248 & 0.566 & 0.754 & 0.639 \\
\TaskEcoPSShort{} & sum\_score & 0..1 & holistic & 0.674 & 0.871 & 0.882 &  \\
\midrule
\TaskEcoSUShort{} & filter & Boolean & formal & 0.822 & 0.953 & 0.944 & 0.961 \\
\TaskEcoSUShort{} & coherence & 0--3 & structure & 0.374 & 0.805 & 0.841 & 0.079 \\
\TaskEcoSUShort{} & argumentation & Boolean & domain & 0.360 & 0.734 & 0.788 & 0.645 \\
\TaskEcoSUShort{} & arguments\_count & 0--4 & structure & 0.579 & 0.861 & 0.892 & 0.145 \\
\TaskEcoSUShort{} & specific\_arguments & Boolean & domain & 0.203 & 0.615 & 0.678 & 0.368 \\
\TaskEcoSUShort{} & conclusion & Boolean & formal & 0.535 & 0.839 & 0.839 & 0.592 \\
\TaskEcoSUShort{} & reference & Boolean & formal & 0.250 & 0.700 & 0.769 & 0.579 \\
\TaskEcoSUShort{} & sum\_score & 0..1 & holistic & 0.579 & 0.876 & 0.894 &  \\
\midrule
\TaskEcoWPShort{} & filter & Boolean & formal & 0.844 & 0.943 & 0.957 & 0.962 \\
\TaskEcoWPShort{} & coherence & 0--3 & structure & 0.486 & 0.779 & 0.838 & 0.238 \\
\TaskEcoWPShort{} & argumentation & Boolean & domain & 0.362 & 0.636 & 0.715 & 0.625 \\
\TaskEcoWPShort{} & arguments\_count & 0--4 & structure & 0.603 & 0.843 & 0.882 & 0.288 \\
\TaskEcoWPShort{} & specific\_arguments & Boolean & domain & 0.626 & 0.845 & 0.862 & 0.762 \\
\TaskEcoWPShort{} & conclusion & Boolean & formal & 0.417 & 0.702 & 0.746 & 0.588 \\
\TaskEcoWPShort{} & reference & Boolean & formal & 0.227 & 0.623 & 0.738 & 0.600 \\
\TaskEcoWPShort{} & sum\_score & 0..1 & holistic & 0.707 & 0.887 & 0.906 &  \\
        \bottomrule

    \caption{Inter-annotator agreements in the economics tasks among the two to four raters (see \autoref{tab:rater_statistics}) calculated using Krippendorff's $\alpha$ (K$\alpha$) with interval difference function, consistency using listwise Cronbach's $\alpha$ (C$\alpha$), ICC(3,k), and percentage of unanimous agreement (UA); see \autoref{sec:methods_llms}. Criteria, ranges and groups are further described in \autoref{sec:methods_data}. UA is not reported for \CritSumscore{}.}
    \label{tab:iaa_humans}
\end{longtable}

\subsection{Human Rater Agreement}
\label{sec:results_human_iaa}

We present the inter-annotator agreements of the human raters based on Krippendorff's alpha~\citep{Krippendorff:2011,Castro:2017} using the \textit{interval} difference function.
The results of the four economics tasks presented in \autoref{tab:iaa_humans} show, in some cases, considerable differences in the agreement for the same criteria across different tasks.
Good agreement, with alpha values over 0.8~\citep{Carletta:1996,Artstein:Poesio:2008} is only achieved for the \CritFilter{} criterion in case of the two economics tasks \TaskEcoWP{} and \TaskEcoSU{}.
For task \TaskEcoNU{}, the agreement is only tentatively acceptable, with a value of 0.68.
Similarly, the criterion \CritSumscore{} in tasks \TaskEcoWP{} and \TaskEcoPS{} are placed within the same agreement range.
The inter-annotator agreement for the other criteria can be considered unreliable, with values well below 0.5 in multiple cases.
This is particularly evident for \CritConclusion{} in task \TaskEcoNU{}, for which the alpha value is close to 0 and thus only slightly above the level of chance.

We report unanimous rating agreement as an additional metric.
In the event of highly skewed rating distributions, a low IAA may result from disagreement over a small number of instances of under-represented values, which can have a strong impact on the agreement score.
This is visible for the \CritFilter{} criterion in \TaskEcoPS{}:
here we see an unanimous rating in 93\% of the responses, but an IAA alpha of only 0.15.

The intraclass correlation coefficients ICC(3,k) range from 0.586 to 0.957, indicating mostly good to excellent inter-rater reliability~\citep{Shrout:Fleiss:1979,Cicchetti:Sparrow:1981,Vallat:2018}.
The ICC results generally align with the Krippendorff's alpha agreements.
Similarly, Cronbach's alpha (calculated using the \texttt{listwise} deletion method) estimates internal item consistency ranging from poor to excellent, with values from 0.35 to 0.953~\citep{Cronbach:1951,George:Mallery:2016,Vallat:2018}.

These measurements emphasise the challenges involved in rating the economics tasks based on the presented criteria.
In our analysis, we acknowledge that these human ratings differ and that no consolidated gold-standard annotation was provided as part of the original assessment data.
We therefore evaluate the LLM-based ratings in this multi-rater setting without assuming or constructing a single gold-standard annotation.
Instead, each human rating is treated as an equally plausible reference annotation, and the LLM is evaluated by the agreement between the individual raters in the rater group rather than against a single gold label.
Looking at the three groups of criteria, in the first group of \textit{formal} criteria, we notice extremely low human rater agreement on the \CritConclusion{} criterion in the \TaskEcoNU{} task.
In contrast, the other two criteria display the highest agreement values.
This suggests a difficulty or misunderstanding regarding what qualifies as a conclusion in the context of the task.
This appears to be a consequence of the ratings performed by rater B.
However, this seems to be specific to this task, as the other three tasks show agreement values comparable with the other criteria.
This task-difference is also evident for the \CritReference{} criterion, for which \TaskEcoNU{} behaves differently:
agreement is relatively high with 0.648, while the other three tasks display alpha values around 0.24.
In the second group, which contains the \textit{structure} criteria, no clear distinction in agreement is evident when comparing the different value ranges.
Despite their larger value range, the two non-Boolean criteria exhibit IAA scores in the mid-range.
The IAA among the human raters varies in third category comprising \textit{domain knowledge}, with the \CritArgumentation{} criterion being lower except for the \TaskEcoNU{} task.
This could suggest a problem with rater calibration.
Since this criterion accepts only \enquote{yes} or \enquote{no} as a rating, the raters might use different thresholds when assessing economic argumentation.
Similar to the \CritConclusion{} criterion, rater B displays different behaviour in task \TaskEcoNU{}; however, this time, the difference appears only in comparison with rater A and not with rater C.

In the pairwise Krippendorff's alpha inter-annotator agreements between the human raters we observe that the \CritSumscore{} results are stable despite the varying outcomes of the other criteria.
For the other criteria, we notice low agreements e.g., with rater B in task \TaskEcoNU{} or rater A in \TaskEcoSU{}.
This suggests that there are systematic differences in evaluation rather than random variation.
This finding shows that, although the raters disagree on multiple criteria, the composite score can compensate for differences in how the human raters understand, evaluate and apply the criteria.

\newpage
\begin{longtable}{l|l|l|l|r|r|r|r}
        \toprule
        Task & Criteria & Range & Group & K$\alpha$ & C$\alpha$ & ICC & UA \\
        \midrule
        \endfirsthead
        
        \toprule
        Task & Criteria & Range & Group & K$\alpha$ & C$\alpha$ & ICC & UA \\
        \midrule
        \endhead
        
        \midrule
        \multicolumn{6}{r}{Continued on next page}\\
        \endfoot

        \endlastfoot

\TaskMedATShort{} & filter & Boolean & formal & 0.481 & 0.738 & 0.891 & 0.935 \\
\TaskMedATShort{} & coherence & 0--3 & structure & 0.437 & 0.711 & 0.812 & 0.478 \\
\TaskMedATShort{} & cr\_a1 & Boolean & diagnoses & 0.614 & 0.831 & 0.844 & 0.717 \\
\TaskMedATShort{} & cr\_b1 & Boolean & diagnoses & 0.418 & 0.715 & 0.725 & 0.587 \\
\TaskMedATShort{} & cr\_b2 & Boolean & diagnoses & 0.851 & 0.946 & 0.947 & 0.957 \\
\TaskMedATShort{} & cr\_b3 & Boolean & diagnoses & 0.839 & 0.942 & 0.943 & 0.935 \\
\TaskMedATShort{} & cr\_b4 & Boolean & diagnoses & 0.255 & 0.672 & 0.678 & 0.500 \\
\TaskMedATShort{} & cr\_c1 & Boolean & treatments & 0.683 & 0.870 & 0.877 & 0.761 \\
\TaskMedATShort{} & cr\_c2 & Boolean & treatments & 0.684 & 0.873 & 0.891 & 0.826 \\
\TaskMedATShort{} & cr\_c3 & Boolean & treatments & 0.850 & 0.948 & 0.948 & 0.935 \\
\TaskMedATShort{} & cr\_c4 & Boolean & treatments & 0.763 & 0.906 & 0.907 & 0.935 \\
\TaskMedATShort{} & cr\_c5 & Boolean & treatments & 0.968 & 0.989 & 0.990 & 0.978 \\
\TaskMedATShort{} & cr\_c6 & Boolean & treatments & 0.916 & 0.971 & 0.971 & 0.978 \\
\TaskMedATShort{} & cr\_c7 & Boolean & treatments & 0.863 & 0.953 & 0.953 & 0.935 \\
\TaskMedATShort{} & cr\_d1 & Boolean & treatments & 0.077 & 0.580 & 0.582 & 0.435 \\
\TaskMedATShort{} & cr\_e1 & Boolean & treatments & 0.265 & 0.628 & 0.645 & 0.457 \\
\TaskMedATShort{} & sum\_score & 0..1 & holistic & 0.717 & 0.930 & 0.930 &  \\
\midrule
\TaskMedAUShort{} & filter & Boolean & formal & 0.506 & 0.804 & 0.934 & 0.865 \\
\TaskMedAUShort{} & coherence & 0--3 & structure & 0.232 & 0.703 & 0.864 & 0.096 \\
\TaskMedAUShort{} & cr\_a1 & Boolean & diagnoses & 0.698 & 0.902 & 0.910 & 0.731 \\
\TaskMedAUShort{} & cr\_b1 & Boolean & diagnoses & 0.323 & 0.695 & 0.787 & 0.442 \\
\TaskMedAUShort{} & cr\_b2 & Boolean & diagnoses & 0.287 & 0.674 & 0.683 & 0.635 \\
\TaskMedAUShort{} & cr\_b3 & Boolean & diagnoses & 0.220 & 0.645 & 0.728 & 0.327 \\
\TaskMedAUShort{} & cr\_b4 & Boolean & diagnoses & 0.047 & 0.168 & 0.186 & 0.788 \\
\TaskMedAUShort{} & cr\_c1 & Boolean & diagnoses & 0.515 & 0.833 & 0.838 & 0.673 \\
\TaskMedAUShort{} & cr\_c2 & Boolean & diagnoses & 0.385 & 0.734 & 0.738 & 0.808 \\
\TaskMedAUShort{} & cr\_c3 & Boolean & diagnoses & 0.384 & 0.730 & 0.737 & 0.712 \\
\TaskMedAUShort{} & cr\_c4 & Boolean & diagnoses & -0.025 & -0.077 & -0.068 & 0.885 \\
\TaskMedAUShort{} & cr\_d1 & Boolean & diagnoses & 0.523 & 0.818 & 0.821 & 0.846 \\
\TaskMedAUShort{} & sum\_score & 0..1 & holistic & 0.568 & 0.864 & 0.872 &  \\
\midrule
\TaskMedKLShort{} & filter & Boolean & formal & 0.894 & 0.962 & 0.980 & 0.981 \\
\TaskMedKLShort{} & coherence & 0-3 & structure & 0.370 & 0.733 & 0.839 & 0.423 \\
\TaskMedKLShort{} & cr\_a1 & Boolean & diagnoses & 0.354 & 0.721 & 0.746 & 0.519 \\
\TaskMedKLShort{} & cr\_b1 & Boolean & diagnoses & 0.679 & 0.879 & 0.880 & 0.865 \\
\TaskMedKLShort{} & cr\_b2 & Boolean & diagnoses & 1.000 & 1.000 & 1.000 & 1.000 \\
\TaskMedKLShort{} & cr\_b3 & Boolean & diagnoses & 0.497 & 0.750 & 0.750 & 0.981 \\
\TaskMedKLShort{} & cr\_b4 & Boolean & diagnoses & 0.922 & 0.973 & 0.973 & 0.981 \\
\TaskMedKLShort{} & cr\_b5 & Boolean & diagnoses & 1.000 & NaN & NaN & 1.000 \\
\TaskMedKLShort{} & cr\_b6 & Boolean & diagnoses & -0.006 & 0.000 & 0.000 & 0.962 \\
\TaskMedKLShort{} & cr\_b7 & Boolean & diagnoses & 1.000 & NaN & NaN & 1.000 \\
\TaskMedKLShort{} & cr\_c05 & Boolean & symptoms & -0.021 & 0.274 & 0.301 & 0.385 \\
\TaskMedKLShort{} & cr\_c1 & Boolean & symptoms & 0.462 & 0.755 & 0.776 & 0.596 \\
\TaskMedKLShort{} & cr\_d05 & Boolean & symptoms & 0.085 & 0.510 & 0.513 & 0.519 \\
\TaskMedKLShort{} & cr\_d1 & Boolean & symptoms & 0.316 & 0.677 & 0.693 & 0.500 \\
\TaskMedKLShort{} & cr\_e1 & Boolean & treatments & 0.731 & 0.892 & 0.892 & 0.942 \\
\TaskMedKLShort{} & cr\_e2 & Boolean & treatments & 0.903 & 0.966 & 0.967 & 0.942 \\
\TaskMedKLShort{} & cr\_e3 & Boolean & treatments & 0.911 & 0.968 & 0.969 & 0.962 \\
\TaskMedKLShort{} & cr\_e4 & Boolean & treatments & 0.930 & 0.976 & 0.977 & 0.962 \\
\TaskMedKLShort{} & cr\_e5 & Boolean & treatments & 0.589 & 0.817 & 0.817 & 0.962 \\
\TaskMedKLShort{} & cr\_e6 & Boolean & treatments & 0.851 & 0.946 & 0.946 & 0.981 \\
\TaskMedKLShort{} & sum\_score & 0..1 & holistic & 0.713 & 0.928 & 0.928 &  \\
        \midrule
\TaskMedMOShort{} & filter & Boolean & formal & 0.756 & 0.962 & 0.887 & 0.923 \\
\TaskMedMOShort{} & coherence & 0-3 & structure & 0.597 & 0.900 & 0.882 & 0.516 \\
\TaskMedMOShort{} & cr\_a1 & Boolean & diagnoses & 0.119 & 0.652 & 0.712 & 0.275 \\
\TaskMedMOShort{} & cr\_b1 & Boolean & diagnoses & 0.334 & 0.734 & 0.722 & 0.604 \\
\TaskMedMOShort{} & cr\_c1 & Boolean & diagnoses & 0.665 & 0.907 & 0.873 & 0.703 \\
\TaskMedMOShort{} & cr\_d1 & Boolean & diagnoses & 0.372 & 0.719 & 0.724 & 0.747 \\
\TaskMedMOShort{} & cr\_e1 & Boolean & treatments & 0.646 & 0.915 & 0.872 & 0.747 \\
\TaskMedMOShort{} & cr\_e2 & Boolean & treatments & 0.372 & 0.824 & 0.773 & 0.473 \\
\TaskMedMOShort{} & cr\_e3 & Boolean & treatments & 0.649 & 0.910 & 0.880 & 0.725 \\
\TaskMedMOShort{} & cr\_f1 & Boolean & treatments & 0.532 & 0.869 & 0.836 & 0.560 \\
\TaskMedMOShort{} & cr\_g1 & Boolean & symptoms & 0.859 & 0.971 & 0.941 & 0.912 \\
\TaskMedMOShort{} & cr\_g2 & Boolean & symptoms & 0.607 & 0.880 & 0.853 & 0.923 \\
\TaskMedMOShort{} & cr\_g3 & Boolean & symptoms & 0.873 & 0.959 & 0.937 & 0.989 \\
\TaskMedMOShort{} & cr\_g4 & Boolean & symptoms & 0.814 & 0.935 & 0.913 & 0.978 \\
\TaskMedMOShort{} & cr\_g5 & Boolean & symptoms & 0.773 & 0.934 & 0.908 & 0.978 \\
\TaskMedMOShort{} & cr\_g6 & Boolean & symptoms & 0.911 & 0.964 & 0.944 & 0.956 \\
\TaskMedMOShort{} & cr\_g7 & Boolean & symptoms & 0.710 & 0.934 & 0.900 & 0.978 \\
\TaskMedMOShort{} & cr\_g8 & Boolean & symptoms & 1.000 & NaN & NaN & 1.000 \\
\TaskMedMOShort{} & cr\_gx & Boolean & symptoms & 0.143 & 0.396 & 0.437 & 0.802 \\
\TaskMedMOShort{} & cr\_h1 & Boolean & symptoms & 0.029 & 0.221 & 0.189 & 0.901 \\
\TaskMedMOShort{} & sum\_score & 0..1 & holistic & 0.754 & 0.944 & 0.901 &  \\
        \bottomrule
        
    \caption{Inter-annotator agreements in the medicine tasks among the three to five raters (see \autoref{tab:rater_statistics}) calculated using Krippendorff's $\alpha$ (K$\alpha$) with interval difference function, consistency using listwise Cronbach's $\alpha$ (C$\alpha$), ICC(3,k), and percentage of unanimous agreement (UA); see \autoref{sec:methods_llms}. Criteria, ranges and groups are further described in \autoref{sec:methods_data}. UA is not reported for \CritSumscore{}.}
    \label{tab:iaa_humans_med}
\end{longtable}
Human raters tend to disagree on most of the economics criteria, which may suggest that they have different expectations regarding the students' statements.
The \CritSumscore{}, which is composed of the other criteria and can therefore partially compensate for differences and disagreements, shows comparatively high agreement across all four economics tasks.
While still below 0.80 in all cases, the agreement is often higher than that of the other individual criteria.
This implies that the human raters do not fully agree on how to assess students' responses on the criteria level, but they do agree more on a holistic assessment and can produce reliable overall performance ratings.
Additionally, as indicated by Cronbach's alpha, they mostly rank the criteria similarly.
The \CritSumscore{} may therefore be more suitable for assigning a holistic performance level to student responses.
Future work should study the direct generation of such a score by LLMs in comparison with a calculated \CritSumscore{}.

Compared with these results, the inter-annotator agreements for the medical tasks criteria show similar tendencies (see \autoref{tab:iaa_humans_med}).
We observe comparable \CritSumscore{} agreement values, which demonstrates the practicality of this holistic criterion across domains.
Similarly to the IAA patterns across the economics criteria, we do not see any systematic behaviour in agreement across the three groups of medical criteria.
The agreement for comparable criteria varies per task; overall, this pattern is similar across the two domains.

Together, the inter-annotator agreements across the eight tasks highlight the challenge of manual annotation when assessing student writing:
in both domains, ratings range from low to good, depending on the task and criterion, with a larger share of criteria reaching at least tentatively acceptable agreement in medicine (32 of 72) than in economics (5 of 31).
We use these results to set the baseline for model-based annotation, analysing to what extent LLMs can replace individual human raters while maintaining the level of agreement among the raters (RQ1).
In other words, given these divergent inter-annotator agreements, how effectively can LLMs replace one or two human raters in this rating task?
Maintaining the agreement does not necessarily mean that the assessment is correct:
if the LLM provided a more accurate rating than human raters, agreement would decrease.
However, our aim is to verify whether an LLM can replace a rater in this context based on these human results and similar instructions and information.

\subsection{LLMs Replacing Human Raters}
\label{sec:results_llm_replace}

\begin{figure}[t]
    \centering
    \begin{subfigure}{1\linewidth}
        \centering
    \includegraphics[trim=0 14cm 35cm 0,clip,width=1\linewidth]{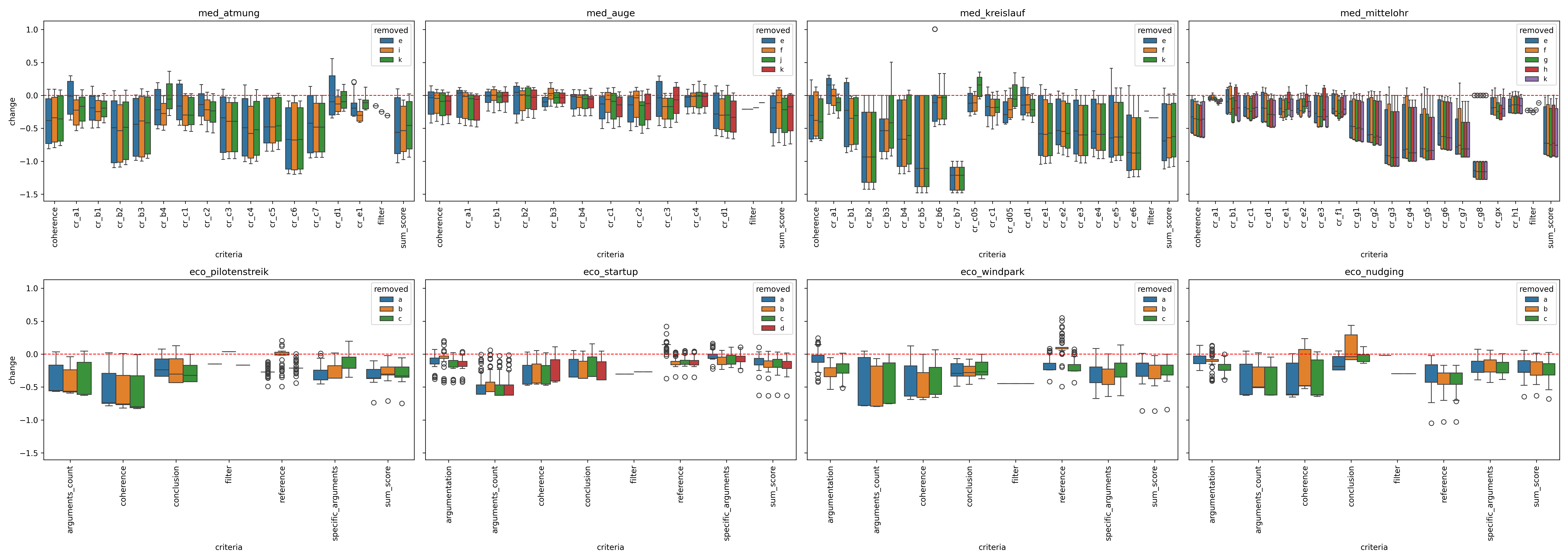}
    \includegraphics[trim=36.5cm 14cm 0 0,clip,width=1\linewidth]{changes_only_human_filterTrue_all_tasks.png}
    \caption{Medicine tasks}
    \end{subfigure}
\end{figure}
\begin{figure}[!ht]
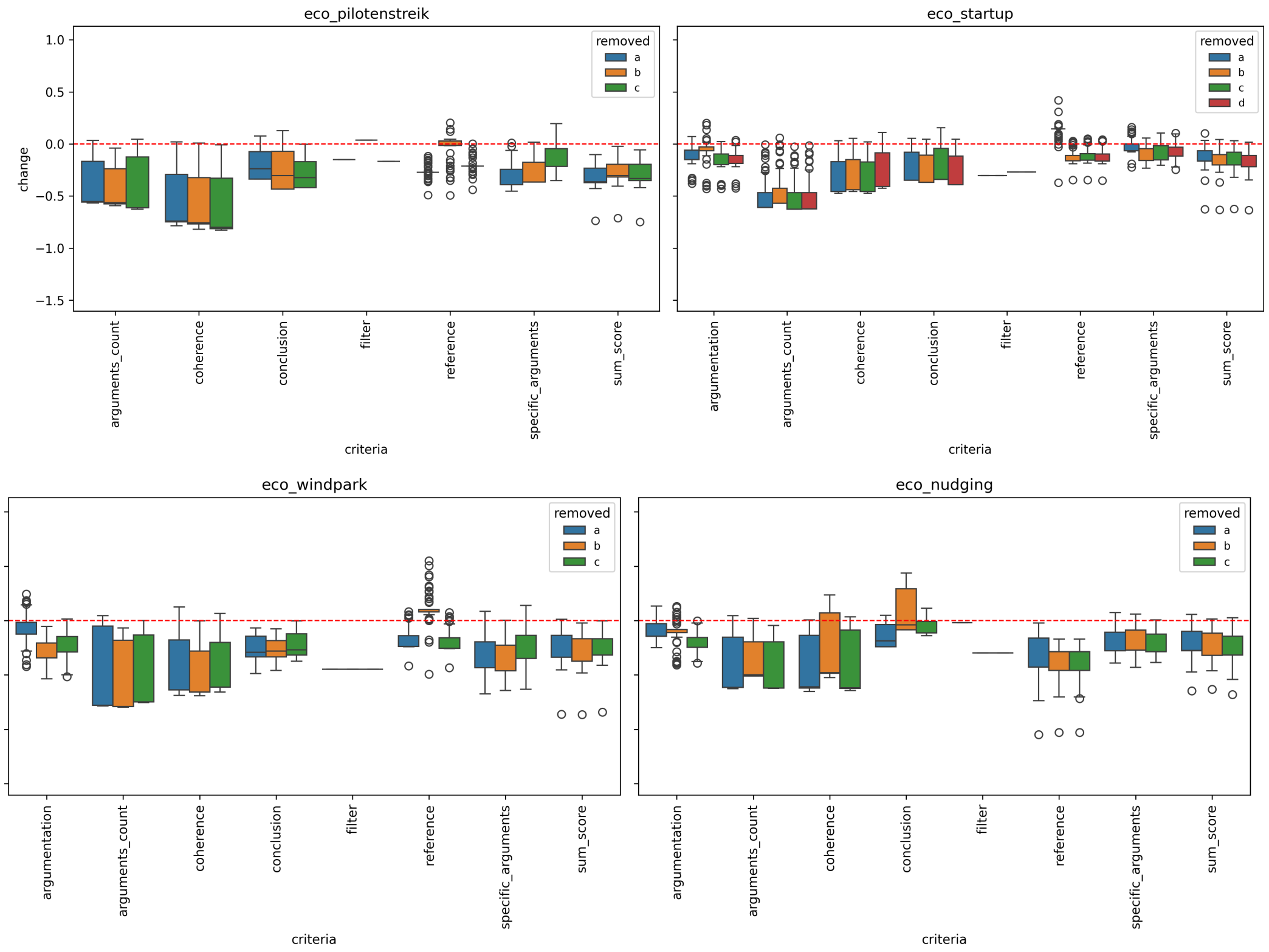

\ContinuedFloat
    \centering
    \begin{subfigure}{1\linewidth}
        \centering
    \includegraphics[trim=0 0 35cm 12cm,clip,width=1\linewidth]{changes_only_human_filterTrue_all_tasks.png}
    \includegraphics[trim=36.5cm 0 0 12cm,clip,width=1\linewidth]{changes_only_human_filterTrue_all_tasks.png}
    \caption{Economics tasks}
    \end{subfigure}
    \caption{Overview of changes in IAA when replacing a human annotator with an LLM in \ExpAll{}, per task (panel) and criterion (x-axis).
    Each box summarises all LLM--prompt configurations for one replaced human rater (distinguished by colour; the letters identify the human raters).
    Values above the dashed zero line indicate higher agreement of the rater group including the LLM than of the original human group.
    Panel titles use internal task names: \textit{med\_atmung} (\TaskMedAT{}), \textit{med\_auge} (\TaskMedAU{}), \textit{med\_kreislauf} (\TaskMedKL{}), \textit{med\_mittelohr} (\TaskMedMO{}), \textit{eco\_pilotenstreik} (\TaskEcoPS{}), \textit{eco\_startup} (\TaskEcoSU{}), \textit{eco\_windpark} (\TaskEcoWP{}), \textit{eco\_nudging} (\TaskEcoNU{}).}
    \label{fig:results_iaa_change_per_human_all_tasks}
\end{figure}

\begin{figure}[t]
    \centering
    \includegraphics[trim=0 0 52cm 0,clip,width=.6\linewidth]{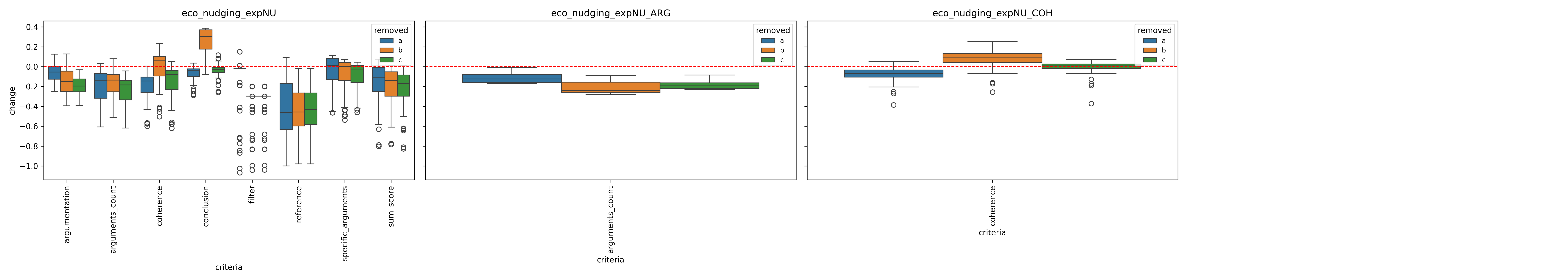}
    \includegraphics[trim=19cm 0 17cm 0,clip,width=1\linewidth]{changes_only_human_filterTrue_all_tasks_expNU.png}
    \caption{Overview of changes in IAA when replacing a human annotator with an LLM in \ExpNu{}, for prompts rating all criteria (\textit{eco\_nudging\_expNU}) and prompts focusing only on \CritArgumentsCount{} (\textit{ARG}) or \CritCoherence{} (\textit{COH}).
    Box elements as in \autoref{fig:results_iaa_change_per_human_all_tasks}.}
    \label{fig:results_iaa_change_per_human_nu}
\end{figure}

We demonstrate the effects of replacing one of the human raters with an LLM-generated rating for the 103 criteria across all eight tasks covered in \ExpAll{} in \autoref{fig:results_iaa_change_per_human_all_tasks}.
The boxplots show the change in inter-annotator agreement among the LLM--human rater groups after each human rater has been replaced by each LLM one by one.
While the overall average across all tested combinations is predominantly negative, we observe improvements in IAA for at least one LLM--human combination for most criteria, except for the \CritFilter{} criterion.
There, we do not see any variance:
the prompts and models behave very similarly and achieve a positive change only for the \TaskEcoPS{} task.
However as mentioned in \autoref{sec:results_human_iaa}, the \CritFilter{} criterion displays a highly skewed distribution towards the \enquote{yes} option, and small deviations in the minority option result in higher disagreement.
We notice an improvement for the \CritReference{} criterion in the three tasks except \TaskEcoNU{}, which confirms that it is more difficult to achieve improvements in agreement when the human raters already display high levels of agreement.
Compared to the economics tasks, the medicine tasks show a wider distribution of agreement changes for most criteria.
These findings show that we can build LLM--prompt configurations that can replace one human rater for most criteria under our agreement-based evaluation.
In total, this is the case for 86.4\% of the criteria (89 of 103).
These criteria show increased agreement among human raters and LLMs in at least one case when one human rater is replaced with an LLM.

The results of the second experiment, \ExpNu{}, are shown in \autoref{fig:results_iaa_change_per_human_nu}.
Generally, we observe similar distributions across the different human raters, which is expected, since the prompts in both experiments are based on similar variants.
However, whereas six of the eight criteria displayed a positive change in inter-annotator agreement, we now see an improvement in all eight criteria of the \TaskEcoNU{} task, including the \CritFilter{}.
Based on the results of this experiment, we are able to replace a human rater for 91 of the 103 criteria (88.3\%).
We contrast these results with the \textit{Alternative Annotator Test} in the next section.

The prompts considered so far work by asking the model for all criteria in a single request or through chained messages.
%
As part of \ExpNu{}, we additionally examine LLM and prompt configurations that focus on \CritCoherence{} and \CritArgumentsCount{}.
Here, we observe no positive change in IAA for the prompts concerning the \CritArgumentsCount{} criterion.
Compared to the non-specialised prompts, the agreement alpha increases slightly to a new maximum value for the \CritCoherence{} criterion.
Based on this test, it cannot be inferred whether the LLMs are able to produce more agreeable ratings when prompted to consider all criteria simultaneously rather than to separately.

Tables~\ref{tab:results_iaa_change_per_human_all_tasks_eco_ps} to~\ref{tab:results_iaa_change_per_human_all_tasks_med_mo} in the appendix provide an overview of the top IAA values and models per task.
The tables show the highest IAA score as well as the change per criterion when replacing a human rater with an LLM.
The list is sorted by the top IAA score and subsequently by the largest positive IAA change.
The count (\texttt{\#}) indicates the number of models achieving the same IAA value; the models are sorted alphabetically.
For economics, we report prompts from four of the five categories as the top-performing prompts per criterion in \ExpAll{}, only the \texttt{OPT} category is missing.
A \textit{PE2}-optimised prompt is top-performing in \ExpNu{} for the \CritSpecificArguments{} criterion, and a few-shot prompt outperforms the prompts for the \CritCoherence{} criterion.
We do not observe the economics-fine-tuned LLM Plutus among the actual top results for the economics criteria, with the exception of \CritFilter{}; however, in this case, the top position is shared with about 50 other configurations, and the change is negative.
Plutus is also shown as top-performing for the medical \TaskMedKL{} task on \CritMed{B7}, this criterion displays a drop from perfect agreement among the human raters to an IAA of 0, and this top position is also shared with three other configurations.
The medical LLM similarly appears only in one negative case.
These findings suggest that, in the context of this study, using more specialised LLMs offers no apparent advantage.

In total, 14.6\% (2992 of 20464) of the tested configurations per criterion show no decrease in IAA, 70\% of which concern the medical criteria.
Relative to each domain, 12\% of the economics samples and 16\% of the medical samples pass the human replacement test.
The distribution is similar to that of the total passing criteria, with \TaskEcoNU{} in first place (55.87\%), where we can replace a human rater with an LLM for all criteria, followed by \TaskEcoSU{} (19.53\% pass, replacement for 87.5\% of criteria), \TaskEcoPS{} (13.32\%; 85.7\%) and \TaskEcoWP{} (11.29\%; 75\%).
However, the results for the \TaskEcoNU{} task may be skewed because of the additional tests in \ExpNu{}.
This pattern is replicated for the medical tasks, where we report \TaskMedAU{} (36.70\%), \TaskMedMO{} (25.31\%), \TaskMedKL{} (22.41\%) and \TaskMedAT{} (15.57\%).
These results further confirm a certain degree of task-specific difficulty, which corresponds to the mean human IAA per task: lower human agreement can be matched more easily by the LLMs.

We most frequently report Llama 3.3 (70b) among the top-performing economics configurations in \ExpAll{}, with 17 occurrences, followed by Nemotron (70b) with 4 occurrences.
The most frequent prompt classes are \texttt{TRAINING} (10), \texttt{CHAIN} (9), \texttt{SIMPLE} (6) and \texttt{FEATURES} (4).
GPT-OSS (120b) achieves top performance for 14 medical criteria.
However, as in economics, Llama 3.3 (70b) performs best overall, with 31 occurrences, with Nemotron (70b) in third place with 10 occurrences.
For the medicine tasks, the distribution of the best prompts shifts to the \texttt{CHAIN} group (37), followed by \texttt{OPT} (24) and \texttt{FEATURES} (7).
Looking at the top performances, we see a variety of LLM families and prompt groups, but with a strong focus on one LLM in both domains and different prompt groups per domain. 

\begin{figure}
    \centering
    \includegraphics[trim=0 0 10cm 0,angle=90,origin=c,height=.9\textheight]{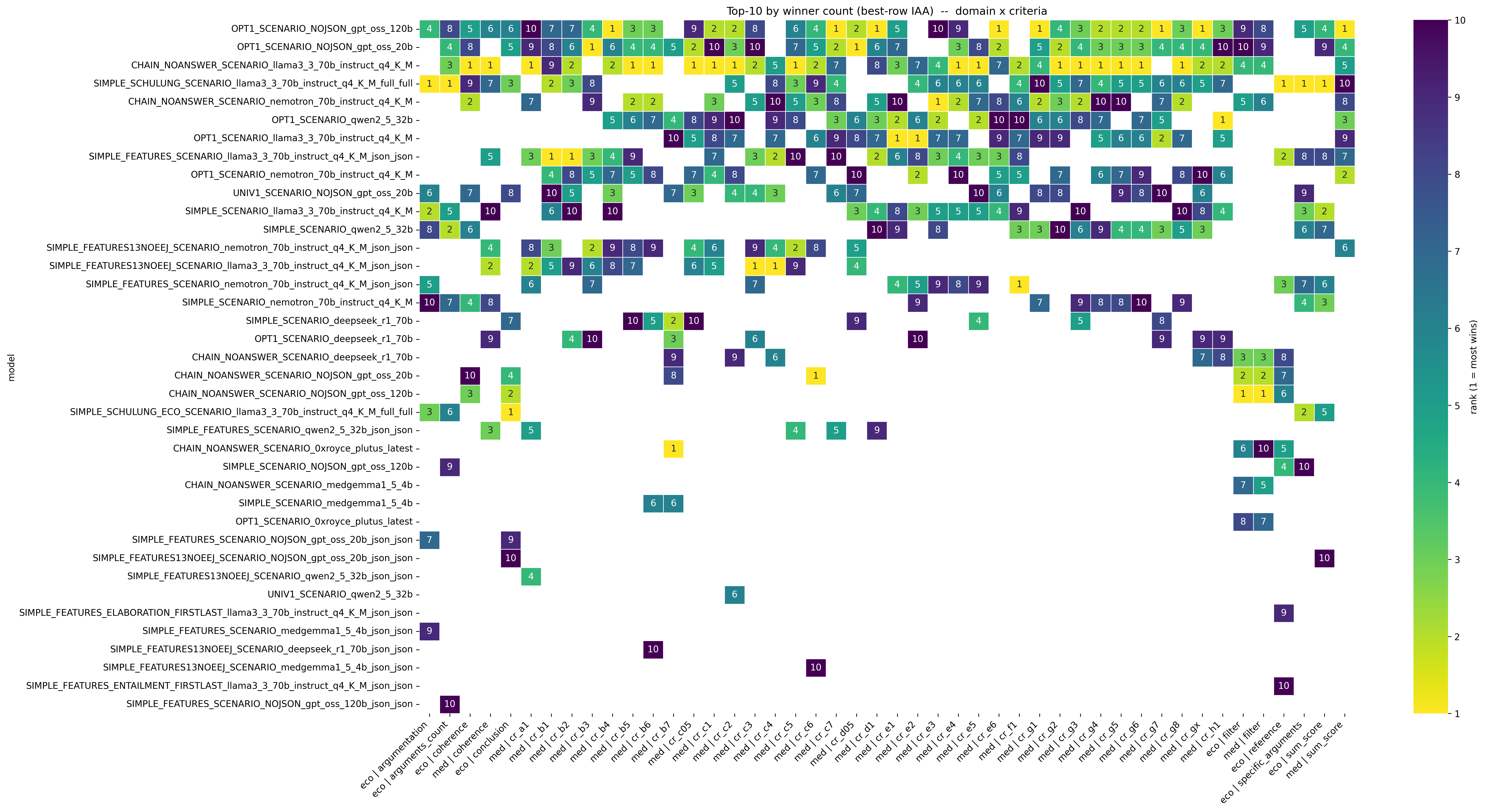}
    \caption{%
    Heatmap showing model overlap among the top 10 models per domain and criterion.}
    \label{fig:results_model_overlap}
\end{figure}

The results so far consider only the top-performing configurations.
We show the ranking of the top 10 LLM--prompt configurations per domain and criterion in \autoref{fig:results_model_overlap}, and their Jaccard similarity in the appendix in \autoref{fig:results_model_overlap_jaccard}.
We report an average of similarity of 0.36 and, for \CritFilter{}, full overlap.
Two clusters can be observed, generally containing only medical and economics criteria, respectively.
Overall, we see the GPT-OSS models of both sizes perform best, using the prompt optimized by Opus -- this is however skewed by the larger number of medicine criteria.
Ranking the prompts by domain separately, we see \texttt{TRAINING} prompts rank highly for the economics criteria, and \texttt{CHAIN} prompts rank highly for medicine.
The larger number of medical criteria could benefit from more fine-grained, criterion-by-criterion rating generation.
However, as with economics, we see instances of all prompt classes among the top performing configurations.
The Jaccard similarities are slightly lower for economics (0.31) and higher for medicine (0.42).
The highest overlap is between the two criteria \CritArgumentation{} and \CritSpecificArguments{}, both placed in the \textit{domain knowledge} group.
These results suggest that the model does not require specific training per domain or task, but that the prompt is sensitive to both, criteria and task.

\subsection{Alternative Annotator Test}
\label{sec:results_alt_test}

\begin{sidewaysfigure}
    \centering
    \makebox[\linewidth][c]{%
    \includegraphics[trim=0 0 8cm 0,width=1.1\textheight]{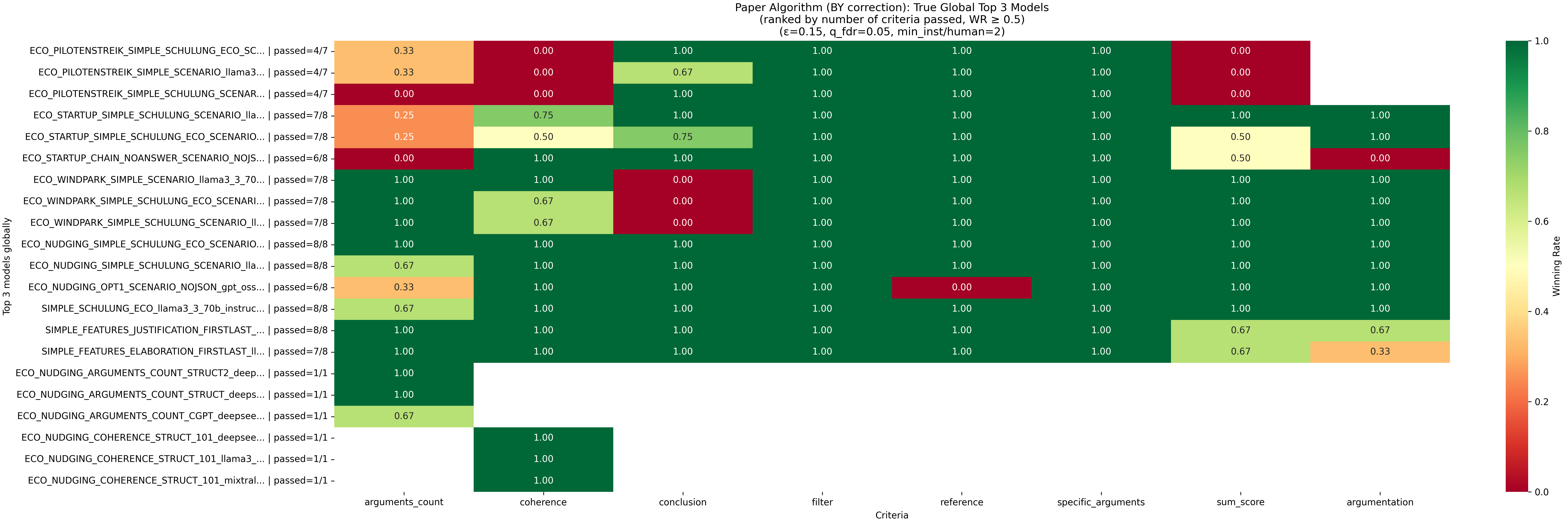}
    }
    \caption{Winning rates of the \textit{Alternative Annotator Test} using a cost-benefit parameter of 0.15 to denote skilled annotators, showing the top three LLM--prompt combinations per task in the economics tasks, ranked by the number of passed criteria.
    Columns show the criteria and rows the LLM--prompt configurations; empty cells denote criteria not rated by the configuration.
    Row labels show internal identifiers (task, prompt, and model; truncated) followed by the number of passing criteria; SCHULUNG denotes TRAINING prompts.
    For \TaskEcoNU{}, the rows include the top three configurations each of \ExpAll{}, \ExpNu{}, and the prompts focusing on \CritArgumentsCount{} and \CritCoherence{}.
    }
    \label{fig:alt_test_eco}
\end{sidewaysfigure}

\begin{sidewaysfigure}
    \centering    
    \makebox[\linewidth][c]{%
    \includegraphics[trim=0 0 8cm 0,width=1.1\textheight]{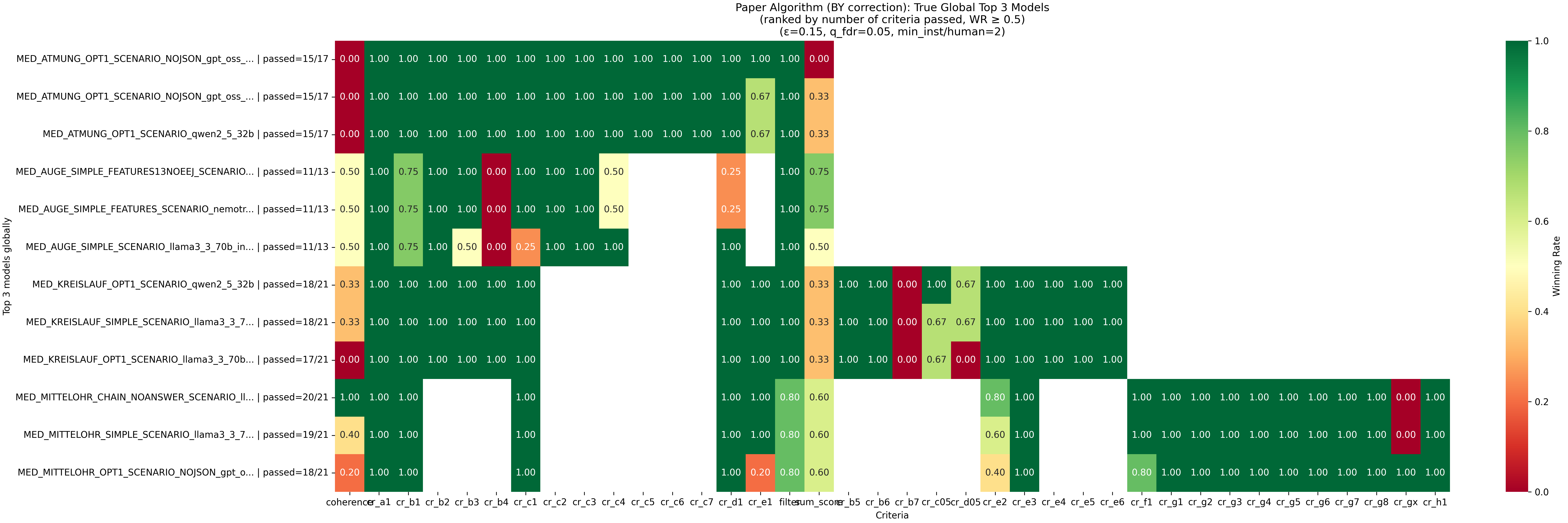}
    }
    \caption{Winning rates of the \textit{Alternative Annotator Test} using a cost-benefit parameter of 0.15 to denote skilled annotators, showing the top three LLM--prompt combinations per task in the medicine tasks, ranked by the number of passed criteria.
    Columns show the criteria and rows the LLM--prompt configurations; empty cells denote criteria not rated by the configuration.
    Row labels show internal identifiers (task, prompt, and model; truncated) followed by the number of passing criteria; SCHULUNG denotes TRAINING prompts.}
    \label{fig:alt_test_med}
\end{sidewaysfigure}

Our results so far show that we are able to replace a human rater without lowering the agreement for about 88\% of the criteria.
This was evaluated by examining whether any LLM--prompt combination can replace any human rater.
In other words, we consider only peak performance to determine the upper bound of what is possible with the studied configurations.
We confirm our findings by using a second metric in addition to inter-annotator agreement:
We present the results of the \textit{Alternative Annotator Test} (alt-test)~\citep{Calderon:et:al:2025} to determine which LLMs and prompts can replace human raters.
See Figures~\ref{fig:alt_test_eco} (economics) and~\ref{fig:alt_test_med} (medicine) for an overview of the results of the alt-test, showing the top three models per task.

In contrast to the inter-annotator metric, we now examine the top LLM--prompt combinations based on the number of criteria for which they pass the alt-test.
In total, we report that 87.4\% of the criteria (90 of 103) can be graded using an LLM-based automatic rating according to the alt-test.
This shows that at least one LLM--prompt combination can replace at least 50\% of the human annotators for a given criterion.

We show the effect of the cost-benefit parameter in \autoref{fig:alt_test_sensitivity} in the appendix.
Even when considering a value of 0.1, which \citet{Calderon:et:al:2025} recommended for unskilled crowdworkers, we still see most criteria passing.

In total, we report that 41.87\%, i.e., \numprint{2423} of the \numprint{5787} evaluated combinations, pass the alt-test.
The passing prompt classes behave differently compared with the IAA-change metric:
we now see prompts from the \texttt{OPT} and \texttt{FEATURES} groups in first and second position with 33\% and 24\%, respectively, both overall and per domain.
Considering the configurations ranked by number of passing criteria, we again find the \texttt{TRAINING} prompt group at the top for economics, compared with \texttt{OPT}, \texttt{FEATURES} and \texttt{CHAIN} prompts for medicine.
This is more in line with our findings using IAA change as a metric and again highlights the need for domain-specific prompts.
The results of the alt-test suggest the same conclusion as the IAA-change results.

\subsection{Replacing Two Human Raters}
\label{sec:results_replacing_two}

\begin{figure}[th!]
    \centering
    \makebox[\linewidth][c]{%
    \includegraphics[trim=0 52.5cm 0 0,clip,width=1.3\linewidth]{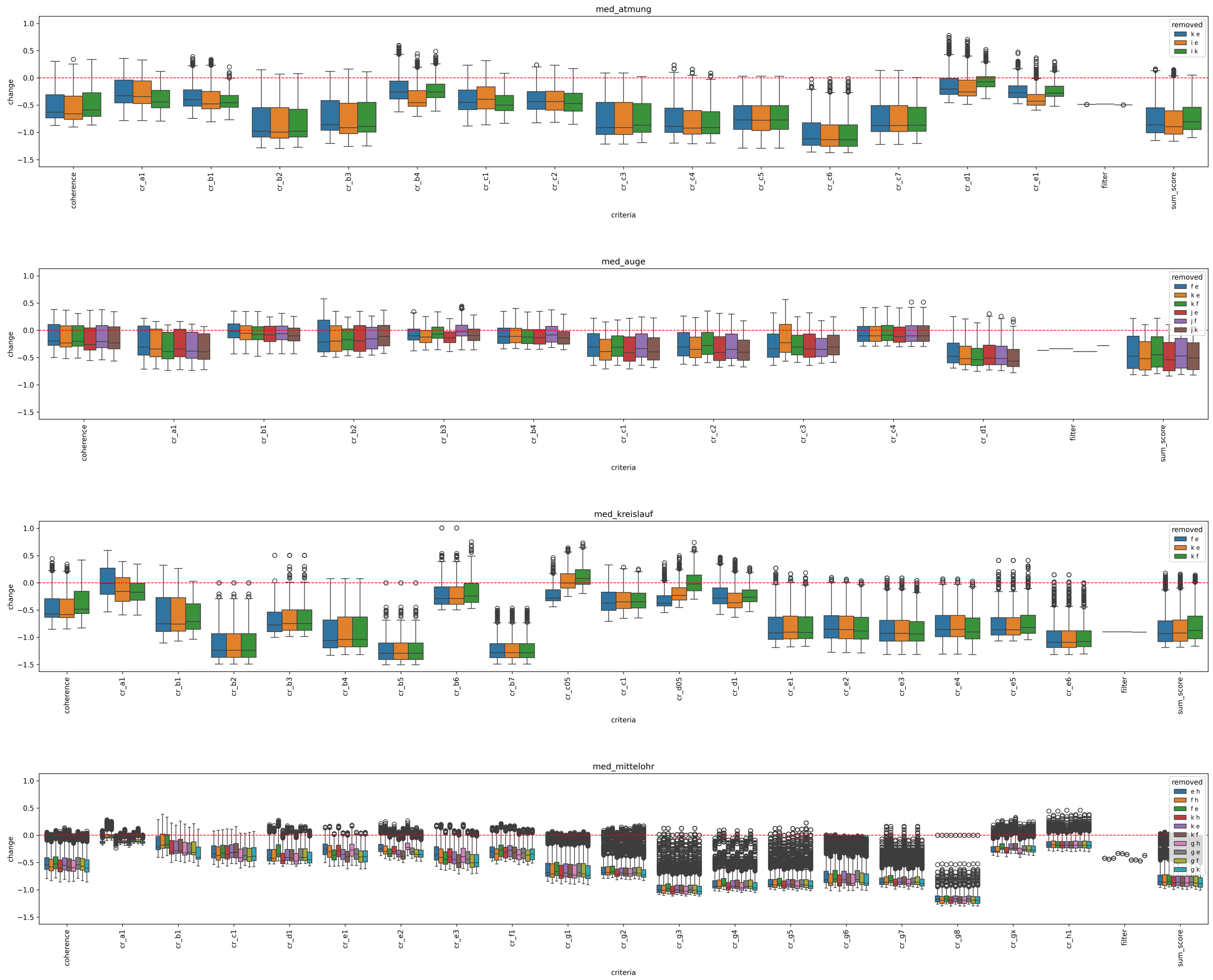}
    }
    \caption{Inter-annotator agreement when replacing two human raters with all LLM--prompt combinations in the medical tasks.}
    \label{fig:humans_minus_two_med}
\end{figure}
\begin{figure}[th!]
    \centering
    \makebox[\linewidth][c]{%
    \includegraphics[trim=0 0 0 50cm,clip,width=1.3\linewidth]{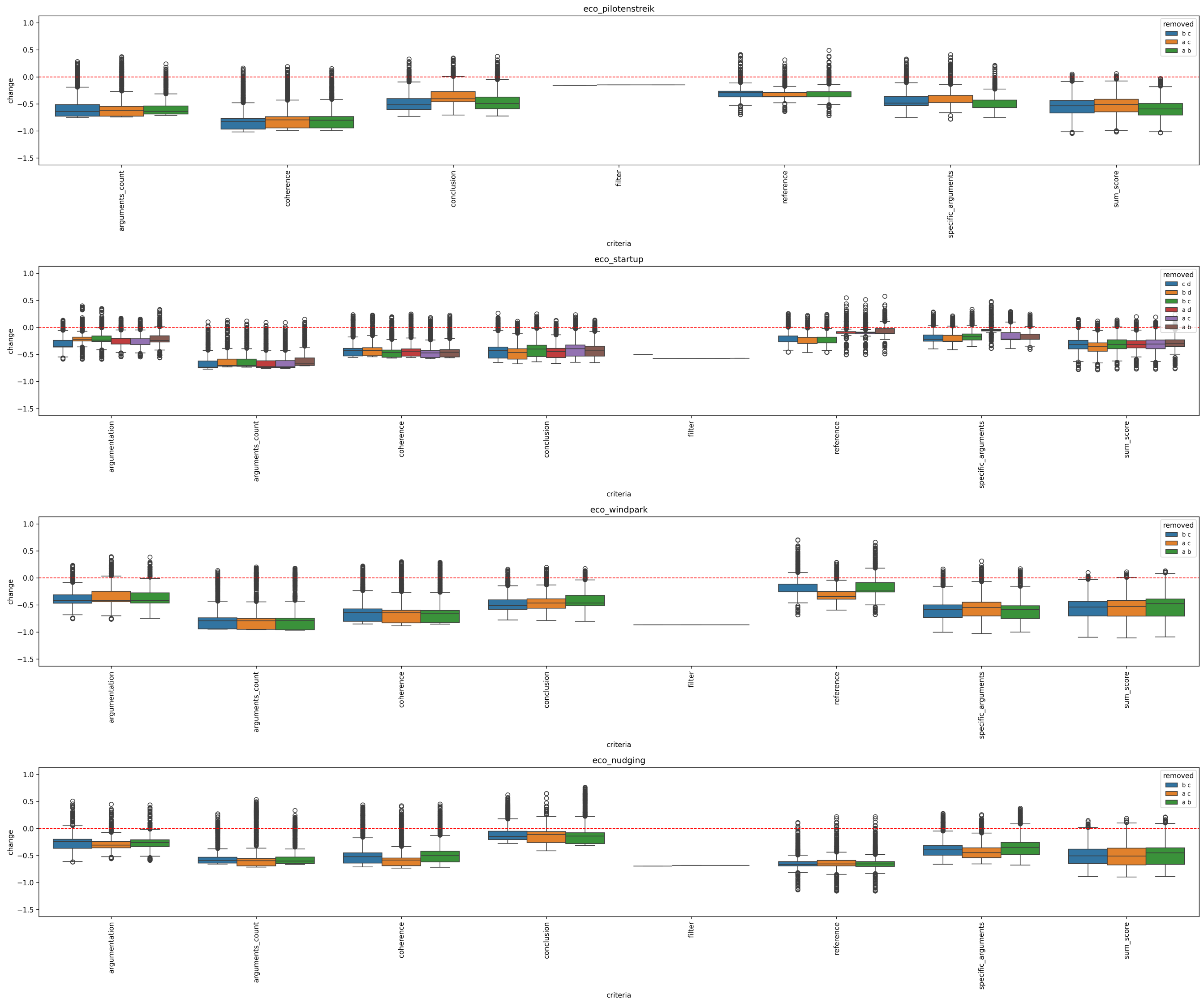}
    }
    \caption{Inter-annotator agreement when replacing two human raters with all LLM--prompt combinations in the economy tasks.}
    \label{fig:humans_minus_two_eco}
\end{figure}

We have demonstrated that an LLM can replace one of the human raters across all eight tasks.
We further assess the extent to which LLMs can replace two raters in \ExpAll{}.
Figures~\ref{fig:humans_minus_two_med} (medicine) and~\ref{fig:humans_minus_two_eco} (economics) show the resulting inter-annotator agreement for all eight tasks.

As expected from existing research, there are fewer differences in the agreement distributions than when only one human rater is replaced.
On average, the top agreement changes are higher for each task, the LLMs tend to agree more among themselves.
In \ExpAll{}, we observe an improvement of four additional passing criteria, totaling 90.3\% of the criteria (93 of 103) with a non-negative IAA change.
Two of these four criteria come from the medical task \TaskMedMO{} and one each from economics tasks \TaskEcoWP{} and \TaskEcoNU{}.
The medical task has the highest average number of human raters per student response, while both economics tasks have the lowest, thus leaving only one human rater in the agreement calculation.
However, for both economics tasks, two human raters cannot be replaced for all criteria.
Overall, these results are therefore quite stable.
This suggests that, even in a setting such as the one studied here, a reduced number of human raters may suffice to ground the LLM-based approaches.

\subsection{LLM-based Rating}
\label{sec:results_iaa_alt}

\begin{figure}[t]
    \centering
    \includegraphics[width=1\linewidth]{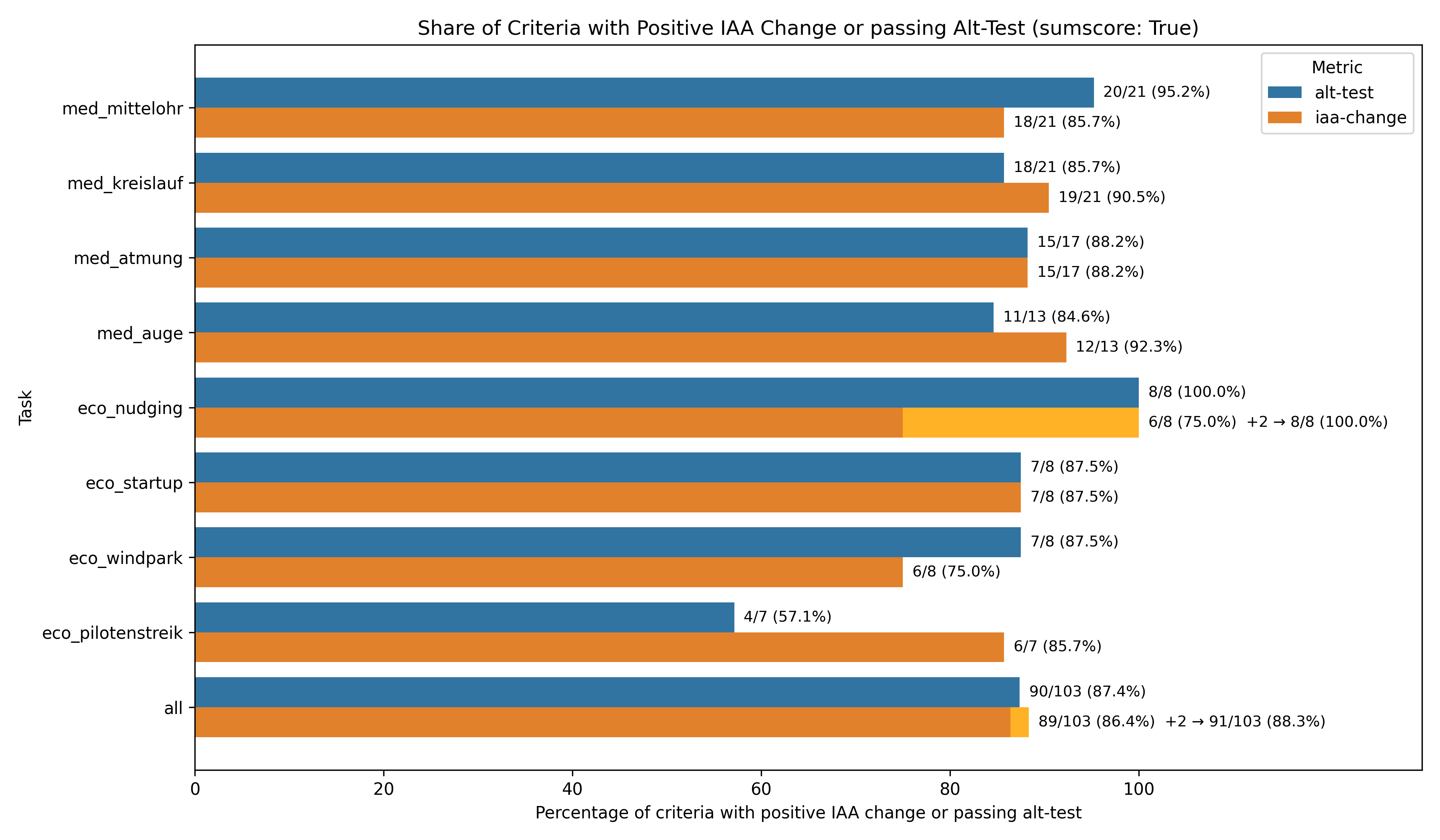}
    \caption{Results of the IAA-change and alt-test analyses for the 103 criteria considered across all eight tasks.}
    \label{fig:results_criteria_both}
\end{figure}

\begin{figure}[t]
    \centering
    \includegraphics[width=1\linewidth]{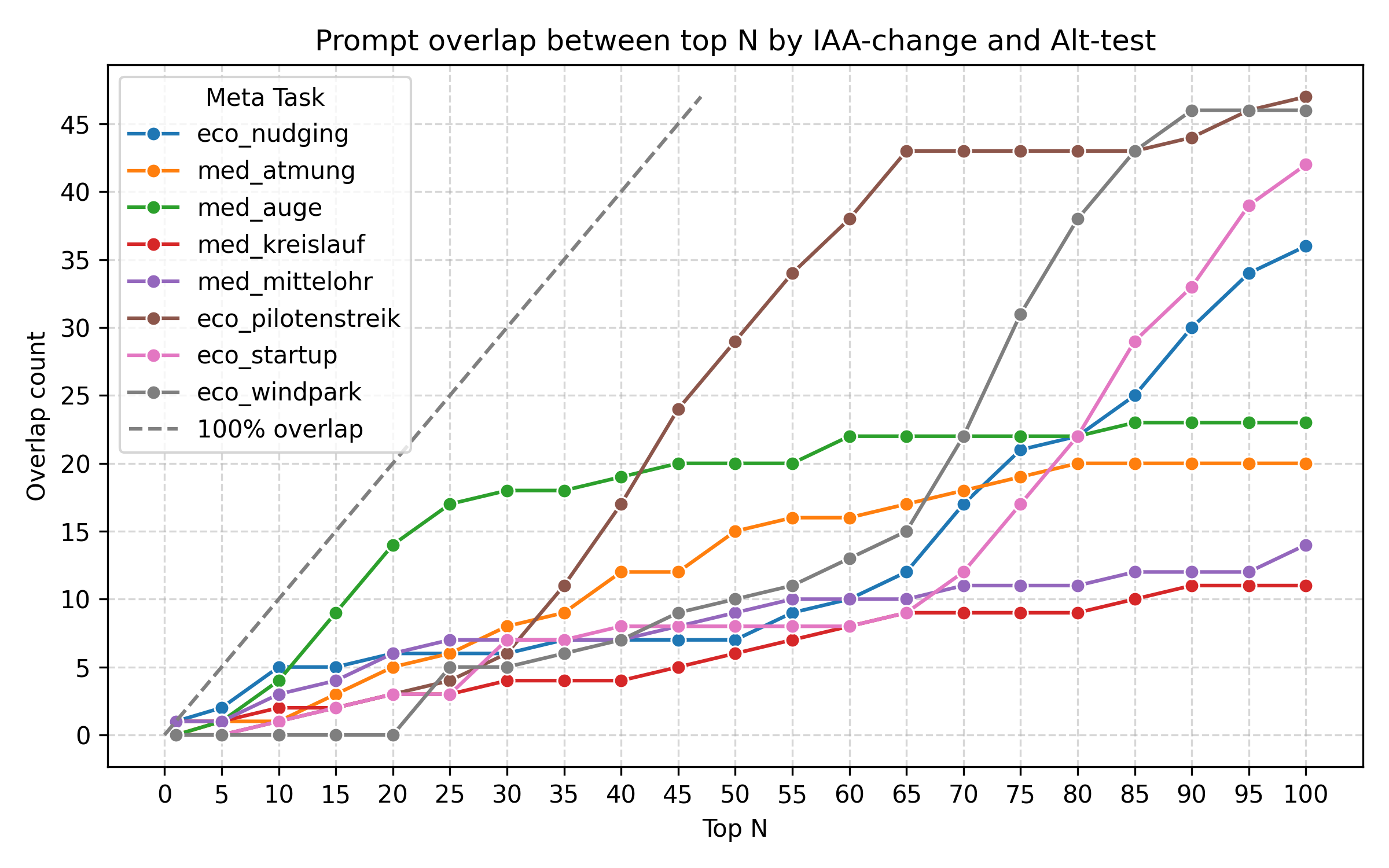}
    \caption{%
    LLM--prompt configuration overlap when comparing the top $N$ performers according to IAA-change and alt-test.}
    \label{fig:iaa_alt_overlap}
\end{figure}

The two metrics presented in \autoref{sec:results_llm_replace} and \autoref{sec:results_alt_test} are summarised in \autoref{fig:results_criteria_both}.
The improvements in the \TaskEcoNU{} task over the results of \ExpAll{} are shown in yellow.
As the \textit{Alternative Annotator Test} already shows all criteria passing for this task, this applies only to the inter-annotator agreement metric.
Together, these results address research questions \textit{RQ1} and \textit{RQ2}:
both metrics point in the same direction and indicate that, for most criteria, at least one LLM--prompt configuration can replace one or even two human raters without lowering the agreement (RQ1); this holds for all eight tasks and both domains, although the share of criteria varies by task and the best configuration differ by task and criterion (RQ2).
We show the LLM--prompt configuration overlap, considering the top $N$ best performing prompts as determined by the two metrics per task, in \autoref{fig:iaa_alt_overlap}.
Considering only the top configuration, we report an overlap with a \texttt{CHAIN} prompt for the two medicine tasks \TaskMedKL{} and \TaskMedMO{} (Llama 3.3 70b in both cases) and task \TaskEcoNU{} (Gemma 2 27b).
At the top 10, we see an average overlap of 21\% across all tasks except \TaskEcoWP{}, for which overlap appears only at the top 25.
At top 5, only economy tasks \TaskEcoWP{} and \TaskEcoSU{} show no overlap.
These findings support the feasibility of the LLM-rater approach across the eight tasks. 

We now investigate the influence of the chosen LLMs and prompts using a \textit{SHAP} analysis.

\subsection{SHAP Results}
\label{sec:results_shap}

\begin{figure}
    \centering
    \begin{subfigure}{1\linewidth}
    \makebox[\linewidth][c]{%
    \includegraphics[trim=0 0 0 7.5cm,width=1\linewidth]{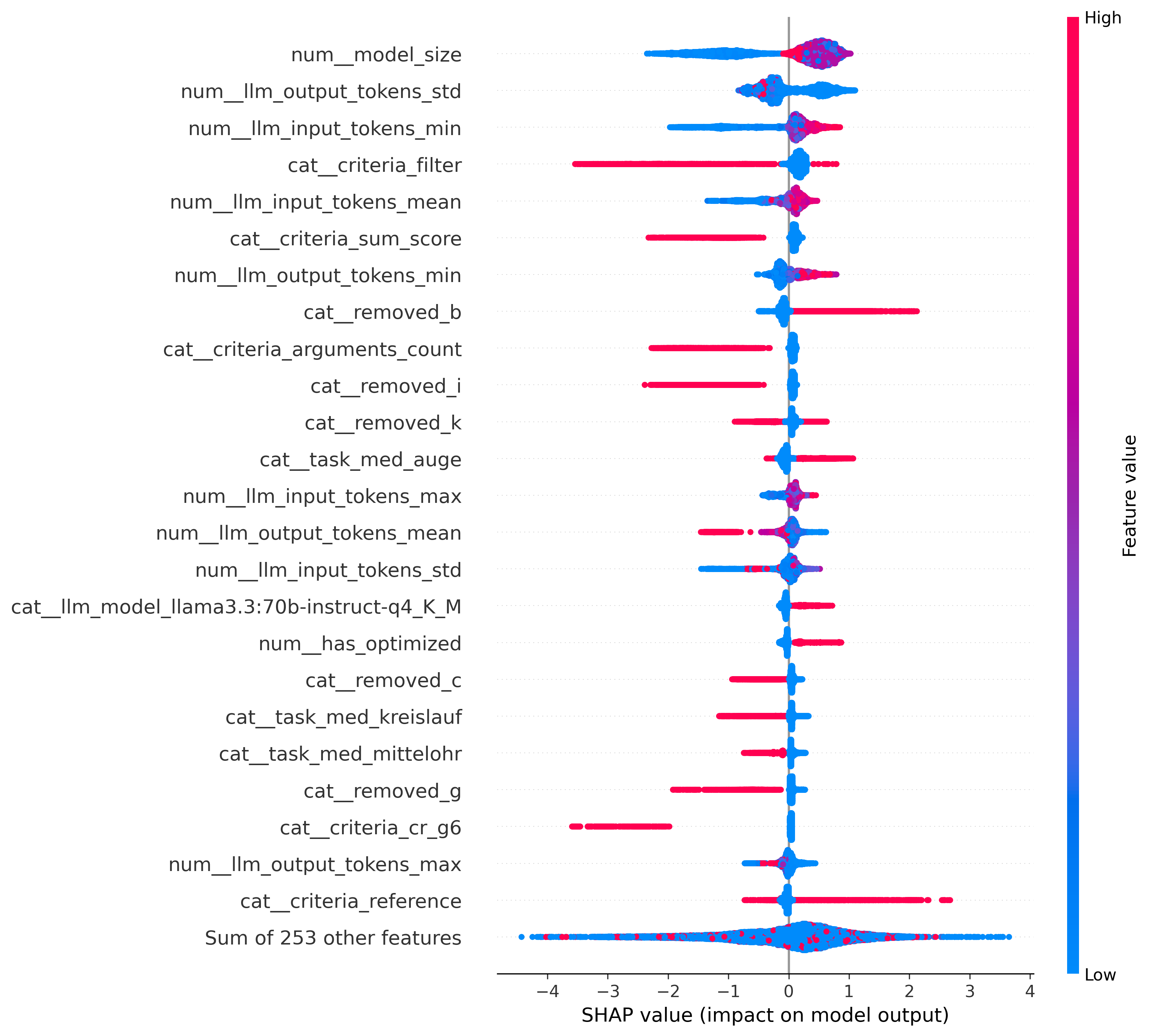}
    }
    \caption{Classifier}
    \end{subfigure}
    \vfill
    \begin{subfigure}{1\linewidth}
    \centering
    \makebox[\linewidth][c]{%
    \includegraphics[width=1\linewidth]{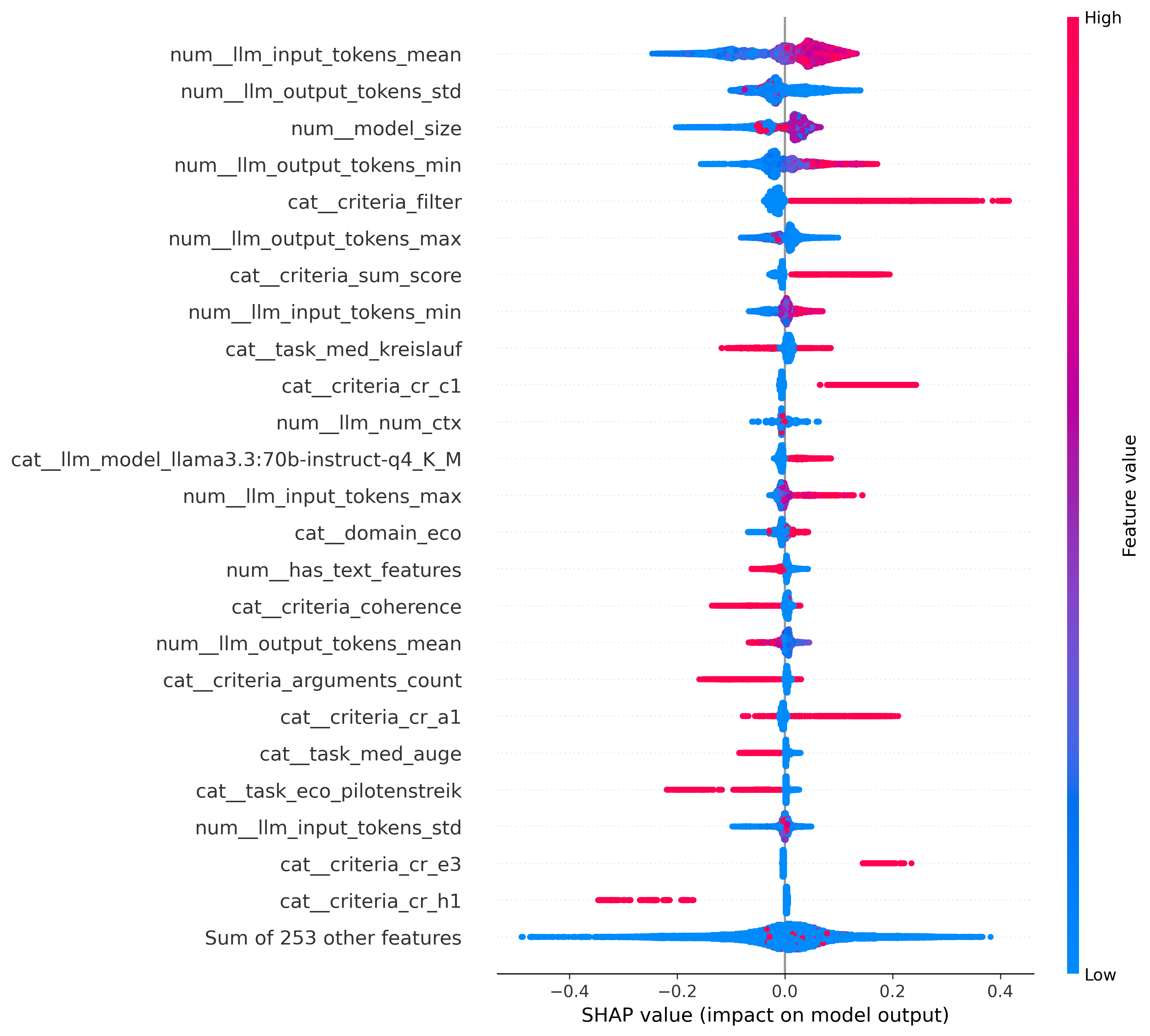}
    }
    \caption{Regression}
    \end{subfigure}
    \caption{Beeswarm plots showing feature importance based on the \textit{SHAP} analysis.}
    \label{fig:results_shap_beeswarm}
\end{figure}

We perform a \textit{SHAP} analysis to determine the feature importance of the LLM--prompt configurations by using two types of proxy models.
The trained proxy models achieve a performance of $R^2 = 0.73$ for regression and $F_1 = 0.79$ for binary classification.
For comparison, the results of a baseline model and two SVM variants are shown in \autoref{tab:shap_models} in the appendix.
The \textit{SHAP} analysis attributes the greatest influence to features related to the LLM parameter size and the number of input and output tokens, as shown in the beeswarm plots in \autoref{fig:results_shap_beeswarm}.
That is, larger models tend to result in better replaceability and IAA prediction, with a similar pattern observed for higher numbers of input tokens.
The more variable the number of output tokens is, the lower the performance.
This may indicate problems with prompt stability, as the output should not vary substantially with the student response given.
The Llama 3.3 (70b) model appears in the list of top features with a positive influence on both the IAA value and the IAA change.
This is supported by our analysis of the LLMs, in which this model was the best-performing model most often.
The \CritFilter{} criterion shows a mirrored behaviour in the regression and classification models.
This can be explained by the fact that we are able to correctly predict the IAA value, which, as shown in our analysis, mostly does not lead to a positive change.

\begin{figure}[t]
    \centering
    \includegraphics[width=1\linewidth]{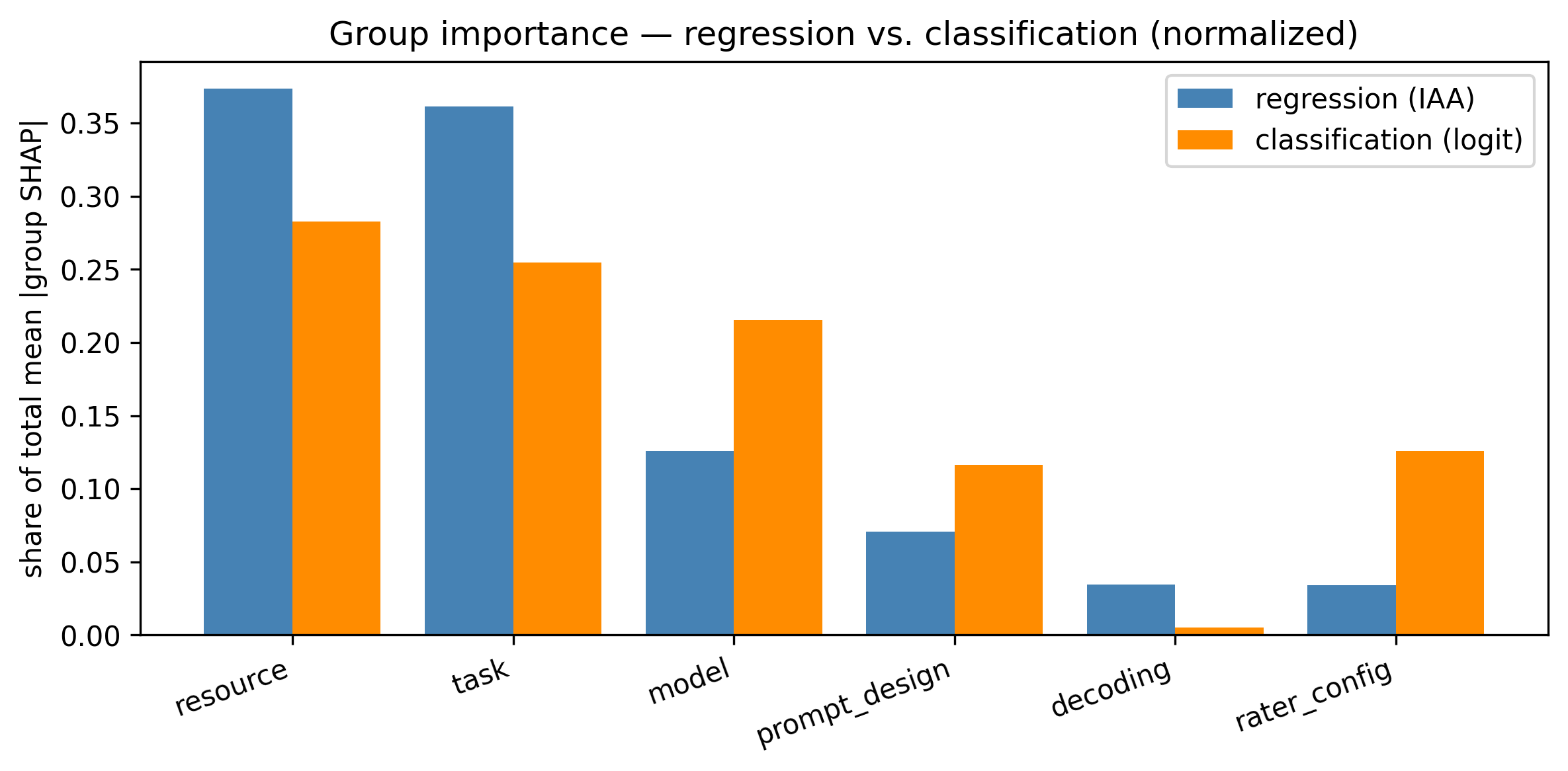}
    \caption{Grouped feature-importance values according to the \textit{SHAP} analysis.}
    \label{fig:results_shap_comparison}
\end{figure}
\begin{figure}[!h]
    \centering
    \makebox[\linewidth][c]{%
    \includegraphics[width=1.15\linewidth]{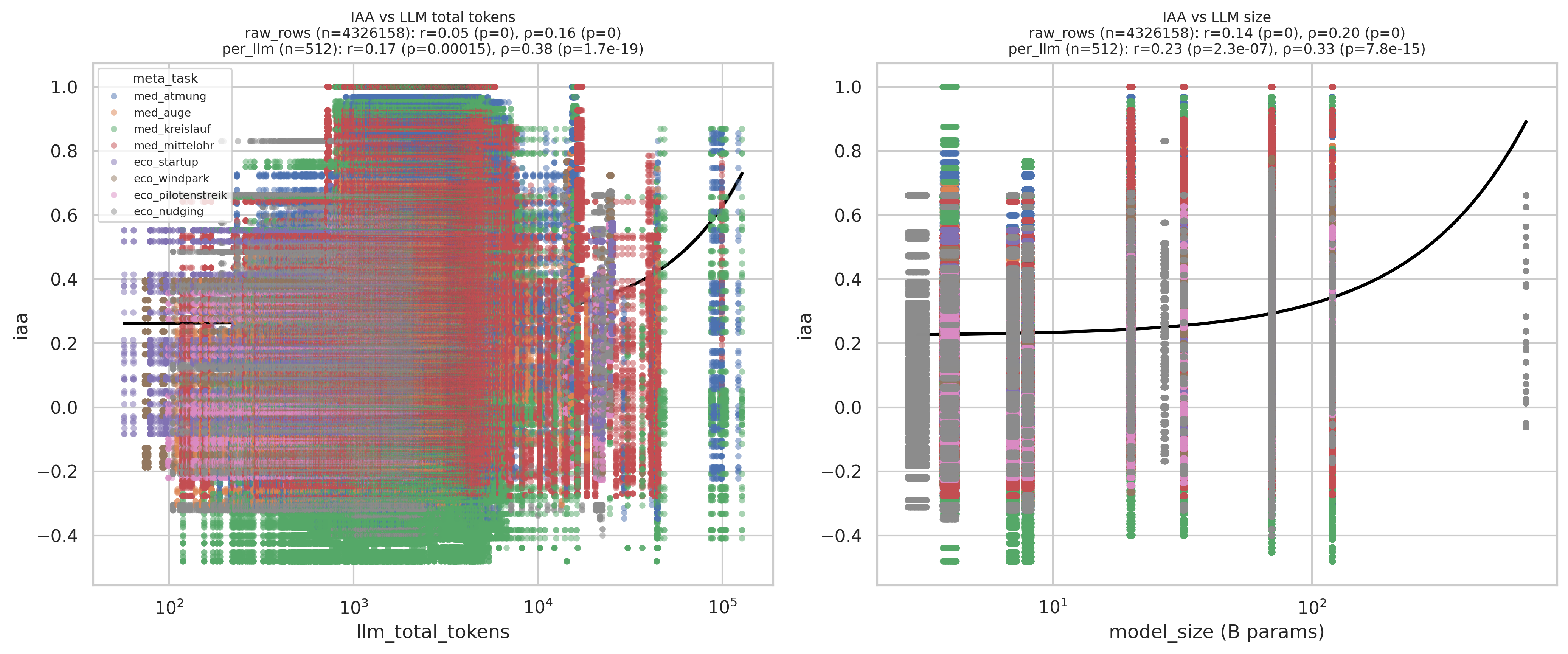}
    }
    \caption{%
    Comparison of the resulting IAA score after replacing one human rater against the total number of tokens consumed and generated by the LLMs (left) and the number of LLM parameters (right).
    Log-scaled x-axis.}
    \label{fig:size_correlation}
\end{figure}

Overall, the importance of the six feature groups is presented in \autoref{fig:results_shap_comparison}, where we show the normalized share of the absolute \textit{SHAP} values per group.
For predicting the IAA values, the \textit{rater} configuration matters less than for determining whether the change in IAA is non-negative, and the \textit{decoding} features also do not play a major role.
For both methods, the \textit{resource}- and \textit{task}-related groups are the most influential, suggesting that the outcomes are sensitive to the actual task, as well as the prompt and context size.
These groups are more important for the prediction of exact IAA values using the regression model.
By comparison, whether an LLM can replace a human rater, as shown by the classification model, depends more on the LLM \textit{model} selection, \textit{prompt design}, and the replaced rater.

The results of the \textit{SHAP} analysis provide an answer to research question \textit{RQ3} and highlight the importance of the task sensitivity.
Within the model-related features, model size appears to be more informative than the concrete model.
To further evaluate this finding, we compare the IAA scores with the total number of tokens and the LLM model size (number of parameters) in \autoref{fig:size_correlation}.
We report statistically significant ($p \leq 0.05$) small to moderate effects using Pearson and Spearman correlations for both variables: token amount ($r=0.17$, $\rho=0.38$) and model size ($r=0.23$, $\rho=0.33$).
Both correlations are positive, confirming an increase in IAA with larger token counts and model sizes, as indicated by the \textit{SHAP} analysis using the proxy model.
However, the effect is not as strong as suggested by \textit{SHAP} and varies depending on the task.
In addition, the plot also shows high performances for smaller model sizes and token counts.

\subsection{Rating Consistency}
\label{sec:results_consistency}

\begin{figure}[t]
    \centering
    \includegraphics[width=.75\linewidth]{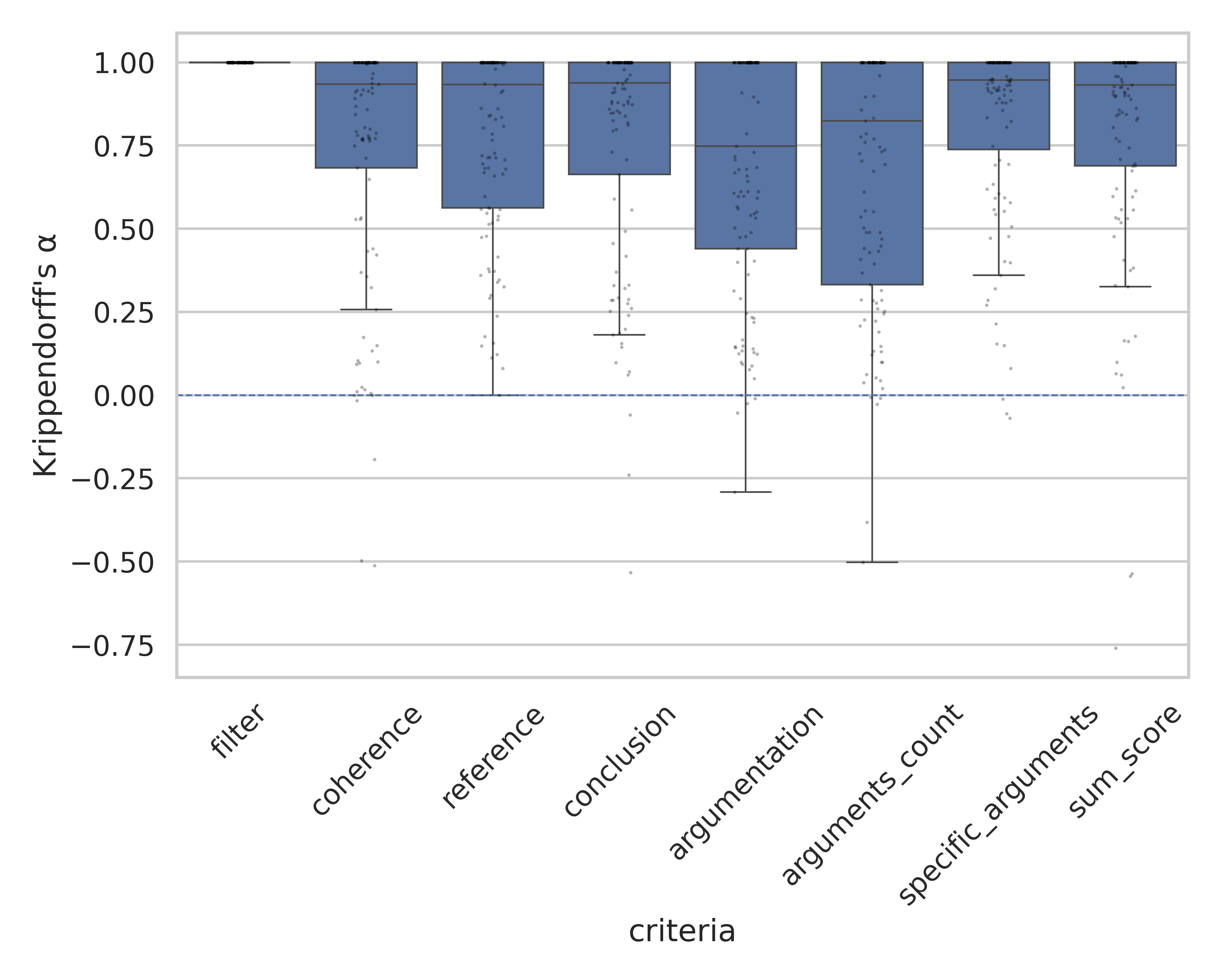}
    \caption{Consistency of LLM ratings based on inter-annotator agreement across five repetitions.}
    \label{fig:llm_consistency}
\end{figure}

\autoref{fig:llm_consistency} shows the inter-annotator agreement across the five repeated ratings for the subset of LLM--prompt combinations from \ExpNu{}.
The results show no deviation for the \CritFilter{} criterion.
While the other criteria do show outliers with negative agreement, we observe high consistency in their ratings on average.

\section{Discussion}
\label{sec:discussion}

Unlike other studies that conclude that LLMs are currently unable to replace human annotators (e.g.,~\citep{Shen:et:al:2023,Chamieh:et:al:2024,Yoshida:2025}), our results suggest a more promising outlook in line with studies reporting LLM performance matching conventional models (e.g.,~\citep{Yancey:et:al:2023,Cong:et:al:2026}).
We show that, for most criteria, at least one LLM--prompt configuration can replace an individual human annotator in essay-grading tasks in educational economics and medicine without lowering the agreement among the raters (\textit{RQ1}).
We demonstrate this for eight tasks across these two domains and per-task rubrics comprising 103 criteria in total (\textit{RQ2}).
We confirm our finding using two metrics:
(1) inter-annotator agreement based on Krippendorff's alpha, which shows replaceability based on agreement patterns;
and (2) the \textit{Alternative Annotator Test}, which provides statistical significance and additionally accounts for the cost of the human annotation.
Our evaluation shows that the agreement is preserved or increased for most criteria when replacing one or two human raters with LLMs, and that multiple LLMs pass the alt-test, meaning they can replace at least 50\% of the human raters.

These are best-case results:
the best configuration differs by task and criterion, and most tested configurations lower the agreement.
Since we compare the LLM ratings with those of the human raters and do not evaluate their absolute quality, a change in agreement does not show whether the LLM-based ratings are better or worse.
An increase may also result from replacing the only rater who rated a response correctly, while a decrease may result from LLM ratings that are more accurate than those of the human raters.
The same applies to the alt-test, which also uses the human raters as the reference.
Instead, we address the question that arises in practice, where gold-standard ratings are rarely available and human raters partly disagree:
whether an LLM can take the place of one of these human raters without lowering their agreement.

We rely on real-world data from the original assessment, manually annotated by multiple trained human raters.
Their inter-annotator agreement shows that constructing a single gold label can be challenging (as shown in \autoref{tab:iaa_humans}).
We therefore treat each rater's score as an equally plausible reference and compare our LLM ratings with those of all human raters:
separately in the case of annotator agreement and combined using the alt-test.
In future work, this \textit{rating indeterminacy} could additionally be evaluated more explicitly, e.g., following the framework presented by \citet{Guerdan:et:al:2025}.
Although our aim is not to imitate individual raters, the replaced rater matters:
replacing a rater whose ratings deviate from the others can increase agreement (\autoref{fig:results_iaa_change_per_human_all_tasks}), and the SHAP analysis identifies the replaced rater as relevant for whether a non-negative change is achieved (\autoref{fig:results_shap_beeswarm}).

While the alt-test shows that LLM--prompt configurations can replace a human rater for all criteria in \TaskEcoNU{} and nearly all criteria overall, including \CritSumscore{}, the IAA results show that the best agreement is achieved by different LLM--prompt combinations.
Using an ensemble of LLMs therefore maximises their effectiveness.
However, our experiments with criterion-focused prompts for task \TaskEcoNU{}, in which the LLM is queried for only one criterion rather than all criteria, show no clear improvement.
This suggests that the LLMs benefit from the explicit inclusion of multiple criteria in their rating process and the interdependencies among criteria in the overall student rating.
Querying one criterion after another while keeping the message history, rather than asking for ratings of all criteria at once, has been shown to be effective for the medical tasks, where the number of criteria is much higher than in economics.
This directly relates to the finding that model size and the number of tokens influence the outcome, as shown in the \textit{SHAP} analysis \textit{RQ3}.
Using a larger model appears to be more important than prompt design.
However, our results also show that, for the majority of tested configurations, the agreement change when replacing a human rater is negative.
\autoref{fig:size_correlation} shows the prompt size, measured as the number of input and output tokens, compared with the change in inter-annotator agreement.
Note that the number of tokens can be considered only an estimate, as it comprises the prompt query, the student response text, additional features, and the LLM outputs.
Although there is a general trend indicating that agreement alphas increase with a larger number of tokens, the results also show that, for all tasks, the best IAAs are not achieved with the largest prompt.
Using an LLM to grade student answers reduces costs and enables applications such as real-time feedback, where prediction speed matters.
This depends directly on the number and size of the prompts.

The ability to use local LLMs is relevant because of AI regulations, data-protection rules, and the need for reproducibility.
Additionally, the use of proprietary LLMs can be complex or forbidden (cf.~\citep{Jayaram:2024,Christakis:2024}) and expensive, especially for larger reasoning models (cf.~\citep{Knoop:Holtmann:2026,Du:2026}).
Therefore, given the data-protection reasons regarding the student statements, our evaluation is restricted to locally hosted, open-weight LLMs.
This conclusion is supported by our results on model parameter size and by the best-performing models based on the alt-test.
These models include a mixture of sizes and families, albeit with a focus on the Llama 3.3 70b LLM, which was also used in the most experiments, suggesting that smaller models and shorter prompts can still deliver sufficient performance.
However, larger models tend to perform better and show greater influence than prompt design.
Our results demonstrate the potential of LLM-based AES, particularly for open-weight LLMs, which have been shown to perform worse than closed LLMs in previous studies (e.g.,~\citep{Mansour:et:al:2024,Stasuik:2025}).
We show that open state-of-the-art models can reach agreement comparable to that among our human annotators.
To this end, while proprietary LLMs might still outperform open models, the quality of the latter has increased in the recent years and now enables their use in this scenario.
The same applies to the need for fine-tuning LLMs and to the development and training of classical approaches.
Although the research by \citet{Chamieh:et:al:2024,Kortemeyer:2023,Latif:Zhai:2024} mentions improvements achieved through fine-tuning, our study suggests that the general knowledge of current LLMs already leads to agreement comparable to those of human annotators.
Not having to fine-tune or further train a LLM makes it more feasible for non-experts to use and develop such a rubric-based LLM-AES system.
This improves the already high level of usability enabled by the ability to control the LLM using natural language.
The results still show that, in the best case, a human rater can be replaced without lowering the agreement for about 88\% of the criteria.
This finding further supports the use of self-hosted open-weight models, reducing resource utilisation, reliance on proprietary LLM providers, and issues related to privacy concerns.

Our results confirm that, while fine-tuning is not necessarily required, the agreement is sensitive to the prompt, as also reported by, e.g., \citet{Mansour:et:al:2024,Lee:et:al:2024}.
However, we also observe that the \texttt{SIMPLE} prompts perform on a par with the more advanced ones when using the Qwen 2.5 and Nemotron LLMs.
The rubric definition in these prompts is kept very concise.
Research shows that LLMs perform worse when irrelevant information is included in the prompt~\citep{Yang:et:al:2025b}, and OpenAI's recommendation for prompting state-of-the-art reasoning models is to use \enquote{straightforward prompts}~\citep{OpenAI:2026}.
By comparison, the better-performing prompts for the Llama model are larger prompts, containing e.g., extracted textual features or training transcripts.
Considering the distribution of the top-performing models and the results of the \textit{SHAP} analysis, our results indicate advantages of more complex prompts regarding most of the criteria covered in our study, similar to the findings of \citet{Huang:et:al:2025}.

We use a direct approach in LLM-based AES, where we ask the LLM for a concrete value and interpret its answer as a rating without further processing, except simple parsing or transformation.
Using the output of an LLM as input for another system -- e.g., as feature generator for a classifier or another LLM -- rather than relying exclusively on the LLM's output for the rating, may improve performance.
This also removes the need for model fine-tuning but instead requires the construction of subsequent tools.
Additionally, combining the outputs of multiple LLMs can balance different model biases and compensate for errors.
See, e.g.,~\citet{Shibata:Miyamura:2025,Sahoo:et:al:2025,Baumartz:et:al:2026}, who evaluate such a classifier-based method in the domain of medical education for assigning holistic ratings.
This should also be evaluated in the field of economics and for more fine-grained rubrics in future work.

To assess the reproducibility of LLM-based ratings, we examined how consistent the LLMs are in their ratings (\autoref{sec:results_consistency}).
Since a temperature of 0.8 permits stochastic generation, this also indicates whether the selected model configuration results in unstable ratings.
The high average consistency suggests stability in the LLMs' application of the rubric, which can facilitate the evaluation of known biases in LLMs and fairness in AI use (cf.~\citep{Johnson:Zhang:2024,Chen:et:al:2024,Weidmann:et:al:2025}).

For holistic essay scoring, earlier work by \citet{Tate:et:al:2024} found no statistically difference between human and ChatGPT scores.
More recent research by \citet{Emirtekin:Ozarslan:2026}, hovever, shows greater agreement for LLM-generated ratings on criteria that are based more on factual and definitional aspects.
They mention, that a holistic score, such as the \CritSumscore{} in our study, is harder for an LLM to rate.
By contrast, for a human rater, this is an easier task than assessing analytical criteria.
The results of our study indicate that this is the case.
We report non-negative IAA changes and mostly high agreement values across all eight tasks for the best human-replacing LLM.
However, the \CritSumscore{} is not predicted directly by the LLM using a prompting approach, but is calculated from the values of the other criteria.
This is consistent with how the \CritSumscore{} is generated from the human ratings -- human raters did not assign a holistic score directly.
This means that the quality of all other criteria factors into the outcome of the composite \CritSumscore{}.
This shows that the performance differences can be smoothed out by the \CritSumscore{} for both human and LLM-generated ratings.
Direct prediction should be further explored in future work, and, ideally, also compared with direct assignment by human raters.

In total, we see that the ratings generated by the LLMs do not lower the group agreement when one human rater is replaced by one LLM for 91 of the 103 criteria.
This trend continues when we replace two humans with LLMs:
we observe a tendency toward higher agreement among LLMs than among human raters, a finding that was also reported by \citet{Frohn:et:al:2025}.

Unlike the results presented by \citet{Deng:et:al:2025}, we do not observe a decrease in agreement for the non-Boolean criteria that we studied.
Across all categories, automatic LLM-based grading is able to produce ratings that achieve a comparable level of agreement with the human annotators for all criterion categories, i.e., regarding \textit{formal}, \textit{structural} and \textit{domain-specific} criteria in economics and medical education.

Our results also show that human review is still required before employing LLM-based AES.
While simple prompts can perform adequately, the variation in IAA scores and the influence of the tasks itself suggests a validation is necessary, especially when introducing new tasks, questions, or domains into the assessments.
In practice, LLM-based scoring should therefore be validated against human ratings per task and criterion before use, since the best configuration varies.
As starting point, our results suggest larger models and prompts that include training material for economics and criterion-wise chaining for medicine.

\section{Limitations}
\label{sec:limitations}

This study evaluates whether LLM-based systems can replace human raters in a group without reducing agreement.
We do not assess or analyse whether the LLMs produce objectively correct ratings.
We provide empirical evidence that current state-of-the-art open-weight LLMs can replace individual human annotators in assessing student responses in economics and medical education without lowering the agreement:
our results suggest preserved or increasing agreement among the top performers when replacing one or two human raters with LLM-based systems.
However, the majority of tested LLM--prompt configurations lead to decreasing agreement.
This indicates that a full replacement of all human raters is not possible and that human validation is still required.

The central limitation concerns the quality of the human ratings we use for reference in our evaluation.
This is reflected by the level of disagreement measured by Krippendorff's alpha, especially in economics, where only 5 of 31 criteria reach at least tentatively acceptable agreement, and could indicate an issue with the rubric we study.
The student statements were evaluated by two independent rater groups, which received task-specific instructions and training, including trial ratings.
However, no consolidated gold rating is available for use in our study.
Our results must therefore be interpreted as empirical evidence in the context of the observed human ratings and cannot make any assumption on the objective correctness of the LLM-generated ratings.
In particular, for criteria with low human agreement, not lowering the agreement is a weaker requirement than for criteria on which the human raters agree, and a change in agreement does not indicate whether the LLMs are better or worse.
We provide complementary measures using Cronbach's alpha and ICC(3,k) to offer another perspective on the human rating agreement.
Additionally, the \textit{Alternative Annotator Test} further provides insights into the replaceability in relation to the human rater group as a whole. 

Further limitations concern the number of student responses we analyse, as well as our restriction to a subset of tasks within each of the two disciplines.
In addition, the tasks, their length requirements, and the rubrics were designed separately for each domain as part of the original assessment, differences between the domains, e.g., in human agreement, may partly reflect these design differences and should be interpreted with caution.

We use only open-weight and locally hostable LLMs for data-protection reasons. 
Consequently, we do not compare our results with proprietary models.
While this may seem limiting at first, the literature reviewed in \autoref{sec:related_work} suggests that proprietary models could achieve even better scores because of their larger training corpora and parameter spaces~\citep{Stasuik:2025}.
Our results show a small to moderate correlation between model parameter size and observed IAA-based performance.

Finally, we only consider criteria that do not require the LLM to use other content than the students' responses.
Using agentic or tool-based LLM workflows for on-demand access to additional data, such as online resources, or using a RAG-based approach, should be considered in future work.

\section{Conclusion}
\label{sec:conclusion}

We have shown the feasibility of using open-weight LLM-based AES for student assessments in the fields of educational economics and medicine.
Our results, which are based on real-world data from four tasks per domain, show that, for most criteria, at least one LLM--prompt configuration can replace an individual human rater without the need for fine-tuning while maintaining comparable inter-annotator agreements, as measured by Krippendorff's alpha.
For 88\%, i.e., 91 of the 103 criteria in our rubric, the agreement is not lowered when replacing one human rater, with similar results when replacing two.
At least one configuration also passes the \textit{Alternative Annotator Test} with statistical significance, confirming the potential to replace at least 50\% of human raters across 87\% of the criteria.
These are best-case results: most tested configurations lower the agreement.
We show that the best-performing prompts and LLMs vary by criterion, however, there is no clear indication that criteria-specific prompts lead to improvements.
As no gold standard for the ratings is available, our results show agreement with the observed human raters rather than the absolute correctness of the LLM ratings.
Performance is associated more with model and prompt size than with prompt design.
There is no visible difference in LLM performance across the three criteria categories; the models are able to assign ratings similar to those of the human raters for \textit{formal}, \textit{structural}, and \textit{domain-knowledge} criteria across both domains.
While LLMs avoid the individual effects of the human raters they replace, they introduce their own biases into the ratings.
However, using automated methods can lead to more stable assessments and enable faster processes, including real-time feedback to students while they are writing.
Human validation is still needed to ground LLM-based AES systems during development.
Together, these findings suggest the use of LLM ensembles to benefit from higher top performances per criterion and to mitigate model and prompt biases.

\section*{Ethical Considerations}
\label{sec:ethics}

The study uses written statements produced by participants in an educational setting, as well as rating data generated by human raters.
As part of a multicenter research project, this study was approved by the ethics committee of LMU (approval reference no.~23-0688). 
We received the dataset in pseudonymised form from the project partners as part of an existing educational assessment in the context of the CORE project.
Although we had access to the original written texts, all personally identifiable information had been removed before we received the data.
All processing of student statements took place on local infrastructure.

\section*{Conflict of Interest}
\label{sec:conflict_interest_statement}

The authors declare no potential conflict of interests.

\section*{Acknowledgments}
\label{sec:acknowledgements}

This work was supported by the research group \emph{Critical Online Reasoning in Higher Education} (CORE, FOR 5404, grant number 462702138) sub-projects B05 and C08, and the Infrastructure Priority Programme \emph{New Data Spaces for the Social Sciences} (SPP 2431, grant number 539634240) sub-project ENTAILab, both funded by the German Research Foundation (DFG).

\section*{Declaration of Generative AI and AI-assisted Technologies in the Manuscript Preparation Process}

During the preparation of this work the authors used ChatGPT and Claude in order to identify relevant literature, support with programming, and improve the clarity and quality of the language.
After using this tool/service, the authors reviewed and edited the content as needed and take full responsibility for the content of the published article.

\clearpage

\appendix

\section{Rating Distribution}
\label{sec:appendix:rating_distribution}

\begin{figure}[th]
    \centering
    \begin{subfigure}{1\linewidth}
    \centering
    \includegraphics[trim=0 0 25.5cm 0,clip,width=1\linewidth]{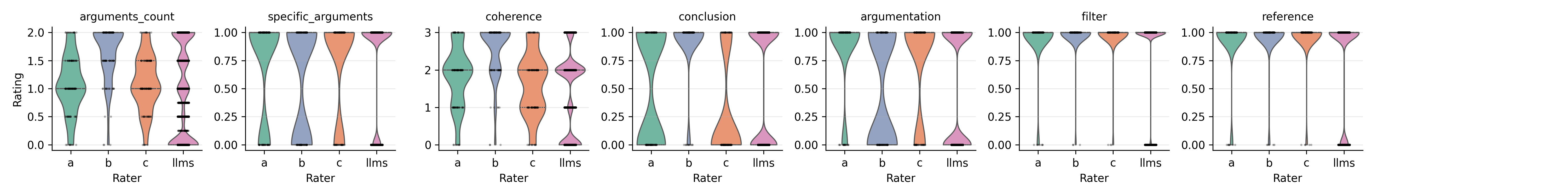}
    \includegraphics[trim=25.5cm 0 0 0,clip,width=1\linewidth]{humans_llms_rating_distribution_nudging.png}
    \caption{Economics task \TaskEcoNU{}}
    \end{subfigure}
    \begin{subfigure}{1\linewidth}
    \centering
    \includegraphics[trim=0 0 25.5cm 0,clip,width=1\linewidth]{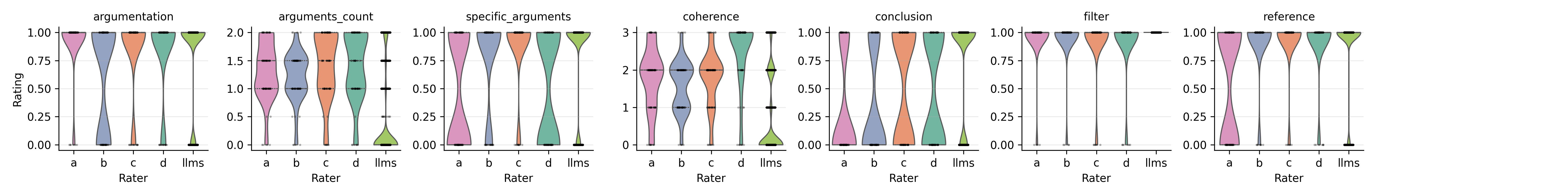}
    \includegraphics[trim=25.5cm 0 0 0,clip,width=1\linewidth]{humans_llms_rating_distribution_startup.png}
    \caption{Economics task \TaskEcoSU{}}
    \end{subfigure}
\end{figure}

\begin{figure}[th]
    \ContinuedFloat

    \centering
    \begin{subfigure}{1\linewidth}
    \centering
    \includegraphics[trim=0 0 25.5cm 0,clip,width=1\linewidth]{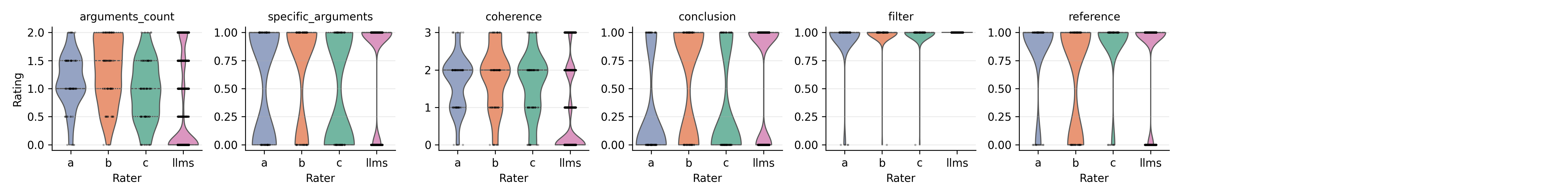}
    \includegraphics[trim=25.5cm 0 0 0,clip,width=1\linewidth]{humans_llms_rating_distribution_pilotenstreik.png}
    \caption{Economics task \TaskEcoPS{}}
    \end{subfigure}
    \begin{subfigure}{1\linewidth}
    \centering
    \includegraphics[trim=0 0 25.5cm 0,clip,width=1\linewidth]{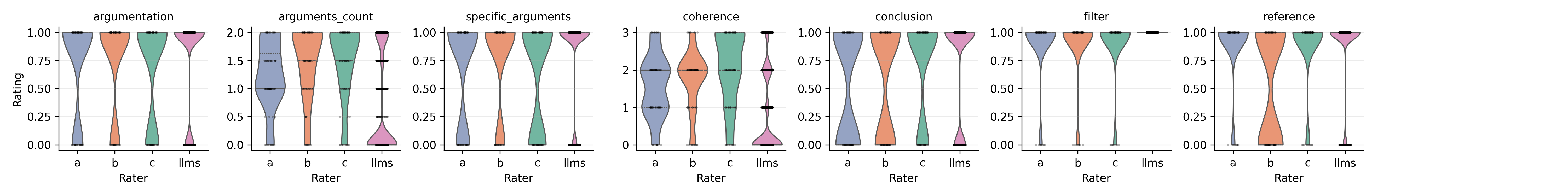}
    \includegraphics[trim=25.5cm 0 0 0,clip,width=1\linewidth]{humans_llms_rating_distribution_windpark.png}
    \caption{Economics task \TaskEcoWP{}}
    \end{subfigure}
    \caption{Distribution of ratings for each economics criterion assigned by human annotators and LLMs.}
    \label{fig:humans_llms_rating_distribution_eco}
\end{figure}

\begin{figure}[th]
    \centering
    \begin{subfigure}{1\linewidth}
    \centering
    \includegraphics[trim=0 7cm 25cm 0,clip,width=1\linewidth]{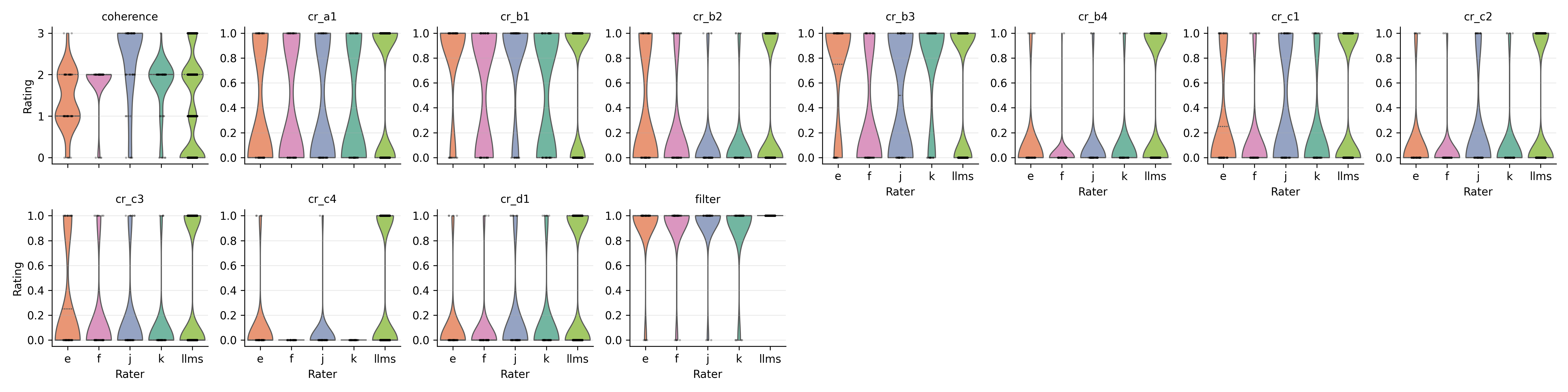}
    \includegraphics[trim=25.5cm 7.25cm 0 0,clip,width=1\linewidth]{humans_llms_rating_distribution_auge.png}
    \includegraphics[trim=.76cm 0 25cm 5.5cm,clip,width=1\linewidth]{humans_llms_rating_distribution_auge.png}
    \caption{Medicine task \TaskMedAU{}}
    \end{subfigure}  
\end{figure}

\begin{figure}[th]
    \ContinuedFloat
    \begin{subfigure}{1\linewidth}
    \centering
    \includegraphics[trim=0 7cm 25cm 0,clip,width=1\linewidth]{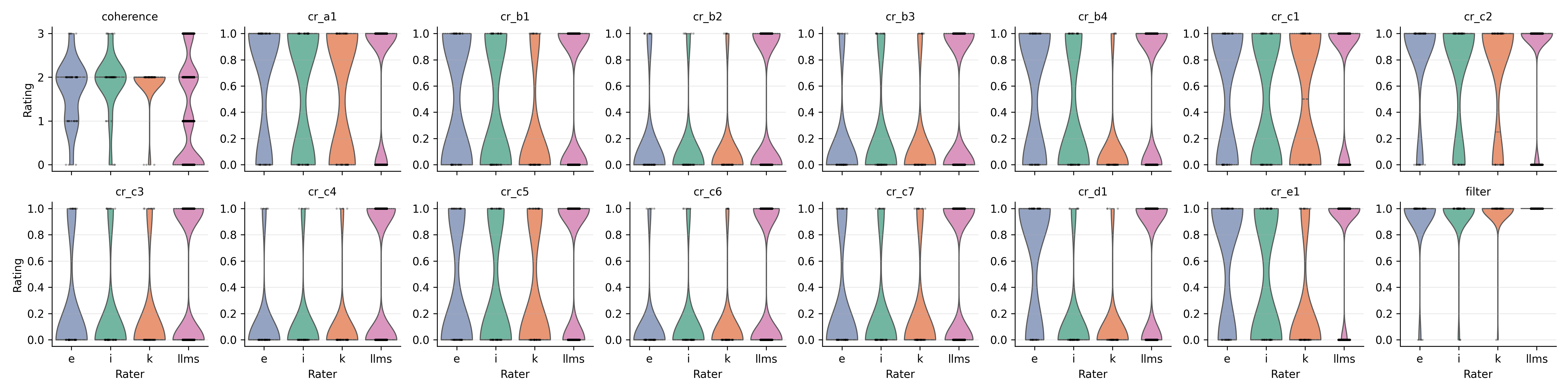}
    \includegraphics[trim=25.5cm 7.25cm 0 0,clip,width=1\linewidth]{humans_llms_rating_distribution_atmung.png}
    \includegraphics[trim=.76cm 0 25cm 5.5cm,clip,width=1\linewidth]{humans_llms_rating_distribution_atmung.png}
    \includegraphics[trim=25.5cm 0 0 6cm,clip,width=1\linewidth]{humans_llms_rating_distribution_atmung.png}
    \caption{Medicine task \TaskMedAT{}}
    \end{subfigure}
\end{figure}

\begin{figure}[th]
    \ContinuedFloat
    \begin{subfigure}{1\linewidth}
    \centering
    \includegraphics[trim=0 13cm 25cm 0,clip,width=1\linewidth]{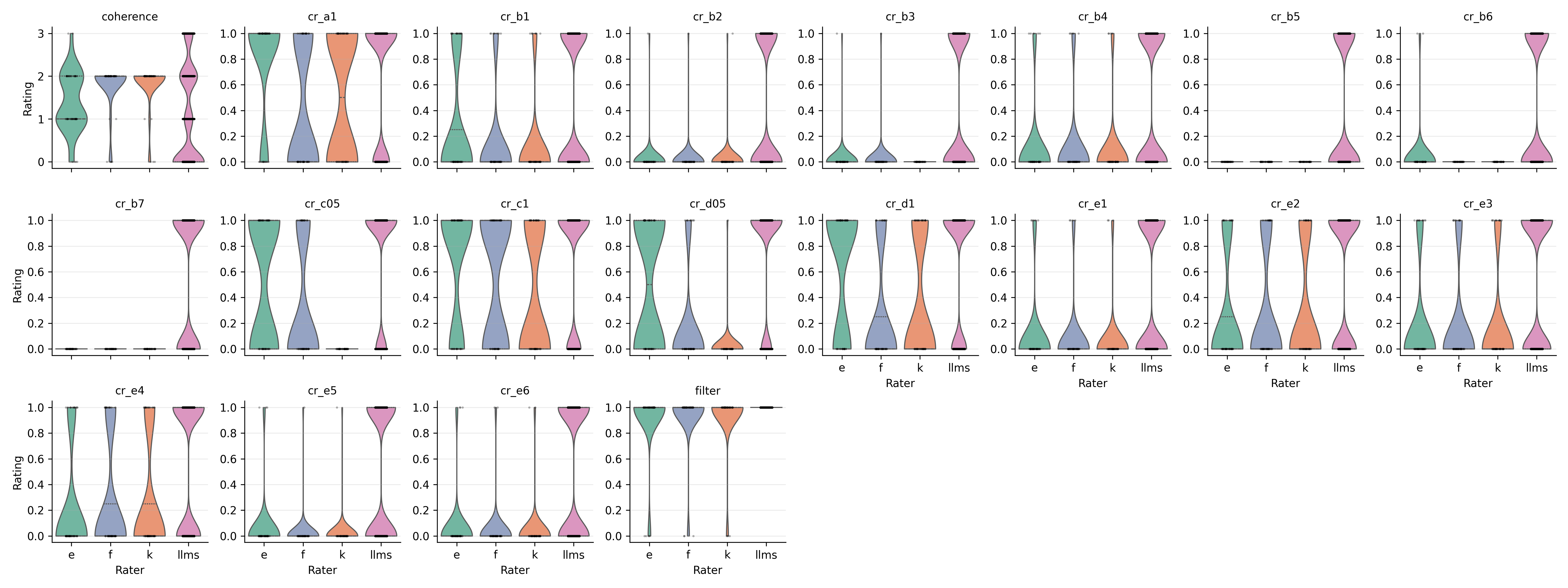}
    \includegraphics[trim=25.5cm 13cm 0 0,clip,width=1\linewidth]{humans_llms_rating_distribution_kreislauf.png}
    \includegraphics[trim=.76cm 7.25cm 25cm 6cm,clip,width=1\linewidth]{humans_llms_rating_distribution_kreislauf.png}
    \includegraphics[trim=25.5cm 7.25cm 0 6cm,clip,width=1\linewidth]{humans_llms_rating_distribution_kreislauf.png}
    \includegraphics[trim=.76cm 0 25cm 12cm,clip,width=1\linewidth]{humans_llms_rating_distribution_kreislauf.png}
    \caption{Medicine task \TaskMedKL{}}
    \end{subfigure}
\end{figure}

\begin{figure}[th]
    \ContinuedFloat
    \begin{subfigure}{1\linewidth}
    \centering
    \includegraphics[trim=0 13cm 25cm 0,clip,width=1\linewidth]{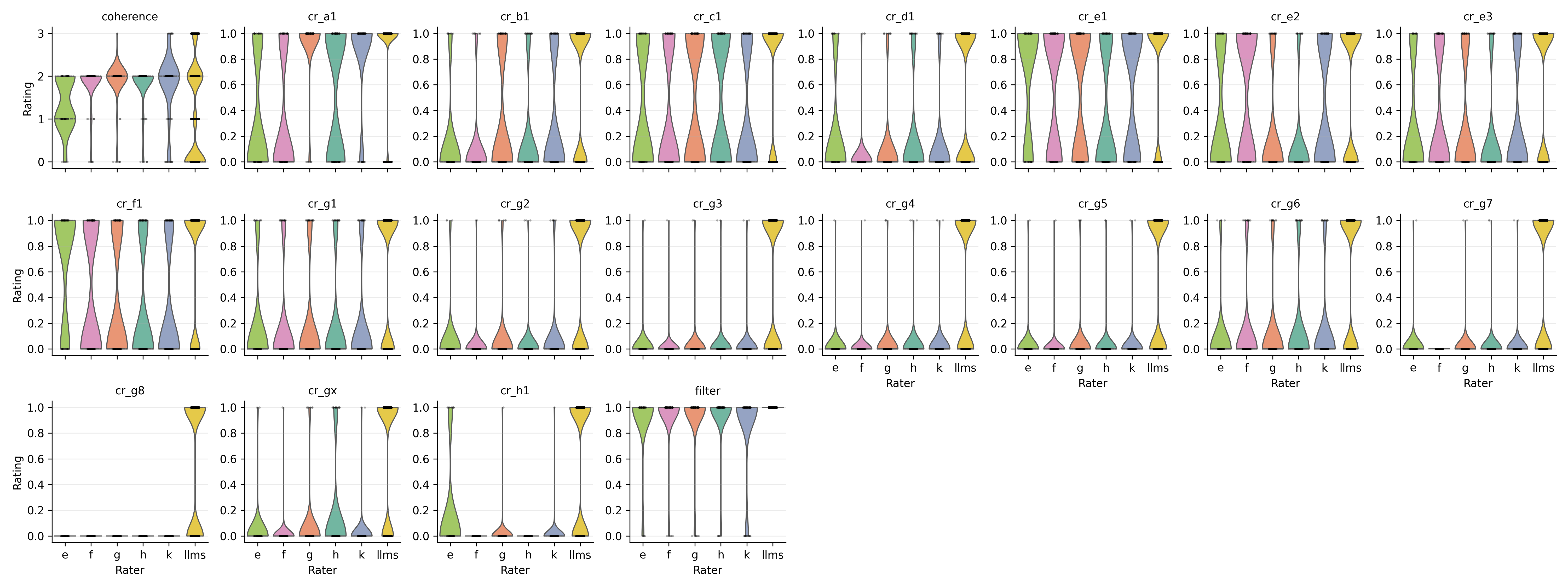}
    \includegraphics[trim=25.5cm 13cm 0 0,clip,width=1\linewidth]{humans_llms_rating_distribution_mittelohr.png}
    \includegraphics[trim=.76cm 7.25cm 25cm 6cm,clip,width=1\linewidth]{humans_llms_rating_distribution_mittelohr.png}
    \includegraphics[trim=25.5cm 7.25cm 0 6cm,clip,width=1\linewidth]{humans_llms_rating_distribution_mittelohr.png}
    \includegraphics[trim=.76cm 0 25cm 12cm,clip,width=1\linewidth]{humans_llms_rating_distribution_mittelohr.png}
    \caption{Medicine task \TaskMedMO{}}
    \end{subfigure}
    \caption{Distribution of ratings for each medical criterion assigned by human annotators and LLMs.}
    \label{fig:humans_llms_rating_distribution_med}
\end{figure}

\clearpage

\section{LLM-based IAA Scores}
\label{sec:appendic:llm_iaa_change_tables}

\begin{table}[th]
    \centering
    \begin{tabular}{l|r|r|r|l|l}
        \toprule
        Criteria & IAA & Change & \# & LLM & Prompt \\
        \midrule
filter & 0.185 & 0.038 & 51 & Plutus & CHAIN NOANSWER \\ 
coherence & 0.658 & 0.019 & 1 & Nemotron 70b & SIMPLE \\ 
arguments & 0.437 & 0.045 & 1 & Qwen 2.5 32b & SIMPLE \\ 
balance & 0.571 & 0.195 & 1 & Llama 3.3 70b & TRAINING ECO (full text) \\ 
conclusion & 0.448 & 0.126 & 1 & GPT-OSS 20b & CHAIN NOANSWER \\ 
problem & 0.451 & 0.203 & 1 & Llama 3.3 70b & TRAINING (full text) \\ 
 sum\_score & 0.651 & -0.023 & 1 & Nemotron 70b & FEATURES (JSON) \\ 
\bottomrule
    \end{tabular}
    \caption{%
    Overview of the top IAA scores after replacing a human rater with an LLM-based system in \ExpAll{} task \TaskEcoPS{}.
    }
    \label{tab:results_iaa_change_per_human_all_tasks_eco_ps}
\end{table}

\begin{table}[th]
    \centering
    \begin{tabular}{l|r|r|r|l|l}
        \toprule
        Criteria & IAA & Change & \# & LLM & Prompt \\
        \midrule
filter & 0.552 & -0.270 & 51 & Plutus & CHAIN NOANSWER \\ 
coherence & 0.485 & 0.111 & 1 & Llama 3.3 70b & CHAIN NOANSWER \\ 
argumentation & 0.560 & 0.200 & 1 & Qwen 2.5 32b & SIMPLE \\ 
arguments & 0.639 & 0.060 & 1 & Llama 3.3 70b & TRAINING (full text) \\ 
balance & 0.364 & 0.161 & 1 & Qwen 2.5 32b & SIMPLE \\ 
conclusion & 0.691 & 0.156 & 1 & Nemotron 70b & CHAIN NOANSWER \\ 
problem & 0.669 & 0.419 & 1 & Llama 3.3 70b & TRAINING (full text) \\ 
sum\_score & 0.680 & 0.101 & 1 & Llama 3.3 70b & TRAINING (full text) \\ 
\bottomrule
    \end{tabular}
    \caption{%
    Overview of the top IAA scores after replacing a human rater with an LLM-based system in \ExpAll{} task \TaskEcoSU{}.
    }
    \label{tab:results_iaa_change_per_human_all_tasks_eco_su}
\end{table}

\begin{table}[th]
    \centering
    \begin{tabular}{l|r|r|r|l|l}
        \toprule
        Criteria & IAA & Change & \# & LLM & Prompt \\
        \midrule
filter & 0.396 & -0.448 & 49 & GPT-OSS 120b & CHAIN NOANSWER \\ 
coherence & 0.610 & 0.124 & 1 & Nemotron 70b & SIMPLE \\ 
argumentation & 0.606 & 0.245 & 1 & Llama 3.3 70b & SIMPLE \\ 
arguments & 0.649 & 0.046 & 1 & Llama 3.3 70b & TRAINING (full text) \\ 
balance & 0.766 & 0.140 & 1 & Llama 3.3 70b & TRAINING (full text) \\ 
conclusion & 0.413 & -0.004 & 1 & Llama 3.3 70b & TRAINING ECO (full text) \\ 
problem & 0.778 & 0.550 & 1 & Llama 3.3 70b & FEATURES (JSON) \\ 
sum\_score & 0.719 & 0.012 & 1 & Llama 3.3 70b & TRAINING (full text) \\ 
\bottomrule
    \end{tabular}
    \caption{%
    Overview of the top IAA scores after replacing a human rater with an LLM-based system in \ExpAll{} task \TaskEcoWP{}.
    }
    \label{tab:results_iaa_change_per_human_all_tasks_eco_wp}
\end{table}

\begin{table}[th]
    \centering
    \begin{tabular}{l|r|r|r|l|l}
        \toprule
        Criteria & IAA & Change & \# & LLM & Prompt \\
        \midrule
filter & 0.661 & -0.019 & 51 & Plutus & CHAIN NOANSWER \\ 
coherence & 0.538 & 0.235 & 1 & Nemotron 70b & SIMPLE \\ 
reference & 0.625 & -0.023 & 2 & MedGemma 1.5 4b & FEATURES (JSON) \\ 
conclusion & 0.471 & 0.436 & 1 & GPT-OSS 120b & CHAIN NOANSWER \\ 
argumentation & 0.305 & 0.132 & 1 & Llama 3.3 70b & TRAINING (full text) \\ 
arguments\_count & 0.357 & 0.046 & 1 & Qwen 2.5 32b & SIMPLE \\ 
specific\_arguments & 0.477 & 0.073 & 1 & Llama 3.3 70b & SIMPLE \\ 
sum\_score & 0.709 & 0.056 & 1 & Llama 3.3 70b & SIMPLE \\ 
\bottomrule
    \end{tabular}
    \caption{%
    Overview of the top IAA scores after replacing a human rater with an LLM-based system in \ExpAll{} task \TaskEcoNU{}.
    }
    \label{tab:results_iaa_change_per_human_all_tasks_eco_nu_all}
\end{table}

\begin{table}[th]
    \centering
    \begin{tabular}{l|r|r|r|l|l}
        \toprule
        Criteria & IAA & Change & \# & LLM & Prompt \\
        \midrule
filter & 0.830 & 0.150 & 1 & Gemma 2 27b & CHAIN \\ 
coherence & 0.536 & 0.234 & 1 & DeepSeek-R1 70b & FEATURES (JSON) \\ 
reference & 0.741 & 0.093 & 3 & Llama 3.3 70b & FEATURES \\ 
 &  &  &  &  & ELABORATION (JSON) \\ 
conclusion & 0.422 & 0.388 & 3 & DeepSeek-R1 70b & CHAIN \\ 
argumentation & 0.301 & 0.128 & 1 & Mixtral 8x7b & CHAIN \\ 
arguments\_count & 0.389 & 0.079 & 1 & Llama 3.2 3b & CHAIN \\ 
specific\_arguments & 0.518 & 0.114 & 1 & Llama 3.3 70b & OPT \\ 
sum\_score & 0.729 & 0.076 & 1 & Llama 3.3 70b & FEATURES \\ 
 &  &  & &  &  ELABORATION (JSON) \\ 
\midrule
arguments\_count & 0.304 & -0.007 & 1 & DeepSeek-R1 70b & STRUCT \\ 
\midrule
coherence & 0.555 & 0.253 & 1 & Llama 3.3 70b & STRUCT CHAIN FEWSHOT \\ 
\bottomrule
    \end{tabular}
    \caption{%
    Overview of the top IAA scores after replacing a human rater with an LLM-based system in \ExpNu{}.
    }
    \label{tab:results_iaa_change_per_human_all_tasks_eco_nu_2}
\end{table}

\begin{table}[th]
    \centering
    \begin{tabular}{l|r|r|r|l|l}
        \toprule
        Criteria & IAA & Change & \# & LLM & Prompt \\
        \midrule
filter & 0.323 & -0.158 & 49 & GPT-OSS 120b & CHAIN NOANSWER \\ 
coherence & 0.531 & 0.093 & 1 & Llama 3.3 70b & CHAIN NOANSWER \\ 
cr\_a1 & 0.912 & 0.298 & 4 & Llama 3.3 70b & CHAIN NOANSWER \\ 
cr\_b1 & 0.557 & 0.139 & 1 & Nemotron 70b & FEATURES (JSON) \\ 
cr\_b2 & 0.930 & 0.078 & 1 & DeepSeek-R1 70b & OPT \\ 
cr\_b3 & 0.939 & 0.100 & 1 & DeepSeek-R1 70b & OPT \\ 
cr\_b4 & 0.620 & 0.364 & 1 & GPT-OSS 120b & OPT \\ 
cr\_c1 & 0.913 & 0.230 & 5 & GPT-OSS 120b & OPT \\ 
cr\_c2 & 0.848 & 0.164 & 3 & Llama 3.3 70b & CHAIN NOANSWER \\ 
cr\_c3 & 0.942 & 0.092 & 2 & Llama 3.3 70b & FEATURES (JSON) \\ 
cr\_c4 & 0.921 & 0.158 & 4 & Llama 3.3 70b & CHAIN NOANSWER \\ 
cr\_c5 & 1.000 & 0.032 & 3 & Llama 3.3 70b & CHAIN NOANSWER \\ 
cr\_c6 & 0.909 & -0.006 & 1 & GPT-OSS 120b & CHAIN NOANSWER \\ 
cr\_c7 & 1.000 & 0.137 & 4 & GPT-OSS 120b & OPT \\ 
cr\_d1 & 0.636 & 0.559 & 1 & Llama 3.3 70b & CHAIN NOANSWER \\ 
cr\_e1 & 0.473 & 0.208 & 1 & GPT-OSS 120b & OPT \\ 
sum\_score & 0.818 & 0.100 & 1 & Llama 3.3 70b & CHAIN NOANSWER \\ 
\bottomrule
    \end{tabular}
    \caption{%
    Overview of the top IAA scores after replacing a human rater with an LLM-based system in \ExpAll{} task \TaskMedAT{}.
    }
    \label{tab:results_iaa_change_per_human_all_tasks_med_at}
\end{table}

\begin{table}[th]
    \centering
    \begin{tabular}{l|r|r|r|l|l}
        \toprule
        Criteria & IAA & Change & \# & LLM & Prompt \\
        \midrule
filter & 0.395 & -0.111 & 49 & Plutus & CHAIN NOANSWER \\ 
coherence & 0.377 & 0.145 & 1 & Llama 3.3 70b & CHAIN NOANSWER \\ 
cr\_a1 & 0.816 & 0.118 & 5 & GPT-OSS 120b & OPT \\ 
cr\_b1 & 0.452 & 0.130 & 1 & Llama 3.3 70b & CHAIN NOANSWER \\ 
cr\_c1 & 0.633 & 0.118 & 1 & Nemotron 70b  & CHAIN NOANSWER \\ 
cr\_b2 & 0.477 & 0.190 & 1 & Llama 3.3 70b & TRAINING (full text) \\ 
cr\_c2 & 0.508 & 0.123 & 1 & Llama 3.3 70b & TRAINING (full text) \\ 
cr\_b3 & 0.412 & 0.192 & 1 & GPT-OSS 120b & OPT \\ 
cr\_c3 & 0.679 & 0.295 & 3 & Llama 3.3 70b & CHAIN NOANSWER \\ 
cr\_b4 & 0.155 & 0.108 & 1 & GPT-OSS 20b & OPT \\ 
cr\_c4 & 0.191 & 0.216 & 1 & DeepSeek-R1 70b & CHAIN NOANSWER \\ 
cr\_d1 & 0.677 & 0.154 & 1 & GPT-OSS 120b & OPT \\ 
sum\_score & 0.683 & 0.115 & 1 & Nemotron 70b & FEATURES (JSON) \\ 
\bottomrule
    \end{tabular}
    \caption{%
    Overview of the top IAA scores after replacing a human rater with an LLM-based system in \ExpAll{} task \TaskMedAU{}.
    }
    \label{tab:results_iaa_change_per_human_all_tasks_med_au}
\end{table}

\begin{table}[th]
    \centering
    \begin{tabular}{l|r|r|r|l|l}
        \toprule
        Criteria & IAA & Change & \# & LLM & Prompt \\
        \midrule
filter & 0.655 & -0.238 & 49 & Plutus & CHAIN NOANSWER \\ 
coherence & 0.605 & 0.234 & 1 & Llama 3.3 70b & CHAIN NOANSWER \\ 
cr\_a1 & 0.662 & 0.307 & 1 & Nemotron 70b & FEATURES (JSON) \\ 
cr\_b1 & 0.940 & 0.261 & 10 & Llama 3.3 70b & CHAIN NOANSWER \\ 
cr\_b2 & 1.000 & 0.000 & 30 & Llama 3.3 70b & CHAIN NOANSWER \\ 
cr\_b3 & 1.000 & 0.503 & 9 & Nemotron 70b & CHAIN NOANSWER \\ 
cr\_b4 & 1.000 & 0.078 & 10 & Llama 3.3 70b & CHAIN NOANSWER \\ 
cr\_b5 & 1.000 & 0.000 & 39 & Llama 3.3 70b & CHAIN NOANSWER \\ 
cr\_b6 & 1.000 & 1.006 & 6 & Llama 3.3 70b & CHAIN NOANSWER \\ 
cr\_b7 & 0.000 & -1.000 & 3 & Plutus & CHAIN NOANSWER \\ 
cr\_c1 & 0.590 & 0.128 & 1 & Qwen 2.5 32b & OPT \\ 
cr\_c05 & 0.333 & 0.353 & 1 & Llama 3.3 70b & CHAIN NOANSWER \\ 
cr\_d1 & 0.590 & 0.275 & 1 & Qwen 2.5 32b & OPT \\ 
cr\_d05 & 0.429 & 0.344 & 1 & GPT-OSS 20b & OPT \\ 
cr\_e1 & 0.894 & 0.163 & 4 & Nemotron 70b & CHAIN NOANSWER \\ 
cr\_e2 & 0.968 & 0.066 & 3 & Llama 3.3 70b & FEATURES (JSON) \\ 
cr\_e3 & 1.000 & 0.089 & 5 & Nemotron 70b & CHAIN NOANSWER \\ 
cr\_e4 & 1.000 & 0.070 & 6 & Llama 3.3 70b & CHAIN NOANSWER \\ 
cr\_e5 & 1.000 & 0.411 & 6 & Llama 3.3 70b & CHAIN NOANSWER \\ 
cr\_e6 & 1.000 & 0.149 & 4 & GPT-OSS 120b & OPT \\ 
sum\_score & 0.823 & 0.110 & 1 & Qwen 2.5 32b & OPT \\ 
\bottomrule
    \end{tabular}
    \caption{%
    Overview of the top IAA scores after replacing a human rater with an LLM-based system in \ExpAll{} task \TaskMedKL{}.
    }
    \label{tab:results_iaa_change_per_human_all_tasks_med_kl}
\end{table}

\begin{table}[th]
    \centering
    \begin{tabular}{l|r|r|r|l|l}
        \toprule
        Criteria & IAA & Change & \# & LLM & Prompt \\
        \midrule
filter & 0.642 & -0.115 & 48 & Plutus & CHAIN NOANSWER \\ 
coherence & 0.596 & -0.001 & 1 & GPT-OSS 120b & OPT \\ 
cr\_a1 & 0.156 & 0.037 & 1 & GPT-OSS 20b & OPT \\ 
cr\_b1 & 0.520 & 0.186 & 1 & Nemotron 70b & SIMPLE \\ 
cr\_c1 & 0.720 & 0.055 & 1 & GPT-OSS 120b & OPT \\ 
cr\_d1 & 0.497 & 0.124 & 1 & Llama 3.3 70b & CHAIN NOANSWER \\ 
cr\_e1 & 0.776 & 0.130 & 1 & Nemotron 70b & FEATURES (JSON) \\ 
cr\_e2 & 0.524 & 0.152 & 1 & Llama 3.3 70b & OPT \\ 
cr\_e3 & 0.798 & 0.149 & 1 & Qwen 2.5 32b & SIMPLE \\ 
cr\_f1 & 0.615 & 0.082 & 1 & Nemotron 70b & FEATURES (JSON) \\ 
cr\_g1 & 0.897 & 0.038 & 1 & GPT-OSS 120b & OPT \\ 
cr\_g2 & 0.675 & 0.068 & 2 & Llama 3.3 70b & CHAIN NOANSWER \\ 
cr\_g3 & 1.000 & 0.127 & 7 & Llama 3.3 70b & CHAIN NOANSWER \\ 
cr\_g4 & 0.926 & 0.113 & 4 & Llama 3.3 70b & CHAIN NOANSWER \\ 
cr\_g5 & 0.898 & 0.124 & 5 & Llama 3.3 70b & CHAIN NOANSWER \\ 
cr\_g6 & 0.910 & -0.001 & 3 & Llama 3.3 70b & CHAIN NOANSWER \\ 
cr\_g7 & 0.898 & 0.188 & 3 & GPT-OSS 120b & OPT \\ 
cr\_g8 & 1.000 & 0.000 & 30 & Llama 3.3 70b & CHAIN NOANSWER \\ 
cr\_gx & 0.256 & 0.113 & 1 & GPT-OSS 120b & OPT \\ 
cr\_h1 & 0.090 & 0.061 & 1 & Qwen 2.5 32b & OPT \\ 
sum\_score & 0.757 & 0.003 & 1 & GPT-OSS 20b & OPT \\ 
\bottomrule
    \end{tabular}
    \caption{%
    Overview of the top IAA scores after replacing a human rater with an LLM-based system in \ExpAll{} task \TaskMedMO{}.
    }
    \label{tab:results_iaa_change_per_human_all_tasks_med_mo}
\end{table}

\clearpage

\section{Top LLM Jaccard Similarity}
\label{sec:appendix:llm_jaccard}

\begin{figure}
    \centering
    \makebox[\linewidth][c]{%
    \includegraphics[width=1.5\linewidth]{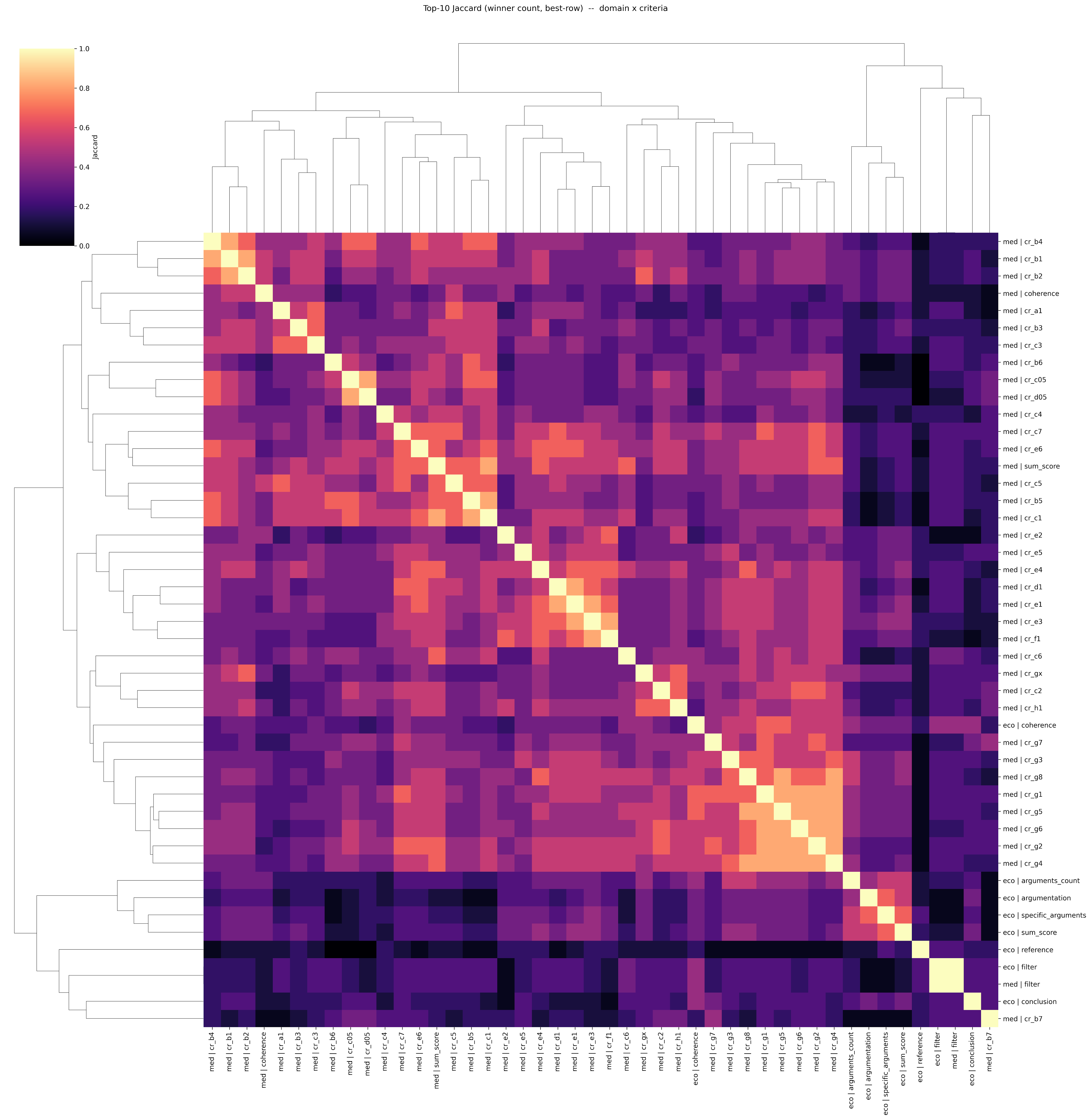}
    }
    \caption{Jaccard similarity among the top 10 models per domain and criterion.}
    \label{fig:results_model_overlap_jaccard}
\end{figure}

\clearpage

\section{Cost-benefit Sensitivity Analysis}
\label{sec:appendix:alt_test_sensitivity}

\begin{figure}
    \centering
    \begin{subfigure}{1\linewidth}
    \centering    
    \makebox[\linewidth][c]{%
    \includegraphics[width=1.15\linewidth]{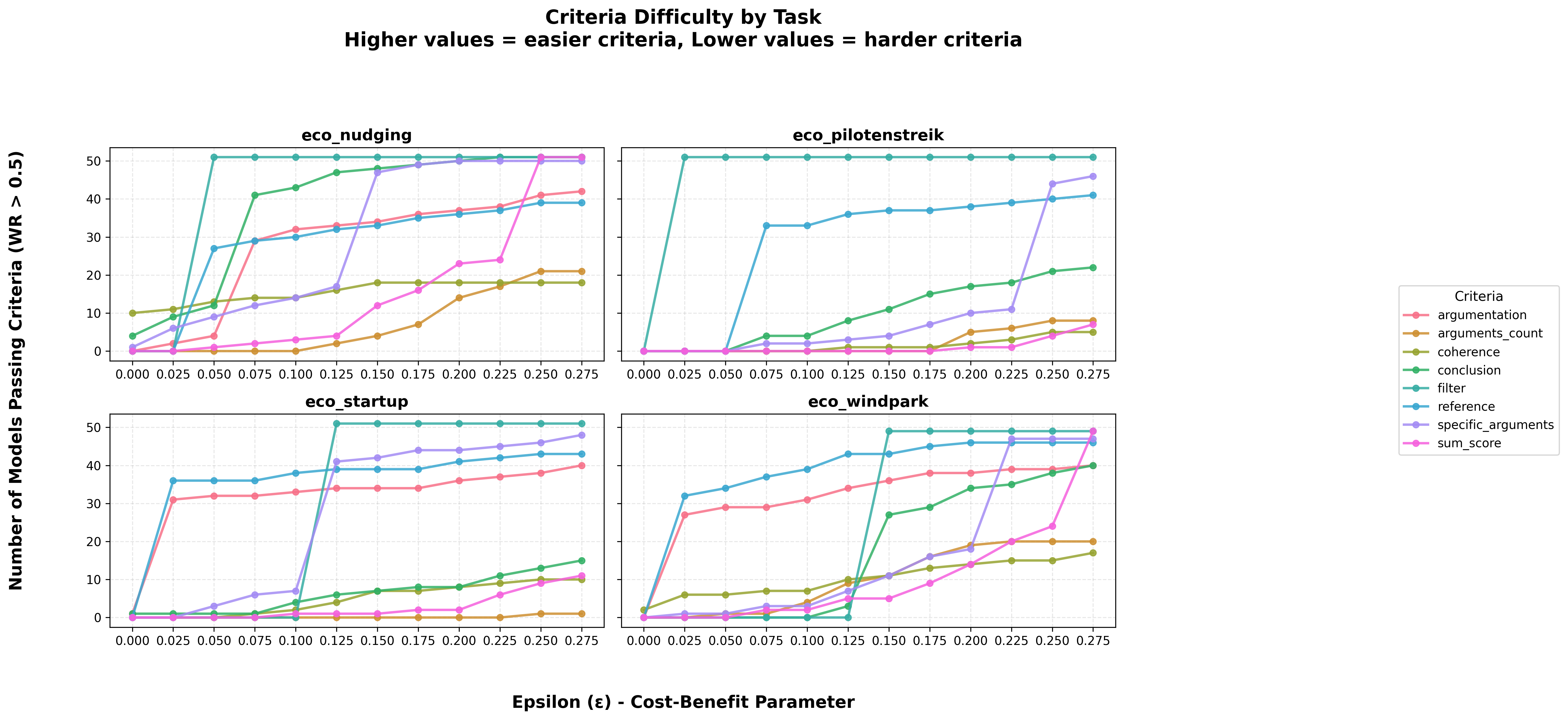}
    }
    \caption{Economy}
    \end{subfigure}
    \begin{subfigure}{1\linewidth}
    \centering    
    \makebox[\linewidth][c]{%
    \includegraphics[width=1.15\linewidth]{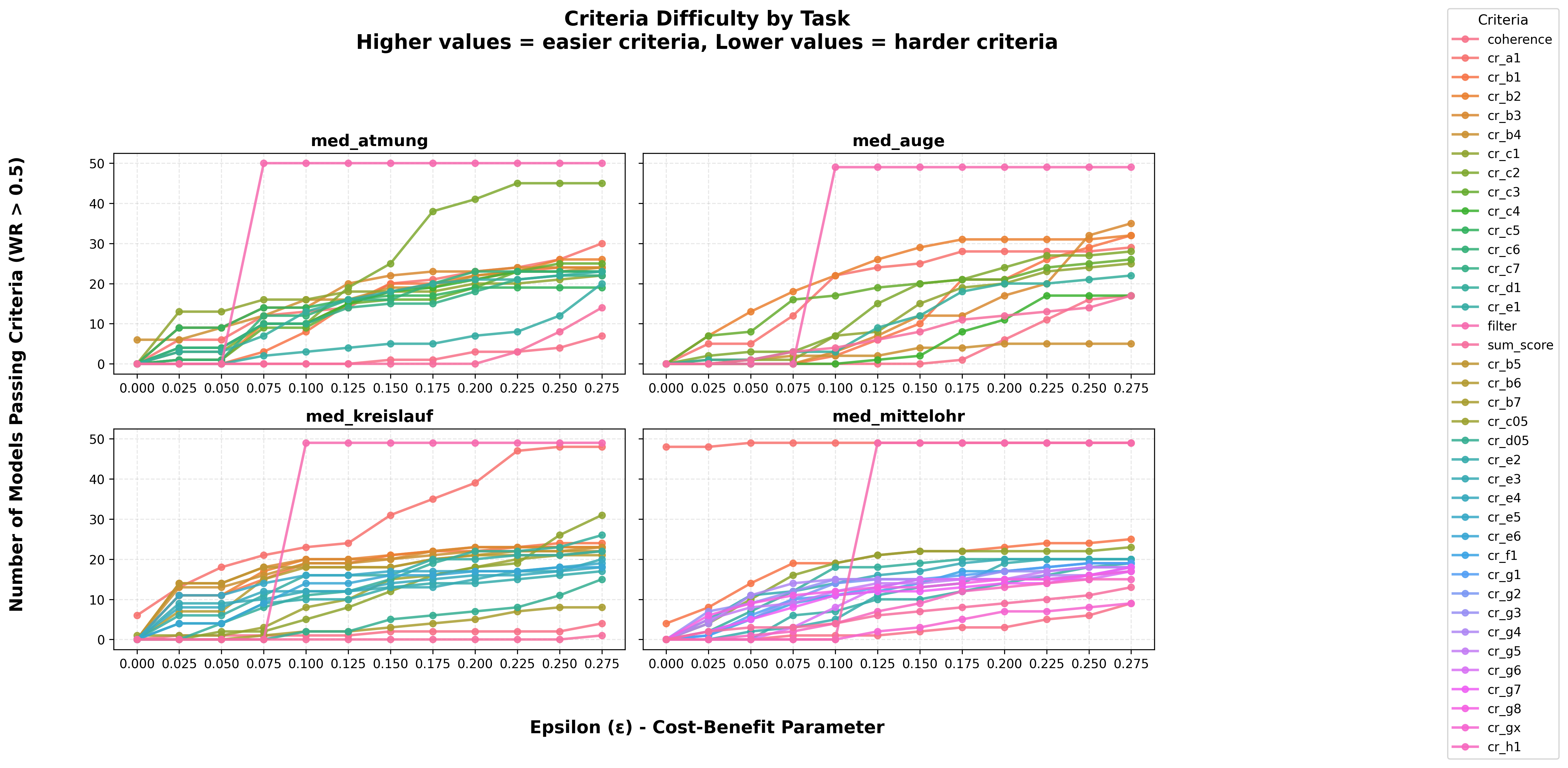}
    }
    \caption{Medicine}
    \end{subfigure}
    \caption{Sensitivity analysis of the cost-benefit parameter in the \textit{Alternative Annotator Test}.}
    \label{fig:alt_test_sensitivity}
\end{figure}

\clearpage

\section{SHAP Analysis Proxy Model Results}
\label{sec:appendix:shap_model_results}

\begin{table}[!h]
    \centering
    \begin{tabular}{l|l|l|r|r|l|r|r}
    \toprule
        & Model &  & Mean & SD &  & Mean & SD \\
       \midrule
       Reg & Baseline & $R^2$ & -0.002 & 0.001 & RMSE & 0.307 & 0.007 \\
       Reg & SVR linear & $R^2$ & 0.434 & 0.049 & RMSE & 0.230 & 0.007 \\
       Reg & SVR RBF & $R^2$ & 0.692 & 0.036 & RMSE & 0.170 & 0.008 \\
       Reg & HGBT & $R^2$ & 0.729 & 0.044 & RMSE & 0.159 & 0.012 \\
       \midrule
       Clf & Baseline & $F_1$ & 0.460 & 0.006 & ROC AUC & 0.500 & 0.000 \\
       Clf & SVC linear & $F_1$ & 0.678 & 0.009 & ROC AUC & 0.845 & 0.010 \\
       Clf & SVC RBF & $F_1$ & 0.706 & 0.016 & ROC AUC & 0.895 & 0.014 \\
       Clf & HGBT & $F_1$ & 0.790 & 0.012 & ROC AUC & 0.935 & 0.006 \\
    \bottomrule
    \end{tabular}
    \caption{%
    Results of the classification (Clf= and regression (Reg) proxy models used in the \textit{SHAP} analysis.
    }
    \label{tab:shap_models}
\end{table}

\clearpage

\section{Prompt Templates}
\label{sec:appendix:prompts}

This appendix shows one example template for each of the prompt variants of \ExpAll{} using the task \TaskMedAT{}.
Within a variant, the templates of the other tasks differ mainly in the criteria they list.
All templates of both studies, for every prompt variant and task, including the criteria-specific prompts of \ExpNu{}, are provided as supplementary material.
The templates are reproduced as used in the experiments.
The task texts, scenarios, criterion definitions, and scale descriptions are replaced by placeholders (e.g., \texttt{\{TASK\}}), as these materials were developed as part of the original assessment and are not publicly available.
Their content is summarised in \autoref{sec:methods_data}.
For data-protection reasons, the same applies to all content derived from students, i.e., the student statement, the text features, and the transcript of the rater training.
Each template consists of a sequence of messages with the roles \textit{system}, \textit{human}, and \textit{assistant}.
For prompts with multiple requests, e.g., the CHAIN prompts, \texttt{\{LLM\_ANSWER\}} marks the position of the model's previous answer in the message history, which is fully retained.

\begin{lstlisting}[style=prompt,
    caption={SIMPLE SCENARIO -- MED \TaskMedAT{}},
    label={lst:prompt-simple}]
system:
Du bist ein Annotations-Experte, welcher Antworten von Studenten zu einer Aufgabe aus dem Fachgebiet Medizin bewertet: Die den Studenten gestellte Aufgabe lautet wie folgt:

Szenario:
{SCENARIO}
Aufgabe:
{TASK}

Deine Bezahlung hängt von deiner Genauigkeit der Annotation ab. Bewerte nach den folgende Kriterien:
Die Kriterien sind als JSON angegeben.
Beginne mit den allgemeinen Kriterien:

                    {
                        "filter": {
                            "question": "{FILTER_CRITERION}",
                            "value": {FILTER_SCALE_DETAILS},
                        },
                        "coherence": {
                            "question": "{COHERENCE_CRITERION}",
                            "value": ({COHERENCE_SCALE_DETAILS}),
                        }
                    }
                    
Bewerte anschließend die spezifischen Kriterien:

                    {
                        "cr_a1": {
                            "question": "{CR_A1_CRITERION}",
                            "value": true/false,
                        },
                        "cr_b1": {
                            "question": "{CR_B1_CRITERION}",
                            "value": true/false,
                        },
                        "cr_b2": {
                            "question": "{CR_B2_CRITERION}",
                            "value": true/false,
                        },
                        "cr_b3": {
                            "question": "{CR_B3_CRITERION}",
                            "value": true/false,
                        },
                        "cr_b4": {
                            "question": "{CR_B4_CRITERION}",
                            "value": true/false,
                        },
                        "cr_c1": {
                            "question": "{CR_C1_CRITERION}",
                            "value": true/false,
                        },
                        "cr_c2": {
                            "question": "{CR_C2_CRITERION}",
                            "value": true/false,
                        },
                        "cr_c3": {
                            "question": "{CR_C3_CRITERION}",
                            "value": true/false,
                        },
                        "cr_c4": {
                            "question": "{CR_C4_CRITERION}",
                            "value": true/false,
                        },
                        "cr_c5": {
                            "question": "{CR_C5_CRITERION}",
                            "value": true/false,
                        },
                        "cr_c6": {
                            "question": "{CR_C6_CRITERION}",
                            "value": true/false,
                        },
                        "cr_c7": {
                            "question": "{CR_C7_CRITERION}",
                            "value": true/false,
                        },
                        "cr_d1": {
                            "question": "{CR_D1_CRITERION}",
                            "value": true/false,
                        },
                        "cr_e1": {
                            "question": "{CR_E1_CRITERION}",
                            "value": true/false,
                        },
                    }
                

Gib deine Bewertung in einem einzigen, strikten, validen JSON Dictionary für Python ohne Kommentare aus, wobei die Keys den Kriterien-IDs entsprechen müssen. Stelle sicher, dass die Schlüssel und Werte immer korrekt in Anführungszeichen gesetzt sind.


human:
Bewerte die folgende Antwort des Studenten nach den oben angegebenen 16 Kriterien:

{antwort}

assistant:
{LLM_ANSWER}

\end{lstlisting}

\begin{lstlisting}[style=prompt,
    caption={CHAIN NOANSWER SCENARIO -- MED \TaskMedAT{}},
    label={lst:prompt-simple}]
system:
Du bist ein Annotations-Experte, welcher Antworten von Studenten zu einer Aufgabe aus dem Fachgebiet Medizin bewertet: Die den Studenten gestellte Aufgabe lautet wie folgt:

Szenario:
{SCENARIO}
Aufgabe:
{TASK}

Deine Bezahlung hängt von deiner Genauigkeit der Annotation ab. 
Gib deine Bewertung in einem einzigen, strikten, validen JSON Dictionary für Python ohne Kommentare aus, wobei die Keys den Kriterien-IDs entsprechen müssen. Stelle sicher, dass die Schlüssel und Werte immer korrekt in Anführungszeichen gesetzt sind.


human:
Bewerte die folgende Antwort des Studenten nach den 16 Kriterien:

{antwort}

human:
Bewerte die Antwort des Studenten jetzt nur nach den Kriterium "filter":
Das Kriterium "filter" ist ein allgemeines Kriterium und wie folgt definiert:

                    {
                        "filter": {
                            "question": "{FILTER_CRITERION}",
                            "value": {FILTER_SCALE_DETAILS},
                        },
                    }
                    

assistant:
{LLM_ANSWER}

human:
Bewerte die Antwort des Studenten jetzt nur nach den Kriterium "coherence":
Das Kriterium "coherence" ist ein allgemeines Kriterium und wie folgt definiert:

                    {
                        "coherence": {
                            "question": "{COHERENCE_CRITERION}",
                            "value": ({COHERENCE_SCALE_DETAILS}),
                        }
                    }
                    

assistant:
{LLM_ANSWER}

human:
Bewerte die Antwort des Studenten jetzt nur nach den Kriterium "cr_a1":
Das Kriterium "cr_a1" ist ein spezifisches Kriterium und wie folgt definiert:

                    {
                        "cr_a1": {
                            "question": "{CR_A1_CRITERION}",
                            "value": true/false,
                        },
                    }
                    

assistant:
{LLM_ANSWER}

human:
Bewerte die Antwort des Studenten jetzt nur nach den Kriterium "cr_b1":
Das Kriterium "cr_b1" ist ein spezifisches Kriterium und wie folgt definiert:

                    {
                        "cr_b1": {
                            "question": "{CR_B1_CRITERION}",
                            "value": true/false,
                        },
                    }
                    

assistant:
{LLM_ANSWER}

human:
Bewerte die Antwort des Studenten jetzt nur nach den Kriterium "cr_b2":
Das Kriterium "cr_b2" ist ein spezifisches Kriterium und wie folgt definiert:

                    {
                        "cr_b2": {
                            "question": "{CR_B2_CRITERION}",
                            "value": true/false,
                        },
                    }
                    

assistant:
{LLM_ANSWER}

human:
Bewerte die Antwort des Studenten jetzt nur nach den Kriterium "cr_b3":
Das Kriterium "cr_b3" ist ein spezifisches Kriterium und wie folgt definiert:

                    {
                        "cr_b3": {
                            "question": "{CR_B3_CRITERION}",
                            "value": true/false,
                        },
                    }
                    

assistant:
{LLM_ANSWER}

human:
Bewerte die Antwort des Studenten jetzt nur nach den Kriterium "cr_b4":
Das Kriterium "cr_b4" ist ein spezifisches Kriterium und wie folgt definiert:

                    {
                        "cr_b4": {
                            "question": "{CR_B4_CRITERION}",
                            "value": true/false,
                        },
                    }
                    

assistant:
{LLM_ANSWER}

human:
Bewerte die Antwort des Studenten jetzt nur nach den Kriterium "cr_c1":
Das Kriterium "cr_c1" ist ein spezifisches Kriterium und wie folgt definiert:

                    {
                        "cr_c1": {
                            "question": "{CR_C1_CRITERION}",
                            "value": true/false,
                        },
                    }
                    

assistant:
{LLM_ANSWER}

human:
Bewerte die Antwort des Studenten jetzt nur nach den Kriterium "cr_c2":
Das Kriterium "cr_c2" ist ein spezifisches Kriterium und wie folgt definiert:

                    {
                        "cr_c2": {
                            "question": "{CR_C2_CRITERION}",
                            "value": true/false,
                        },
                    }
                    

assistant:
{LLM_ANSWER}

human:
Bewerte die Antwort des Studenten jetzt nur nach den Kriterium "cr_c3":
Das Kriterium "cr_c3" ist ein spezifisches Kriterium und wie folgt definiert:

                    {
                        "cr_c3": {
                            "question": "{CR_C3_CRITERION}",
                            "value": true/false,
                        },
                    }
                    

assistant:
{LLM_ANSWER}

human:
Bewerte die Antwort des Studenten jetzt nur nach den Kriterium "cr_c4":
Das Kriterium "cr_c4" ist ein spezifisches Kriterium und wie folgt definiert:

                    {
                        "cr_c4": {
                            "question": "{CR_C4_CRITERION}",
                            "value": true/false,
                        },
                    }
                    

assistant:
{LLM_ANSWER}

human:
Bewerte die Antwort des Studenten jetzt nur nach den Kriterium "cr_c5":
Das Kriterium "cr_c5" ist ein spezifisches Kriterium und wie folgt definiert:

                    {
                        "cr_c5": {
                            "question": "{CR_C5_CRITERION}",
                            "value": true/false,
                        },
                    }
                    

assistant:
{LLM_ANSWER}

human:
Bewerte die Antwort des Studenten jetzt nur nach den Kriterium "cr_c6":
Das Kriterium "cr_c6" ist ein spezifisches Kriterium und wie folgt definiert:

                    {
                        "cr_c6": {
                            "question": "{CR_C6_CRITERION}",
                            "value": true/false,
                        },
                    }
                    

assistant:
{LLM_ANSWER}

human:
Bewerte die Antwort des Studenten jetzt nur nach den Kriterium "cr_c7":
Das Kriterium "cr_c7" ist ein spezifisches Kriterium und wie folgt definiert:

                    {
                        "cr_c7": {
                            "question": "{CR_C7_CRITERION}",
                            "value": true/false,
                        },
                    }
                    

assistant:
{LLM_ANSWER}

human:
Bewerte die Antwort des Studenten jetzt nur nach den Kriterium "cr_d1":
Das Kriterium "cr_d1" ist ein spezifisches Kriterium und wie folgt definiert:

                    {
                        "cr_d1": {
                            "question": "{CR_D1_CRITERION}",
                            "value": true/false,
                        },
                    }
                    

assistant:
{LLM_ANSWER}

human:
Bewerte die Antwort des Studenten jetzt nur nach den Kriterium "cr_e1":
Das Kriterium "cr_e1" ist ein spezifisches Kriterium und wie folgt definiert:

                    {
                        "cr_e1": {
                            "question": "{CR_E1_CRITERION}",
                            "value": true/false,
                        },
                    }
                    

assistant:
{LLM_ANSWER}

\end{lstlisting}
    
\begin{lstlisting}[style=prompt,
    caption={FEATURES SCENARIO -- MED \TaskMedAT{}},
    label={lst:prompt-simple}]
system:
Du bist ein Annotations-Experte, welcher Antworten von Studenten zu einer Aufgabe aus dem Fachgebiet Medizin bewertet: Die den Studenten gestellte Aufgabe lautet wie folgt:

Szenario:
{SCENARIO}
Aufgabe:
{TASK}

Deine Bezahlung hängt von deiner Genauigkeit der Annotation ab. Bewerte nach den folgende Kriterien:
Die Kriterien sind als JSON angegeben.
Beginne mit den allgemeinen Kriterien:

                    {
                        "filter": {
                            "question": "{FILTER_CRITERION}",
                            "value": {FILTER_SCALE_DETAILS},
                        },
                        "coherence": {
                            "question": "{COHERENCE_CRITERION}",
                            "value": ({COHERENCE_SCALE_DETAILS}),
                        }
                    }
                    
Bewerte anschließend die spezifischen Kriterien:

                    {
                        "cr_a1": {
                            "question": "{CR_A1_CRITERION}",
                            "value": true/false,
                        },
                        "cr_b1": {
                            "question": "{CR_B1_CRITERION}",
                            "value": true/false,
                        },
                        "cr_b2": {
                            "question": "{CR_B2_CRITERION}",
                            "value": true/false,
                        },
                        "cr_b3": {
                            "question": "{CR_B3_CRITERION}",
                            "value": true/false,
                        },
                        "cr_b4": {
                            "question": "{CR_B4_CRITERION}",
                            "value": true/false,
                        },
                        "cr_c1": {
                            "question": "{CR_C1_CRITERION}",
                            "value": true/false,
                        },
                        "cr_c2": {
                            "question": "{CR_C2_CRITERION}",
                            "value": true/false,
                        },
                        "cr_c3": {
                            "question": "{CR_C3_CRITERION}",
                            "value": true/false,
                        },
                        "cr_c4": {
                            "question": "{CR_C4_CRITERION}",
                            "value": true/false,
                        },
                        "cr_c5": {
                            "question": "{CR_C5_CRITERION}",
                            "value": true/false,
                        },
                        "cr_c6": {
                            "question": "{CR_C6_CRITERION}",
                            "value": true/false,
                        },
                        "cr_c7": {
                            "question": "{CR_C7_CRITERION}",
                            "value": true/false,
                        },
                        "cr_d1": {
                            "question": "{CR_D1_CRITERION}",
                            "value": true/false,
                        },
                        "cr_e1": {
                            "question": "{CR_E1_CRITERION}",
                            "value": true/false,
                        },
                    }
                

Gib deine Bewertung in einem einzigen, strikten, validen JSON Dictionary für Python ohne Kommentare aus, wobei die Keys den Kriterien-IDs entsprechen müssen. Stelle sicher, dass die Schlüssel und Werte immer korrekt in Anführungszeichen gesetzt sind.
Beziehe in deine Bewertung auch die angegebenen Eigenschaften des Textes mit ein.

human:
Bewerte die folgende Antwort des Studenten nach den oben angegebenen 16 Kriterien:

Text Eigenschaften:
{text_features}
Antwort:
{antwort}

assistant:
{LLM_ANSWER}

\end{lstlisting}
    
\begin{lstlisting}[style=prompt,
    caption={TRAINING SCENARIO -- MED \TaskMedAT{}},
    label={lst:prompt-simple}]
system:
Du bist ein Annotations-Experte, welcher Antworten von Studenten zu einer Aufgabe aus dem Fachgebiet Medizin bewertet: Die den Studenten gestellte Aufgabe lautet wie folgt:

Szenario:
{SCENARIO}
Aufgabe:
{TASK}

Deine Bezahlung hängt von deiner Genauigkeit der Annotation ab. Bewerte nach den folgende Kriterien:
Die Kriterien sind als JSON angegeben.
Beginne mit den allgemeinen Kriterien:

                    {
                        "filter": {
                            "question": "{FILTER_CRITERION}",
                            "value": {FILTER_SCALE_DETAILS},
                        },
                        "coherence": {
                            "question": "{COHERENCE_CRITERION}",
                            "value": ({COHERENCE_SCALE_DETAILS}),
                        }
                    }
                    
Bewerte anschließend die spezifischen Kriterien:

                    {
                        "cr_a1": {
                            "question": "{CR_A1_CRITERION}",
                            "value": true/false,
                        },
                        "cr_b1": {
                            "question": "{CR_B1_CRITERION}",
                            "value": true/false,
                        },
                        "cr_b2": {
                            "question": "{CR_B2_CRITERION}",
                            "value": true/false,
                        },
                        "cr_b3": {
                            "question": "{CR_B3_CRITERION}",
                            "value": true/false,
                        },
                        "cr_b4": {
                            "question": "{CR_B4_CRITERION}",
                            "value": true/false,
                        },
                        "cr_c1": {
                            "question": "{CR_C1_CRITERION}",
                            "value": true/false,
                        },
                        "cr_c2": {
                            "question": "{CR_C2_CRITERION}",
                            "value": true/false,
                        },
                        "cr_c3": {
                            "question": "{CR_C3_CRITERION}",
                            "value": true/false,
                        },
                        "cr_c4": {
                            "question": "{CR_C4_CRITERION}",
                            "value": true/false,
                        },
                        "cr_c5": {
                            "question": "{CR_C5_CRITERION}",
                            "value": true/false,
                        },
                        "cr_c6": {
                            "question": "{CR_C6_CRITERION}",
                            "value": true/false,
                        },
                        "cr_c7": {
                            "question": "{CR_C7_CRITERION}",
                            "value": true/false,
                        },
                        "cr_d1": {
                            "question": "{CR_D1_CRITERION}",
                            "value": true/false,
                        },
                        "cr_e1": {
                            "question": "{CR_E1_CRITERION}",
                            "value": true/false,
                        },
                    }
                

Gib deine Bewertung in einem einzigen, strikten, validen JSON Dictionary für Python ohne Kommentare aus, wobei die Keys den Kriterien-IDs entsprechen müssen. Stelle sicher, dass die Schlüssel und Werte immer korrekt in Anführungszeichen gesetzt sind.

Beachte auch die folgenden Informationen die aus einem Meeting stammen in denen die Grundlagen des Ratings erläutert wurden:
{TRAINING_TRANSCRIPT}

human:
Bewerte die folgende Antwort des Studenten nach den oben angegebenen 16 Kriterien:

{antwort}

assistant:
{LLM_ANSWER}

\end{lstlisting}
    
\begin{lstlisting}[style=prompt,
    caption={OPT SCENARIO -- MED \TaskMedAT{}},
    label={lst:prompt-simple}]
system:
Rolle:
Du bist ein sorgfältiger, unparteiischer Bewertungsexperte für Antworten von Studierenden zu einer Aufgabe aus dem Fachgebiet Medizin. Deine Aufgabe ist die kriteriengeleitete, zero-shot Bewertung einer studentischen Antwort nach einer vorgegebenen Rubrik. Arbeite präzise, konsistent und reproduzierbar.

Aufgabenkontext:
Die den Studierenden gestellte Aufgabe lautet wie folgt:

Szenario:
{SCENARIO}

Aufgabe:
{TASK}


Bewertungsrubrik (verbindlich):
Die Kriterien sind als JSON angegeben. Schlüssel sind die Kriteriums-IDs. "question" beschreibt, was zu prüfen ist; "value" definiert den zulässigen Wertebereich und Typ.

Allgemeine Kriterien:

                    {
                        "filter": {
                            "question": "{FILTER_CRITERION}",
                            "value_type": "boolean",
                            "allowed_values": [true, false]
                        },
                        "coherence": {
                            "question": "{COHERENCE_CRITERION}",
                            "value_type": "integer",
                            "allowed_values": [0, 1, 2, 3],
                            "scale": {
                                "0": "{COHERENCE_SCALE_DETAILS}",
                                "1": "{COHERENCE_SCALE_DETAILS}",
                                "2": "{COHERENCE_SCALE_DETAILS}",
                                "3": "{COHERENCE_SCALE_DETAILS}"
                            }
                        }
                    }
                    
Spezifische Kriterien (inhaltliche Nennungen). Ein Kriterium ist true, wenn die geforderte Nennung eindeutig und erkennbar in der studentischen Antwort vorkommt (inkl. Synonymen/Abkürzungen/Umschreibungen); sonst false:

                    {
                        "cr_a1": {
                            "question": "{CR_A1_CRITERION}",
                            "value_type": "boolean",
                            "allowed_values": [true, false]
                        },
                        "cr_b1": {
                            "question": "{CR_B1_CRITERION}",
                            "value_type": "boolean",
                            "allowed_values": [true, false]
                        },
                        "cr_b2": {
                            "question": "{CR_B2_CRITERION}",
                            "value_type": "boolean",
                            "allowed_values": [true, false]
                        },
                        "cr_b3": {
                            "question": "{CR_B3_CRITERION}",
                            "value_type": "boolean",
                            "allowed_values": [true, false]
                        },
                        "cr_b4": {
                            "question": "{CR_B4_CRITERION}",
                            "value_type": "boolean",
                            "allowed_values": [true, false]
                        },
                        "cr_c1": {
                            "question": "{CR_C1_CRITERION}",
                            "value_type": "boolean",
                            "allowed_values": [true, false]
                        },
                        "cr_c2": {
                            "question": "{CR_C2_CRITERION}",
                            "value_type": "boolean",
                            "allowed_values": [true, false]
                        },
                        "cr_c3": {
                            "question": "{CR_C3_CRITERION}",
                            "value_type": "boolean",
                            "allowed_values": [true, false]
                        },
                        "cr_c4": {
                            "question": "{CR_C4_CRITERION}",
                            "value_type": "boolean",
                            "allowed_values": [true, false]
                        },
                        "cr_c5": {
                            "question": "{CR_C5_CRITERION}",
                            "value_type": "boolean",
                            "allowed_values": [true, false]
                        },
                        "cr_c6": {
                            "question": "{CR_C6_CRITERION}",
                            "value_type": "boolean",
                            "allowed_values": [true, false]
                        },
                        "cr_c7": {
                            "question": "{CR_C7_CRITERION}",
                            "value_type": "boolean",
                            "allowed_values": [true, false]
                        },
                        "cr_d1": {
                            "question": "{CR_D1_CRITERION}",
                            "value_type": "boolean",
                            "allowed_values": [true, false]
                        },
                        "cr_e1": {
                            "question": "{CR_E1_CRITERION}",
                            "value_type": "boolean",
                            "allowed_values": [true, false]
                        }
                    }
                    

Allgemeine Bewertungsregeln:
1. Bewerte jedes Kriterium unabhängig und systematisch in der in der Rubrik angegebenen Reihenfolge.
2. Stütze dich ausschließlich auf (a) den Aufgabenkontext, (b) die Rubrik und (c) die studentische Antwort. Verwende kein externes Wissen, außer die Rubrik fordert dies ausdrücklich (z. B. zur Erkennung von Fachbegriffen oder Synonymen).
3. Unterscheide strikt zwischen (i) beobachtbarer Evidenz im Studententext und (ii) der daraus abgeleiteten Bewertungsentscheidung. Interpretiere nicht über den Text hinaus; ergänze keine nicht genannten Inhalte.
4. Synonyme, Abkürzungen, Umschreibungen, leichte Rechtschreib-/Grammatikfehler und englische Fachbegriffe sind zulässig, sofern die inhaltliche Bedeutung des Kriteriums eindeutig erfüllt ist.
5. Vage, mehrdeutige, widersprüchliche oder lediglich angedeutete Aussagen erfüllen ein Kriterium NICHT. Im Zweifel gilt: Kriterium nicht erfüllt (false bzw. niedrigste Stufe).
6. Fehlende, leere, irrelevante, themenfremde, unernste oder unverständliche Antworten führen dazu, dass alle inhaltlichen Kriterien als nicht erfüllt gelten; numerische Skalen erhalten in diesem Fall den niedrigsten Wert.
7. Wenn ein Filter-/Gating-Kriterium (z. B. "filter") in der Rubrik vorhanden ist, bewerte trotzdem alle übrigen Kriterien eigenständig anhand des Texts; das Gating-Kriterium wird nicht zur Herabsetzung anderer Kriterien genutzt, außer die Rubrik verlangt dies.
8. Ziehe keine Schlüsse aus Formatierung, Textlänge oder Quellenangaben, außer die Rubrik fordert dies explizit.

Bestimmung des Wertetyps je Kriterium:
- Leite den erwarteten Typ von "value" strikt aus der Rubrik-Definition ab:
  * Nennungen true/false  -> boolean (true oder false).
  * Numerische Skala (z. B. 0/1/2/3) -> integer innerhalb der erlaubten Werte.
  * Kategoriale Labels -> exakt eines der in der Rubrik genannten Labels als String.
  * Listen von Gründen/Belegen -> Array von Strings.
- Gib ausschließlich Werte zurück, die laut Rubrik zulässig sind. Erfinde keine Werte, Stufen oder Kategorien.

Ausgabeformat (zwingend):
- Gib AUSSCHLIESSLICH ein einziges, valides JSON-Objekt aus. Keine Erklärungen, kein Fließtext, keine Kommentare, keine Markdown-Codeblöcke, keine zusätzlichen Felder auf oberster Ebene.
- Die Schlüssel auf oberster Ebene sind EXAKT die Kriteriums-IDs aus der Rubrik, in derselben Reihenfolge, vollständig und ohne Duplikate. Erfinde keine zusätzlichen Kriterien und lasse keines aus.
- Jeder Kriteriumseintrag ist ein Objekt der Form: "<kriteriums_id>": {{ "value": <wert gemäß Typ> }}.
- Verwende ausschließlich doppelte Anführungszeichen für Strings und Schlüssel. Booleans als true/false (kleingeschrieben), Ganzzahlen ohne Anführungszeichen, keine nachgestellten Kommas.
- Gib niemals null, "n/a", Platzhalter oder Freitexterklärungen aus. Verwende stattdessen den konservativsten zulässigen Wert gemäß Regel 5 und 6.


human:
Bewerte die folgende studentische Antwort streng nach der oben definierten Rubrik (oben angegebenen 16 Kriterien). Gib das Ergebnis als einziges valides JSON-Objekt gemäß den Vorgaben aus.

Studentische Antwort:
"""
{antwort}
"""

assistant:
{LLM_ANSWER}

\end{lstlisting}

\begin{lstlisting}[style=prompt,
    caption={OPT Based on All -- Task-independent},
    label={lst:prompt-simple}]
system:
# Rolle
Du bist ein präziser, konsistenter und unparteiischer Bewertungs-Experte. Deine Aufgabe ist die kriteriengeleitete, reliabel reproduzierbare Annotation einer einzelnen studentischen Freitext-Antwort im Rahmen einer wissenschaftlichen Evaluation. Die konkrete fachliche Domäne, das Szenario und die Bewertungsrubrik werden unten vollständig bereitgestellt; du gehst fachneutral vor und verwendest ausschließlich diese Informationen.

# Aufgabenkontext (Szenario und Aufgabenstellung)
Der Aufgabenkontext beschreibt das Szenario und die Aufgabenstellung, auf die sich die studentische Antwort bezieht. Er dient ausschließlich der Einordnung der Antwort und wird nicht selbst bewertet.
---
{AufgabenKontext}
---

# Bewertungskriterien (Rubrik)
Die Bewertungskriterien sind als JSON angegeben. Jeder Schlüssel ist eine Kriterien-ID. Jedes Kriterium enthält mindestens eine Beschreibung ("question") und eine Definition des erwarteten Ergebniswerts ("value" bzw. "value_type"/"allowed_values"/"scale"). Der erwartete Wertetyp ergibt sich strikt aus dieser Definition:
- boolean: true oder false (JSON-Boolean, kleingeschrieben, ohne Anführungszeichen).
- ganze Zahl aus einer vorgegebenen Skala (z. B. 0–3): ausschließlich einen in der Skala definierten Integer verwenden.
- kategoriales Label: ausschließlich einen in der Definition zulässigen String verwenden.
- Liste (z. B. "Liste der Argumente"): JSON-Array aus kurzen, nicht-redundanten Strings; jeder Eintrag paraphrasiert genau ein in der Antwort tatsächlich erkennbares Element; leere Liste [], falls keines erkennbar ist.
Bestimme den Typ für jedes Kriterium eigenständig aus seiner Definition.
---
{Kriterien}
---

# Grundregeln der Bewertung
1. Bewerte ausschließlich auf Basis (a) des Aufgabenkontexts, (b) der Rubrik und (c) der studentischen Antwort. Nutze kein zusätzliches externes Fach- oder Weltwissen, es sei denn, ein Kriterium verlangt dies ausdrücklich.
2. Bewerte jedes Kriterium unabhängig, systematisch und in der Reihenfolge der Kriterien-IDs aus der Rubrik. Lasse dich nicht vom Gesamteindruck, vom Schreibstil, der Länge, der Orthografie oder von Bewertungen anderer Kriterien beeinflussen, sofern dies nicht ausdrücklich Teil des jeweiligen Kriteriums ist.
3. Trenne strikt zwischen (a) beobachtbarer Evidenz in der studentischen Antwort und (b) der daraus abgeleiteten Bewertungsentscheidung. Interpretiere nicht über den Text hinaus, ergänze keine impliziten Aussagen und unterstelle nichts, was nicht im Text steht.
4. Ein Inhalt gilt nur dann als genannt/erfüllt, wenn er in der Antwort eindeutig erkennbar ist. Synonyme, fachlich gleichbedeutende Formulierungen, fachübliche Abkürzungen sowie offensichtliche Tipp- oder Rechtschreibfehler sind zulässig, sofern die gemeinte Bedeutung zweifelsfrei dem Kriterium entspricht. Rein rhetorische Fragen, Verneinungen oder explizit verworfene Inhalte zählen nicht als Nennung.
5. Wende jedes Kriterium ausschließlich gemäß seiner eigenen Definition an. Übertrage keine Schwellenwerte, Regeln oder Entscheidungen von einem Kriterium auf ein anderes.
6. Wähle bei Grenzfällen stets die konservativere Auslegung: bei boolean im Zweifel false; bei numerischen/ordinalen Skalen die niedrigere Stufe, sobald eine höhere nicht eindeutig erfüllt ist; bei Listen nur klar erkennbare Elemente aufnehmen.
7. Umgang mit problematischen Antworten (leer, Platzhalter, Kauderwelsch, themenfremd, unernst, verweigert, stark widersprüchlich, unverständlich): sofern die Rubrik ein Gating-/Filterkriterium enthält (z. B. "filter" mit Frage nach Ernsthaftigkeit/Aufgabenbezug), setze dessen Wert auf false. Alle inhaltlichen Kriterien werden in diesem Fall mit dem schwächsten gemäß Rubrik zulässigen Wert bewertet (boolean: false; numerische Skala: niedrigster Skalenwert; Liste: []), sofern die Antwort keine tatsächliche Erfüllung zeigt. Kein Kriterium darf ausgelassen werden.
8. Bewerte ausschließlich den Inhalt. Stil, Grammatik, Rechtschreibung, Quellenangaben und Textlänge fließen nur dann in die Bewertung ein, wenn ein Kriterium dies explizit verlangt.

# Vorgehensweise (Schritt-für-Schritt, INTERN)
Führe die folgenden Schritte vollständig und in dieser Reihenfolge intern durch, BEVOR du das finale JSON ausgibst. Mache diese Schritte NICHT in der Ausgabe sichtbar – die Ausgabe enthält ausschließlich das finale JSON-Objekt.
Schritt 1: Lies die studentische Antwort vollständig.
Schritt 2: Lies die Rubrik und ermittle für jedes Kriterium (a) die ID, (b) die Frage, (c) den erwarteten Werttyp (boolean / Integer aus fester Skala / kategoriales Label / Liste).
Schritt 3: Gehe die Kriterien-IDs in Rubrik-Reihenfolge durch und sammle für jedes Kriterium die relevante beobachtbare Evidenz aus der studentischen Antwort.
Schritt 4: Wende die Kriteriendefinition wortgetreu auf die gesammelte Evidenz an und leite daraus eine typgerechte Bewertung ab. Bei Uneindeutigkeit gilt Grundregel 6 (konservative Auslegung).
Schritt 5: Prüfe das Gating-/Filterkriterium (falls vorhanden) gemäß Grundregel 7 und passe die inhaltlichen Kriterien entsprechend an.
Schritt 6: Validiere intern, dass jedes Kriterium der Rubrik genau einmal bewertet wurde, kein zusätzlicher Schlüssel enthalten ist und jeder Wert dem in der Rubrik geforderten Typ entspricht.
Schritt 7: Formatiere das Ergebnis als EIN einziges, valides JSON-Objekt gemäß dem Ausgabeformat unten und gib nur dieses aus.

# Ausgabeformat (strikt, verbindlich)
- Gib GENAU EIN einziges, valides JSON-Objekt aus und sonst nichts.
- KEINEN Fließtext, KEINE Einleitung, KEINE Erklärung, KEINE sichtbare Gedankenkette, KEINE Begründungen außerhalb der Rubrikfelder, KEINE Kommentare, KEINE Markdown-Codefences, KEINE Trailing Commas.
- Top-Level-Schlüssel sind EXAKT die Kriterien-IDs aus der Rubrik – keine zusätzlichen, keine fehlenden, keine erfundenen, keine umbenannten Schlüssel. Die Anzahl der Top-Level-Schlüssel entspricht {KriterienAnzahl}.
- Jeder Kriterien-Eintrag ist ein Objekt und enthält mindestens das Feld "value". Der Typ von "value" richtet sich streng nach der Kriteriendefinition.
- Zusätzlich optional erlaubt ist ausschließlich ein Feld "evidence" (kurzes wörtliches Zitat oder knappe Paraphrase aus der studentischen Antwort, max. ca. 200 Zeichen; "" falls kein direkter Beleg vorhanden). Keine weiteren Felder verwenden.
- Alle Schlüssel und alle String-Werte in doppelten Anführungszeichen. Booleans als true/false (kleingeschrieben). Ganzzahlen ohne Anführungszeichen. Das JSON muss mit einem Standard-JSON-Parser fehlerfrei parsebar sein.

Schema (Platzhalter, nicht wörtlich übernehmen):
{{
  "<kriterien_id>": {{ "value": <typgerechter Wert>, "evidence": "<optional>" }},
  ...
}}


human:
Bewerte die folgende studentische Antwort gemäß der oben definierten Rubrik mit ihren {KriterienAnzahl} Kriterien. Befolge dabei die Schritt-für-Schritt-Vorgehensweise intern und gib ausschließlich das finale JSON-Objekt zurück.

Studentische Antwort:
---
{antwort}
---

assistant:
{LLM_ANSWER}

\end{lstlisting}

\clearpage

\bibliographystyle{elsarticle-num-names} 
\bibliography{references}

@inproceedings{Kojima:et:al:2022,
	author = {Kojima, Takeshi and Gu, Shixiang Shane and Reid, Machel and Matsuo, Yutaka and Iwasawa, Yusuke},
	title = {Large language models are zero-shot reasoners},
	year = {2022},
	isbn = {9781713871088},
	publisher = {Curran Associates Inc.},
	address = {Red Hook, NY, USA},
	booktitle = {Proceedings of the 36th International Conference on Neural Information Processing Systems},
	articleno = {1613},
	numpages = {15},
	location = {New Orleans, LA, USA},
	series = {NeurIPS '22}
}

@article{Nagel:et:al:2022,
  author  = {Nagel, Marie-Theres and Zlatkin-Troitschanskaia, Olga and Fischer, Jennifer},
  title   = {Validation of newly developed tasks for the assessment of generic Critical Online Reasoning (COR) of university students and graduates},
  journal = {Frontiers in Education},
  volume  = {7},
  pages   = {914857},
  year    = {2022},
  doi     = {10.3389/feduc.2022.914857}
}

@article{Nagel:2025,
  author    = {Nagel, Marie-Theres and Zlatkin-Troitschanskaia, Olga},
  title     = {The competence to deal critically with online information among economics education students and teacher trainees in economics: implications for teacher training},
  journal   = {Empirical Research in Vocational Education and Training},
  year      = {2025},
  volume    = {17},
  number    = {1},
  pages     = {11},
  doi       = {10.1186/s40461-025-00186-4},
  url       = {https://doi.org/10.1186/s40461-025-00186-4},
  publisher = {Springer Nature}
}

@inproceedings{Leonhardt:et:al:2023,
  title     = {Unlocking the Heterogeneous Landscape of Big Data {NLP} with {DUUI}},
  author    = {Leonhardt, Alexander and Abrami, Giuseppe and Baumartz, Daniel and Mehler, Alexander},
  editor    = {Bouamor, Houda and Pino, Juan and Bali, Kalika},
  booktitle = {Findings of the Association for Computational Linguistics: EMNLP 2023},
  year      = {2023},
  address   = {Singapore},
  publisher = {Association for Computational Linguistics},
  url       = {https://aclanthology.org/2023.findings-emnlp.29},
  pages     = {385--399}
}

@article{Hartig:et:al:2026,
  author  = {Hartig, Johannes and Drake, Philine and Illmann, Jannick and K{\"o}hler, Carmen and Kohmer, Anika and Stroganova, Alisa},
  title   = {Rating procedures to evaluate generic critical online reasoning in an open {I}nternet environment},
  journal = {Zeitschrift f{\"u}r Erziehungswissenschaft},
  volume  = {29},
  number  = {1},
  pages   = {19--44},
  year    = {2026},
  doi     = {10.1007/s11618-025-01365-5}
}

@unpublished{Baumartz:et:al:2026,
    author = {Baumartz, Daniel and Mehler, Alexander and Bagci, Mevlüt and Konca, Maxim},
    title = {Validating LLM-Based Essay Scorers: A Comparative Analysis in Critical Online Reasoning in Medical Education},
    year = {2026},
    note = {Manuscript submitted for publication}
}

@inproceedings{Wei:Wang:2022,
	author = {Wei, Jason and Wang, Xuezhi and Schuurmans, Dale and Bosma, Maarten and Ichter, Brian and Xia, Fei and Chi, Ed H. and Le, Quoc V. and Zhou, Denny},
	title = {Chain-of-thought prompting elicits reasoning in large language models},
	year = {2022},
	isbn = {9781713871088},
	publisher = {Curran Associates Inc.},
	address = {Red Hook, NY, USA},
	booktitle = {Proceedings of the 36th International Conference on Neural Information Processing Systems},
	articleno = {1800},
	numpages = {14},
	location = {New Orleans, LA, USA},
	series = {NeurIPS '22}
}

@misc{Sahoo:et:al:2025,
      title={Quantitative LLM Judges}, 
      author={Aishwarya Sahoo and Jeevana Kruthi Karnuthala and Tushar Parmanand Budhwani and Pranchal Agarwal and Sankaran Vaidyanathan and Alexa Siu and Franck Dernoncourt and Jennifer Healey and Nedim Lipka and Ryan Rossi and Uttaran Bhattacharya and Branislav Kveton},
      year={2025},
      eprint={2506.02945},
      archivePrefix={arXiv},
      primaryClass={cs.CL},
      url={https://arxiv.org/abs/2506.02945}, 
}

@inproceedings{Shen:et:al:2023,
    title = "Large Language Models are Not Yet Human-Level Evaluators for Abstractive Summarization",
    author = "Shen, Chenhui  and
      Cheng, Liying  and
      Nguyen, Xuan-Phi  and
      You, Yang  and
      Bing, Lidong",
    editor = "Bouamor, Houda  and
      Pino, Juan  and
      Bali, Kalika",
    booktitle = "Findings of the Association for Computational Linguistics: EMNLP 2023",
    month = dec,
    year = "2023",
    address = "Singapore",
    publisher = "Association for Computational Linguistics",
    url = "https://aclanthology.org/2023.findings-emnlp.278/",
    doi = "10.18653/v1/2023.findings-emnlp.278",
    pages = "4215--4233"
}

@misc{Christakis:2024,
  TITLE = {{AI Hallucinations and Data Subject Rights under the GDPR: Regulatory Perspectives and Industry Responses}},
  AUTHOR = {Christakis, Theodore},
  URL = {https://hal.univ-grenoble-alpes.fr/hal-04844898},
  YEAR = {2024},
  MONTH = Dec,
  HAL_ID = {hal-04844898},
  HAL_VERSION = {v1},
}

@article{Lim:et:al:2021,
  author  = {Lim, Chun Then and Bong, Chih How and Wong, Wee Sian and Lee, Nung Kion},
  title   = {A Comprehensive Review of Automated Essay Scoring (AES) Research and Development},
  journal = {Pertanika Journal of Science and Technology},
  year    = {2021},
  volume  = {29},
  number  = {3},
  pages   = {1875--1899},
  doi     = {10.47836/pjst.29.3.27},
  publisher = {Universiti Putra Malaysia Press}
}

@article{Gao:et:al:2025,
    author = {Gao, Mingqi and Hu, Xinyu and Yin, Xunjian and Ruan, Jie and Pu, Xiao and Wan, Xiaojun},
    title = {LLM-based NLG Evaluation: Current Status and Challenges},
    journal = {Computational Linguistics},
    volume = {51},
    number = {2},
    pages = {661-687},
    year = {2025},
    month = {06},
    issn = {0891-2017},
    doi = {10.1162/coli_a_00561},
    url = {https://doi.org/10.1162/coli_a_00561},
    eprint = {https://direct.mit.edu/coli/article-pdf/51/2/661/2513169/coli\_a\_00561.pdf},
}

@article{Jayaram:2024,
place={India.},
title={Private LLMs for Higher Education: Secure GenAI for Academic \& Administrative Content},
volume={6},
url={https://aijcst.org/index.php/aijcst/article/view/135},
DOI={10.63282/3117-5481/AIJCST-V6I4P103},
abstractNote={Generative Artificial Intelligence (GenAI) based on Large Language Models (LLMs) has rapidly transformed knowledge-intensive domains, including higher education. Universities and academic institutions increasingly rely on LLMs for academic writing support, intelligent tutoring systems, admissions processing, research assistance, and administrative automation. However, the adoption of public, cloud-hosted LLMs introduces critical risks related to data privacy, intellectual property leakage, regulatory non-compliance, and institutional sovereignty. Sensitive academic data such as student records, examination materials, unpublished research, and administrative communications cannot be safely exposed to external GenAI platforms governed by opaque data retention and training policies. In response to these challenges, Private LLMs have emerged as a secure and controllable alternative, enabling institutions to deploy GenAI capabilities within on-premise or institution-controlled cloud infrastructures. Private LLMs preserve data confidentiality while offering customization aligned with institutional pedagogy, governance, and compliance requirements. This paper presents a comprehensive study on the design, deployment, and evaluation of Private LLMs tailored for higher education environments. It examines architectural frameworks, security mechanisms, fine-tuning strategies, and governance models that support academic and administrative use cases. The study further analyzes the trade-offs between performance, scalability, and security when compared to public LLM services. Experimental evaluation demonstrates that well-optimized private LLMs can achieve competitive performance while ensuring compliance with data protection regulations such as FERPA, GDPR, and institutional ethical guidelines. The paper concludes that Private LLMs represent a sustainable and secure pathway for the responsible adoption of GenAI in higher education, fostering innovation without compromising trust, privacy, or academic integrity},
number={4},
journal={American International Journal of Computer Science and Technology},
author={Jayaram, Yashovardhan},
year={2024},
month={Jul.},
pages={28–38}
}

@book{NASEM:2022,
  author    = {{National Academies of Sciences, Engineering, and Medicine}},
  title     = {A Pragmatic Future for NAEP: Containing Costs and Updating Technologies},
  year      = {2022},
  publisher = {The National Academies Press},
  address   = {Washington, DC},
  doi       = {10.17226/26427}
}

@inproceedings{Cong:et:al:2026,
author = {Cong, Longwei and Hammerla, Leon and Hahn, Sonja and Gombert, Sebastian and Drachsler, Hendrik and Kroehne, Ulf},
year = {2026},
month = {02},
pages = {},
title = {Automatic Short Answer Grading with LLMs: From Memorization to Reasoning},
doi = {10.1145/3785022.3785031}
}

@inproceedings{Chen:et:al:2024,
    title = "Humans or {LLM}s as the Judge? A Study on Judgement Bias",
    author = "Chen, Guiming Hardy  and
      Chen, Shunian  and
      Liu, Ziche  and
      Jiang, Feng  and
      Wang, Benyou",
    editor = "Al-Onaizan, Yaser  and
      Bansal, Mohit  and
      Chen, Yun-Nung",
    booktitle = "Proceedings of the 2024 Conference on Empirical Methods in Natural Language Processing",
    month = nov,
    year = "2024",
    address = "Miami, Florida, USA",
    publisher = "Association for Computational Linguistics",
    url = "https://aclanthology.org/2024.emnlp-main.474/",
    doi = "10.18653/v1/2024.emnlp-main.474",
    pages = "8301--8327"
}

@misc{Deng:et:al:2025,
      title={Rubric-Conditioned LLM Grading: Alignment, Uncertainty, and Robustness}, 
      author={Haotian Deng and Chris Farber and Jiyoon Lee and David Tang},
      year={2025},
      eprint={2601.08843},
      archivePrefix={arXiv},
      primaryClass={cs.CL},
      url={https://arxiv.org/abs/2601.08843}, 
}

@article{Wang:et:al:2025b,
title = {Evaluating large language models as raters in large-scale writing assessments: A psychometric framework for reliability and validity},
journal = {Computers and Education: Artificial Intelligence},
volume = {9},
pages = {100481},
year = {2025},
issn = {2666-920X},
doi = {https://doi.org/10.1016/j.caeai.2025.100481},
url = {https://www.sciencedirect.com/science/article/pii/S2666920X25001213},
author = {Yuehan Wang and Jinyan Huang and Lun Du and Yuxin Guo and Ying Liu and Rong Wang}
}

@inproceedings{Zheng:et:al:2023,
 author = {Zheng, Lianmin and Chiang, Wei-Lin and Sheng, Ying and Zhuang, Siyuan and Wu, Zhanghao and Zhuang, Yonghao and Lin, Zi and Li, Zhuohan and Li, Dacheng and Xing, Eric and Zhang, Hao and Gonzalez, Joseph E and Stoica, Ion},
 booktitle = {Advances in Neural Information Processing Systems},
 editor = {A. Oh and T. Naumann and A. Globerson and K. Saenko and M. Hardt and S. Levine},
 pages = {46595--46623},
 publisher = {Curran Associates, Inc.},
 title = {Judging LLM-as-a-Judge with MT-Bench and Chatbot Arena},
 url = {https://proceedings.neurips.cc/paper_files/paper/2023/file/91f18a1287b398d378ef22505bf41832-Paper-Datasets_and_Benchmarks.pdf},
 volume = {36},
 year = {2023}
}

@inproceedings{Yancey:et:al:2023,
    title = "Rating Short {L}2 Essays on the {CEFR} Scale with {GPT}-4",
    author = "Yancey, Kevin P.  and
      Laflair, Geoffrey  and
      Verardi, Anthony  and
      Burstein, Jill",
    editor = {Kochmar, Ekaterina  and
      Burstein, Jill  and
      Horbach, Andrea  and
      Laarmann-Quante, Ronja  and
      Madnani, Nitin  and
      Tack, Ana{\"i}s  and
      Yaneva, Victoria  and
      Yuan, Zheng  and
      Zesch, Torsten},
    booktitle = "Proceedings of the 18th Workshop on Innovative Use of NLP for Building Educational Applications (BEA 2023)",
    month = jul,
    year = "2023",
    address = "Toronto, Canada",
    publisher = "Association for Computational Linguistics",
    url = "https://aclanthology.org/2023.bea-1.49/",
    doi = "10.18653/v1/2023.bea-1.49",
    pages = "576--584"
}

@inproceedings{Mansour:et:al:2024,
    title = "Can Large Language Models Automatically Score Proficiency of Written Essays?",
    author = "Mansour, Watheq Ahmad  and
      Albatarni, Salam  and
      Eltanbouly, Sohaila  and
      Elsayed, Tamer",
    editor = "Calzolari, Nicoletta  and
      Kan, Min-Yen  and
      Hoste, Veronique  and
      Lenci, Alessandro  and
      Sakti, Sakriani  and
      Xue, Nianwen",
    booktitle = "Proceedings of the 2024 Joint International Conference on Computational Linguistics, Language Resources and Evaluation (LREC-COLING 2024)",
    month = may,
    year = "2024",
    address = "Torino, Italia",
    publisher = "ELRA and ICCL",
    url = "https://aclanthology.org/2024.lrec-main.247/",
    pages = "2777--2786"
}

@article{Gu:et:al:2026,
title = {A survey on LLM-as-a-Judge},
journal = {The Innovation},
pages = {101253},
year = {2026},
issn = {2666-6758},
doi = {https://doi.org/10.1016/j.xinn.2025.101253},
url = {https://www.sciencedirect.com/science/article/pii/S2666675825004564},
author = {Jiawei Gu and Xuhui Jiang and Zhichao Shi and Hexiang Tan and Xuehao Zhai and Chengjin Xu and Wei Li and Yinghan Shen and Shengjie Ma and Honghao Liu and Saizhuo Wang and Kun Zhang and Zhouchi Lin and Bowen Zhang and Lionel Ni and Wen Gao and Yuanzhuo Wang and Jian Guo}
}

@ARTICLE{Hahn:et:al:2021,
  author={Hahn, Marcelo Guerra and Navarro, Silvia Margarita Baldiris and De La Fuente Valentín, Luis and Burgos, Daniel},
  journal={IEEE Access}, 
  title={A Systematic Review of the Effects of Automatic Scoring and Automatic Feedback in Educational Settings}, 
  year={2021},
  volume={9},
  number={},
  pages={108190-108198},
  doi={10.1109/ACCESS.2021.3100890}}

@book{Shermis:Burstein:2013,
  editor    = {Shermis, Mark D. and Burstein, Jill},
  title     = {Handbook of Automated Essay Evaluation: Current Applications and New Directions},
  year      = {2013},
  edition   = {1},
  publisher = {Routledge},
  doi       = {10.4324/9780203122761}
}

@inproceedings{Wang:et:al:2024,
author = {Wang, Xinru and Kim, Hannah and Rahman, Sajjadur and Mitra, Kushan and Miao, Zhengjie},
title = {Human-LLM Collaborative Annotation Through Effective Verification of LLM Labels},
year = {2024},
isbn = {9798400703300},
publisher = {Association for Computing Machinery},
address = {New York, NY, USA},
url = {https://doi.org/10.1145/3613904.3641960},
doi = {10.1145/3613904.3641960},
booktitle = {Proceedings of the 2024 CHI Conference on Human Factors in Computing Systems},
articleno = {303},
numpages = {21},
location = {Honolulu, HI, USA},
series = {CHI '24}
}

@article{Li:et:al:2025,
title = {Large language models-powered clinical decision support: enhancing or replacing human expertise?},
journal = {Intelligent Medicine},
volume = {5},
number = {1},
pages = {1-4},
year = {2025},
issn = {2667-1026},
doi = {https://doi.org/10.1016/j.imed.2025.01.001},
url = {https://www.sciencedirect.com/science/article/pii/S2667102625000014},
author = {Jia Li and Zichun Zhou and Han Lyu and Zhenchang Wang}
}

@article{Schopf:et:al:2025,
  author       = {Michael Schopf},
  title        = {Outperformed by AI: Time to Replace Your Analyst? Find Out Which GenAI Model Does It Best},
  year         = {2025},
  month        = {April},
  note         = {Posted May 6, 2025},
  journal      = {SSRN Electronic Journal},
  url          = {https://ssrn.com/abstract=5222427},
  doi          = {10.2139/ssrn.5222427}
}

@article{Carnat:et:al:2024, 
place={Bologna, Italy},
title={From human-in-the-loop to LLM-in-the-loop for high quality legal dataset},
volume={17},
url={https://i-lex.unibo.it/article/view/20421},
DOI={10.6092/issn.1825-1927/20421},
abstractNote={&amp;lt;p&amp;gt;Annotating legal documents with rhetorical structures is difficult and time-consuming, especially if done completely manually. This paper explores two methodologies for optimal results: first, a human-in-the-loop approach based on a multi-step annotation process with domain experts reviewing and revising datasets iteratively. To enhance interpretability, eXplainable Artificial Intelligence (XAI) models are incorporated, aiding in understanding decision-making processes. Second, an LLM-in-the-loop method has humans leveraging generative large language models (LLMs) to assist experts by automating repetitive annotation tasks under supervision. Further research is proposed to develop interaction models that effectively balance automation with human guidance and accountability.&amp;lt;/p&amp;gt;},
number={1},
journal={i-lex},
author={Carnat, Irina and Comandé, Giovanni and Licari, Daniele and De Nigris, Chiara},
year={2024},
month={Jan.},
pages={27–40}
}

@article{Wilson:et:al:2024,
author = {Wilson, Christopher D. and Haudek, Kevin C. and Osborne, Jonathan F. and Buck Bracey, Zoë E. and Cheuk, Tina and Donovan, Brian M. and Stuhlsatz, Molly A. M. and Santiago, Marisol M. and Zhai, Xiaoming},
title = {Using automated analysis to assess middle school students' competence with scientific argumentation},
journal = {Journal of Research in Science Teaching},
volume = {61},
number = {1},
pages = {38-69},
doi = {https://doi.org/10.1002/tea.21864},
url = {https://onlinelibrary.wiley.com/doi/abs/10.1002/tea.21864},
eprint = {https://onlinelibrary.wiley.com/doi/pdf/10.1002/tea.21864},
year = {2024}
}

@Article{Quah:et:al:2024,
author={Quah, Bernadette
and Zheng, Lei
and Sng, Timothy Jie Han
and Yong, Chee Weng
and Islam, Intekhab},
title={Reliability of ChatGPT in automated essay scoring for dental undergraduate examinations},
journal={BMC Medical Education},
year={2024},
month={Sep},
day={03},
volume={24},
number={1},
pages={962},
issn={1472-6920},
doi={10.1186/s12909-024-05881-6},
url={https://doi.org/10.1186/s12909-024-05881-6}
}

@inproceedings{Clarke:Dietz:2025,
  author       = {Charles Clarke and
                  Laura Dietz},
  title        = {LLM-based Relevance Assessment Still Can't Replace HumanRelevance
                  Assessment},
  booktitle    = {Proceedings of the Eleventh International Workshop on Evaluating Information
                  Access, {EVIA} 2025, a Satellite Workshop of the {NTCIR-18} Conference,
                  Tokyo, Japan, June 10, 2025},
  publisher    = {National Institute of Informatics {(NII)}},
  year         = {2025},
  url          = {https://doi.org/10.20736/0002002105},
  doi          = {10.20736/0002002105}
}

@article{Page:1966,
 ISSN = {00317217},
 URL = {http://www.jstor.org/stable/20371545},
 author = {Ellis B. Page},
 journal = {The Phi Delta Kappan},
 number = {5},
 pages = {238--243},
 publisher = {Phi Delta Kappa International},
 title = {The Imminence of... Grading Essays by Computer},
 urldate = {2026-02-18},
 volume = {47},
 year = {1966}
}

@article{Lee:et:al:2024,
title = {Applying large language models and chain-of-thought for automatic scoring},
journal = {Computers and Education: Artificial Intelligence},
volume = {6},
pages = {100213},
year = {2024},
issn = {2666-920X},
doi = {https://doi.org/10.1016/j.caeai.2024.100213},
url = {https://www.sciencedirect.com/science/article/pii/S2666920X24000146},
author = {Gyeong-Geon Lee and Ehsan Latif and Xuansheng Wu and Ninghao Liu and Xiaoming Zhai},
}

@Article{Kortemeyer:2023,
author={Kortemeyer, Gerd},
title={Performance of the pre-trained large language model GPT-4 on automated short answer grading},
journal={Discover Artificial Intelligence},
year={2024},
month={Jul},
day={05},
volume={4},
number={1},
pages={47},
issn={2731-0809},
doi={10.1007/s44163-024-00147-y},
url={https://doi.org/10.1007/s44163-024-00147-y}
}

@inproceedings{Cozma:et:al:2018,
    title = "Automated essay scoring with string kernels and word embeddings",
    author = "Cozma, M{\u{a}}d{\u{a}}lina  and
      Butnaru, Andrei  and
      Ionescu, Radu Tudor",
    editor = "Gurevych, Iryna  and
      Miyao, Yusuke",
    booktitle = "Proceedings of the 56th Annual Meeting of the Association for Computational Linguistics (Volume 2: Short Papers)",
    month = jul,
    year = "2018",
    address = "Melbourne, Australia",
    publisher = "Association for Computational Linguistics",
    url = "https://aclanthology.org/P18-2080/",
    doi = "10.18653/v1/P18-2080",
    pages = "503--509"
}

@misc{Tornberg:2023,
      title={ChatGPT-4 Outperforms Experts and Crowd Workers in Annotating Political Twitter Messages with Zero-Shot Learning}, 
      author={Petter Törnberg},
      year={2023},
      eprint={2304.06588},
      archivePrefix={arXiv},
      primaryClass={cs.CL},
      url={https://arxiv.org/abs/2304.06588}, 
}

@article{Andersen:et:al:2021,
  title={Benefits of Alternative Evaluation Methods for Automated Essay Scoring.},
  author={Andersen, {\O}istein E and Yuan, Zheng and Watson, Rebecca and Cheung, Kevin Yet Fong},
  journal={International Educational Data Mining Society},
  year={2021},
  publisher={ERIC}
}

@article{Jin:et:al:2025,
author = {Jin, Yiwen and Sercu, Lies},
title = {ChatGPT Interventions in Higher Education: A Systematic Review of Experimental Studies},
journal = {Journal of Computer Assisted Learning},
volume = {41},
number = {4},
pages = {e70072},
doi = {https://doi.org/10.1111/jcal.70072},
url = {https://onlinelibrary.wiley.com/doi/abs/10.1111/jcal.70072},
eprint = {https://onlinelibrary.wiley.com/doi/pdf/10.1111/jcal.70072},
note = {e70072 JCAL-24-1479.R2},
year = {2025}
}

@Article{Emirtekin:2025,
AUTHOR = {Emirtekin, Emrah},
TITLE = {Large Language Model-Powered Automated Assessment: A Systematic Review},
JOURNAL = {Applied Sciences},
VOLUME = {15},
YEAR = {2025},
NUMBER = {10},
ARTICLE-NUMBER = {5683},
URL = {https://www.mdpi.com/2076-3417/15/10/5683},
ISSN = {2076-3417},
DOI = {10.3390/app15105683}
}

@Article{Burrows:et:al:2015,
author={Burrows, Steven
and Gurevych, Iryna
and Stein, Benno},
title={The Eras and Trends of Automatic Short Answer Grading},
journal={International Journal of Artificial Intelligence in Education},
year={2015},
month={Mar},
day={01},
volume={25},
number={1},
pages={60-117},
issn={1560-4306},
doi={10.1007/s40593-014-0026-8},
url={https://doi.org/10.1007/s40593-014-0026-8}
}

@book{Mislevy:2018,
  author    = {Robert J. Mislevy},
  title     = {Sociocognitive Foundations of Educational Measurement},
  year      = {2018},
  publisher = {Routledge},
  edition   = {1},
  doi       = {10.4324/9781315871691},
  url       = {https://doi.org/10.4324/9781315871691}
}

@misc{Haller:et:al:2022,
      title={Survey on Automated Short Answer Grading with Deep Learning: from Word Embeddings to Transformers}, 
      author={Stefan Haller and Adina Aldea and Christin Seifert and Nicola Strisciuglio},
      year={2022},
      eprint={2204.03503},
      archivePrefix={arXiv},
      primaryClass={cs.CL},
      url={https://arxiv.org/abs/2204.03503}, 
}

@misc{Ma:2025,
      title={Can Machines Think Like Humans? A Behavioral Evaluation of LLM Agents in Dictator Games}, 
      author={Ji Ma},
      year={2025},
      eprint={2410.21359},
      archivePrefix={arXiv},
      primaryClass={cs.CL},
      url={https://arxiv.org/abs/2410.21359}, 
}

@article{Weidmann:et:al:2025,
   title={Large Language Models Are Democracy Coders with Attitudes},
   volume={59},
   ISSN={1537-5935},
   url={http://dx.doi.org/10.1017/S1049096525101248},
   DOI={10.1017/s1049096525101248},
   number={1},
   journal={PS: Political Science \& Politics},
   publisher={Cambridge University Press (CUP)},
   author={Weidmann, Nils B. and Faulborn, Mats and García, David},
   year={2025},
   month=jul, pages={17–23}
}

@article{Rudner:Gagne:2001,
  author  = {Rudner, Lawrence M. and Gagne, Phill},
  title   = {An Overview of Three Approaches to Scoring Written Essays by Computer},
  journal = {Practical Assessment, Research \& Evaluation},
  volume  = {7},
  number  = {26},
  year    = {2001}
}

@article{Page:1968,
 ISSN = {00208566, 15730638},
 URL = {http://www.jstor.org/stable/3442515},
 author = {Ellis B. Page},
 journal = {International Review of Education / Internationale Zeitschrift für Erziehungswissenschaft / Revue Internationale de l'Education},
 number = {2},
 pages = {210--225},
 publisher = {Springer},
 title = {The Use of the Computer in Analyzing Student Essays},
 urldate = {2026-02-23},
 volume = {14},
 year = {1968}
}

@article{Eckes:2005,
author = {Thomas Eckes},
title = {Examining Rater Effects in TestDaF Writing and Speaking Performance Assessments: A Many-Facet Rasch Analysis},
journal = {Language Assessment Quarterly},
volume = {2},
number = {3},
pages = {197--221},
year = {2005},
publisher = {Routledge},
doi = {10.1207/s15434311laq0203\_2},
URL = {https://doi.org/10.1207/s15434311laq0203\_2},
eprint = {https://doi.org/10.1207/s15434311laq0203\_2}
}

@article{Mahshanian:Shahnazari:2020,
	author = {Amir Mahshanian and Mohammadtaghi Shahnazari},
	title = {The effect of raters fatigue on scoring EFL writing tasks},
	journal = {Indonesian Journal of Applied Linguistics},
	volume = {10},
	number = {1},
	year = {2020},
	issn = {2502-6747},	pages = {1--13},	doi = {10.17509/ijal.v10i1.24956},
	url = {https://ejournal.upi.edu/index.php/IJAL/article/view/24956}
}

@Article{Jin:Chiu:2022,
author={Jin, Kuan-Yu
and Chiu, Ming Ming},
title={A mixture Rasch facets model for rater's illusory halo effects},
journal={Behavior Research Methods},
year={2022},
month={Dec},
day={01},
volume={54},
number={6},
pages={2750-2764},
issn={1554-3528},
doi={10.3758/s13428-021-01721-3},
url={https://doi.org/10.3758/s13428-021-01721-3}
}

@techreport{Livingston:2009,
  author      = {Livingston, Samuel A.},
  title       = {Constructed-Response Test Questions: Why We Use Them; How We Score Them},
  institution = {Educational Testing Service},
  type        = {R\&D Connections},
  number      = {11},
  address     = {Princeton, NJ},
  month       = sep,
  year        = {2009},
  pages       = {8},
  url         = {https://eric.ed.gov/?id=ED507802},
  note        = {ERIC Number: ED507802}
}

@article{Firoozi:et:al:2022,
   title={Using Active Learning Methods to Strategically Select Essays for Automated Scoring},
   volume={42},
   ISSN={1745-3992},
   url={http://dx.doi.org/10.1111/emip.12537},
   DOI={10.1111/emip.12537},
   number={1},
   journal={Educational Measurement: Issues and Practice},
   publisher={Wiley},
   author={Firoozi, Tahereh and Mohammadi, Hamid and Gierl, Mark J.},
   year={2022},
   month=dec, pages={34–43} }

@article{Korkmaz:Correia:2019,
author = {Ceren Korkmaz and Ana-Paula Correia},
title = {A review of research on machine learning in educational technology},
journal = {Educational Media International},
volume = {56},
number = {3},
pages = {250--267},
year = {2019},
publisher = {Routledge},
doi = {10.1080/09523987.2019.1669875},
URL = {https://doi.org/10.1080/09523987.2019.1669875},
eprint = {https://doi.org/10.1080/09523987.2019.1669875}
}

@Article{Wang:et:al:2025,
author={Wang, Angelina
and Morgenstern, Jamie
and Dickerson, John P.},
title={Large language models that replace human participants can harmfully misportray and flatten identity groups},
journal={Nature Machine Intelligence},
year={2025},
month={Mar},
day={01},
volume={7},
number={3},
pages={400-411},
issn={2522-5839},
doi={10.1038/s42256-025-00986-z},
url={https://doi.org/10.1038/s42256-025-00986-z}
}

@Article{Luo:et:al:2025,
author={Luo, Xiaoliang
and Rechardt, Akilles
and Sun, Guangzhi
and Nejad, Kevin K.
and Y{\'a}{\~{n}}ez, Felipe
and Yilmaz, Bati
and Lee, Kangjoo
and Cohen, Alexandra O.
and Borghesani, Valentina
and Pashkov, Anton
and Marinazzo, Daniele
and Nicholas, Jonathan
and Salatiello, Alessandro
and Sucholutsky, Ilia
and Minervini, Pasquale
and Razavi, Sepehr
and Rocca, Roberta
and Yusifov, Elkhan
and Okalova, Tereza
and Gu, Nianlong
and Ferianc, Martin
and Khona, Mikail
and Patil, Kaustubh R.
and Lee, Pui-Shee
and Mata, Rui
and Myers, Nicholas E.
and Bizley, Jennifer K.
and Musslick, Sebastian
and Bilgin, Isil Poyraz
and Niso, Guiomar
and Ales, Justin M.
and Gaebler, Michael
and Ratan Murty, N. Apurva
and Loued-Khenissi, Leyla
and Behler, Anna
and Hall, Chloe M.
and Dafflon, Jessica
and Bao, Sherry Dongqi
and Love, Bradley C.},
title={Large language models surpass human experts in predicting neuroscience results},
journal={Nature Human Behaviour},
year={2025},
month={Feb},
day={01},
volume={9},
number={2},
pages={305-315},
issn={2397-3374},
doi={10.1038/s41562-024-02046-9},
url={https://doi.org/10.1038/s41562-024-02046-9}
}

@inproceedings{Wong:et:al:2021,
    title = "Cross-replication Reliability - An Empirical Approach to Interpreting Inter-rater Reliability",
    author = "Wong, Ka  and
      Paritosh, Praveen  and
      Aroyo, Lora",
    editor = "Zong, Chengqing  and
      Xia, Fei  and
      Li, Wenjie  and
      Navigli, Roberto",
    booktitle = "Proceedings of the 59th Annual Meeting of the Association for Computational Linguistics and the 11th International Joint Conference on Natural Language Processing (Volume 1: Long Papers)",
    month = aug,
    year = "2021",
    address = "Online",
    publisher = "Association for Computational Linguistics",
    url = "https://aclanthology.org/2021.acl-long.548/",
    doi = "10.18653/v1/2021.acl-long.548",
    pages = "7053--7065"
}

@misc{Piot:et:al:2025,
      title={Can LLMs Evaluate What They Cannot Annotate? Revisiting LLM Reliability in Hate Speech Detection}, 
      author={Paloma Piot and David Otero and Patricia Martín-Rodilla and Javier Parapar},
      year={2025},
      eprint={2512.09662},
      archivePrefix={arXiv},
      primaryClass={cs.CL},
      url={https://arxiv.org/abs/2512.09662}, 
}

@Article{Bafna:et:al:2025,
author={Bafna, Khushi
and Yadav, Amit Kumar
and Bagaria, Shaurya
and Gianchandani, Ayushi
and Poduval, Murali
and Bagaria, Vaibhav},
title={Is ChatGPT more Humane than Humans?---Accuracy, Relevance and humanness of the Answers Provided by ChatGPT for Patient Education for Total Knee Replacement},
journal={Indian Journal of Orthopaedics},
year={2025},
month={Nov},
day={01},
volume={59},
number={11},
pages={1880-1886},
issn={1998-3727},
doi={10.1007/s43465-025-01474-7},
url={https://doi.org/10.1007/s43465-025-01474-7}
}

@article{Emirtekin:Ozarslan:2026,
author = {Emirtekin, Emrah and Özarslan, Yasin},
title = {Automatic Short-Answer Grading in Sustainability Education: AI–Human Agreement},
journal = {Journal of Computer Assisted Learning},
volume = {42},
number = {1},
pages = {e70160},
doi = {https://doi.org/10.1002/jcal.70160},
url = {https://onlinelibrary.wiley.com/doi/abs/10.1002/jcal.70160},
eprint = {https://onlinelibrary.wiley.com/doi/pdf/10.1002/jcal.70160},
note = {e70160 JCAL-25-1064},
year = {2026}
}

@inproceedings{Burges:et:al:2005,
author = {Burges, Chris and Shaked, Tal and Renshaw, Erin and Lazier, Ari and Deeds, Matt and Hamilton, Nicole and Hullender, Greg},
title = {Learning to rank using gradient descent},
year = {2005},
isbn = {1595931805},
publisher = {Association for Computing Machinery},
address = {New York, NY, USA},
url = {https://doi.org/10.1145/1102351.1102363},
doi = {10.1145/1102351.1102363},
booktitle = {Proceedings of the 22nd International Conference on Machine Learning},
pages = {89–96},
numpages = {8},
location = {Bonn, Germany},
series = {ICML '05}
}

@inproceedings{Alikaniotis:et:al:2016,
    title = "Automatic Text Scoring Using Neural Networks",
    author = "Alikaniotis, Dimitrios  and
      Yannakoudakis, Helen  and
      Rei, Marek",
    editor = "Erk, Katrin  and
      Smith, Noah A.",
    booktitle = "Proceedings of the 54th Annual Meeting of the Association for Computational Linguistics (Volume 1: Long Papers)",
    month = aug,
    year = "2016",
    address = "Berlin, Germany",
    publisher = "Association for Computational Linguistics",
    url = "https://aclanthology.org/P16-1068/",
    doi = "10.18653/v1/P16-1068",
    pages = "715--725"
}

@inproceedings{Taghipour:et:al:2016,
    title = "A Neural Approach to Automated Essay Scoring",
    author = "Taghipour, Kaveh  and
      Ng, Hwee Tou",
    editor = "Su, Jian  and
      Duh, Kevin  and
      Carreras, Xavier",
    booktitle = "Proceedings of the 2016 Conference on Empirical Methods in Natural Language Processing",
    month = nov,
    year = "2016",
    address = "Austin, Texas",
    publisher = "Association for Computational Linguistics",
    url = "https://aclanthology.org/D16-1193/",
    doi = "10.18653/v1/D16-1193",
    pages = "1882--1891"
}

@inproceedings{Wang:et:al:2018,
    title = "Automatic Essay Scoring Incorporating Rating Schema via Reinforcement Learning",
    author = "Wang, Yucheng  and
      Wei, Zhongyu  and
      Zhou, Yaqian  and
      Huang, Xuanjing",
    editor = "Riloff, Ellen  and
      Chiang, David  and
      Hockenmaier, Julia  and
      Tsujii, Jun{'}ichi",
    booktitle = "Proceedings of the 2018 Conference on Empirical Methods in Natural Language Processing",
    month = oct # "-" # nov,
    year = "2018",
    address = "Brussels, Belgium",
    publisher = "Association for Computational Linguistics",
    url = "https://aclanthology.org/D18-1090/",
    doi = "10.18653/v1/D18-1090",
    pages = "791--797"
}

@article{Chen:et:al:2013,
    author = {Chen, Hongbo and Xu, Jungang and He, Ben},
    title = {Automated Essay Scoring by Capturing Relative Writing Quality1},
    journal = {The Computer Journal},
    volume = {57},
    number = {9},
    pages = {1318-1330},
    year = {2013},
    month = {10},
    issn = {0010-4620},
    doi = {10.1093/comjnl/bxt117},
    url = {https://doi.org/10.1093/comjnl/bxt117},
    eprint = {https://academic.oup.com/comjnl/article-pdf/57/9/1318/1528257/bxt117.pdf},
}

@article{Gilardi:et:al:2023,
author = {Fabrizio Gilardi  and Meysam Alizadeh  and Maël Kubli },
title = {ChatGPT outperforms crowd workers for text-annotation tasks},
journal = {Proceedings of the National Academy of Sciences},
volume = {120},
number = {30},
pages = {e2305016120},
year = {2023},
doi = {10.1073/pnas.2305016120},
URL = {https://www.pnas.org/doi/abs/10.1073/pnas.2305016120},
eprint = {https://www.pnas.org/doi/pdf/10.1073/pnas.2305016120}}

@misc{Jiang:et:al:2024,
      title={Mixtral of Experts}, 
      author={Albert Q. Jiang and Alexandre Sablayrolles and Antoine Roux and Arthur Mensch and Blanche Savary and Chris Bamford and Devendra Singh Chaplot and Diego de las Casas and Emma Bou Hanna and Florian Bressand and Gianna Lengyel and Guillaume Bour and Guillaume Lample and Lélio Renard Lavaud and Lucile Saulnier and Marie-Anne Lachaux and Pierre Stock and Sandeep Subramanian and Sophia Yang and Szymon Antoniak and Teven Le Scao and Théophile Gervet and Thibaut Lavril and Thomas Wang and Timothée Lacroix and William El Sayed},
      year={2024},
      eprint={2401.04088},
      archivePrefix={arXiv},
      primaryClass={cs.LG},
      url={https://arxiv.org/abs/2401.04088}, 
}

@misc{Yang:et:al:2025,
      title={Qwen3 Technical Report}, 
      author={An Yang and Anfeng Li and Baosong Yang and Beichen Zhang and Binyuan Hui and Bo Zheng and Bowen Yu and Chang Gao and Chengen Huang and Chenxu Lv and Chujie Zheng and Dayiheng Liu and Fan Zhou and Fei Huang and Feng Hu and Hao Ge and Haoran Wei and Huan Lin and Jialong Tang and Jian Yang and Jianhong Tu and Jianwei Zhang and Jianxin Yang and Jiaxi Yang and Jing Zhou and Jingren Zhou and Junyang Lin and Kai Dang and Keqin Bao and Kexin Yang and Le Yu and Lianghao Deng and Mei Li and Mingfeng Xue and Mingze Li and Pei Zhang and Peng Wang and Qin Zhu and Rui Men and Ruize Gao and Shixuan Liu and Shuang Luo and Tianhao Li and Tianyi Tang and Wenbiao Yin and Xingzhang Ren and Xinyu Wang and Xinyu Zhang and Xuancheng Ren and Yang Fan and Yang Su and Yichang Zhang and Yinger Zhang and Yu Wan and Yuqiong Liu and Zekun Wang and Zeyu Cui and Zhenru Zhang and Zhipeng Zhou and Zihan Qiu},
      year={2025},
      eprint={2505.09388},
      archivePrefix={arXiv},
      primaryClass={cs.CL},
      url={https://arxiv.org/abs/2505.09388}, 
}

@misc{Jiang:et:al:2023,
      title={Mistral 7B}, 
      author={Albert Q. Jiang and Alexandre Sablayrolles and Arthur Mensch and Chris Bamford and Devendra Singh Chaplot and Diego de las Casas and Florian Bressand and Gianna Lengyel and Guillaume Lample and Lucile Saulnier and Lélio Renard Lavaud and Marie-Anne Lachaux and Pierre Stock and Teven Le Scao and Thibaut Lavril and Thomas Wang and Timothée Lacroix and William El Sayed},
      year={2023},
      eprint={2310.06825},
      archivePrefix={arXiv},
      primaryClass={cs.CL},
      url={https://arxiv.org/abs/2310.06825}, 
}

@misc{OpenAI:et:al:2025,
      title={gpt-oss-120b \& gpt-oss-20b Model Card}, 
      author={OpenAI and Sandhini Agarwal and Lama Ahmad and Jason Ai and Sam Altman and Andy Applebaum and Edwin Arbus and Rahul K. Arora and Yu Bai and Bowen Baker and Haiming Bao and Boaz Barak and Ally Bennett and Tyler Bertao and Nivedita Brett and Eugene Brevdo and Greg Brockman and Sebastien Bubeck and Che Chang and Kai Chen and Mark Chen and Enoch Cheung and Aidan Clark and Dan Cook and Marat Dukhan and Casey Dvorak and Kevin Fives and Vlad Fomenko and Timur Garipov and Kristian Georgiev and Mia Glaese and Tarun Gogineni and Adam Goucher and Lukas Gross and Katia Gil Guzman and John Hallman and Jackie Hehir and Johannes Heidecke and Alec Helyar and Haitang Hu and Romain Huet and Jacob Huh and Saachi Jain and Zach Johnson and Chris Koch and Irina Kofman and Dominik Kundel and Jason Kwon and Volodymyr Kyrylov and Elaine Ya Le and Guillaume Leclerc and James Park Lennon and Scott Lessans and Mario Lezcano-Casado and Yuanzhi Li and Zhuohan Li and Ji Lin and Jordan Liss and Lily and Liu and Jiancheng Liu and Kevin Lu and Chris Lu and Zoran Martinovic and Lindsay McCallum and Josh McGrath and Scott McKinney and Aidan McLaughlin and Song Mei and Steve Mostovoy and Tong Mu and Gideon Myles and Alexander Neitz and Alex Nichol and Jakub Pachocki and Alex Paino and Dana Palmie and Ashley Pantuliano and Giambattista Parascandolo and Jongsoo Park and Leher Pathak and Carolina Paz and Ludovic Peran and Dmitry Pimenov and Michelle Pokrass and Elizabeth Proehl and Huida Qiu and Gaby Raila and Filippo Raso and Hongyu Ren and Kimmy Richardson and David Robinson and Bob Rotsted and Hadi Salman and Suvansh Sanjeev and Max Schwarzer and D. Sculley and Harshit Sikchi and Kendal Simon and Karan Singhal and Yang Song and Dane Stuckey and Zhiqing Sun and Philippe Tillet and Sam Toizer and Foivos Tsimpourlas and Nikhil Vyas and Eric Wallace and Xin Wang and Miles Wang and Olivia Watkins and Kevin Weil and Amy Wendling and Kevin Whinnery and Cedric Whitney and Hannah Wong and Lin Yang and Yu Yang and Michihiro Yasunaga and Kristen Ying and Wojciech Zaremba and Wenting Zhan and Cyril Zhang and Brian Zhang and Eddie Zhang and Shengjia Zhao},
      year={2025},
      eprint={2508.10925},
      archivePrefix={arXiv},
      primaryClass={cs.CL},
      url={https://arxiv.org/abs/2508.10925}, 
}

@misc{Grattafiori:et:al:2024,
      title={The Llama 3 Herd of Models}, 
      author={Aaron Grattafiori and Abhimanyu Dubey and Abhinav Jauhri and Abhinav Pandey and Abhishek Kadian and Ahmad Al-Dahle and Aiesha Letman and Akhil Mathur and Alan Schelten and Alex Vaughan and Amy Yang and Angela Fan and Anirudh Goyal and Anthony Hartshorn and Aobo Yang and Archi Mitra and Archie Sravankumar and Artem Korenev and Arthur Hinsvark and Arun Rao and Aston Zhang and Aurelien Rodriguez and Austen Gregerson and Ava Spataru and Baptiste Roziere and Bethany Biron and Binh Tang and Bobbie Chern and Charlotte Caucheteux and Chaya Nayak and Chloe Bi and Chris Marra and Chris McConnell and Christian Keller and Christophe Touret and Chunyang Wu and Corinne Wong and Cristian Canton Ferrer and Cyrus Nikolaidis and Damien Allonsius and Daniel Song and Danielle Pintz and Danny Livshits and Danny Wyatt and David Esiobu and Dhruv Choudhary and Dhruv Mahajan and Diego Garcia-Olano and Diego Perino and Dieuwke Hupkes and Egor Lakomkin and Ehab AlBadawy and Elina Lobanova and Emily Dinan and Eric Michael Smith and Filip Radenovic and Francisco Guzmán and Frank Zhang and Gabriel Synnaeve and Gabrielle Lee and Georgia Lewis Anderson and Govind Thattai and Graeme Nail and Gregoire Mialon and Guan Pang and Guillem Cucurell and Hailey Nguyen and Hannah Korevaar and Hu Xu and Hugo Touvron and Iliyan Zarov and Imanol Arrieta Ibarra and Isabel Kloumann and Ishan Misra and Ivan Evtimov and Jack Zhang and Jade Copet and Jaewon Lee and Jan Geffert and Jana Vranes and Jason Park and Jay Mahadeokar and Jeet Shah and Jelmer van der Linde and Jennifer Billock and Jenny Hong and Jenya Lee and Jeremy Fu and Jianfeng Chi and Jianyu Huang and Jiawen Liu and Jie Wang and Jiecao Yu and Joanna Bitton and Joe Spisak and Jongsoo Park and Joseph Rocca and Joshua Johnstun and Joshua Saxe and Junteng Jia and Kalyan Vasuden Alwala and Karthik Prasad and Kartikeya Upasani and Kate Plawiak and Ke Li and Kenneth Heafield and Kevin Stone and Khalid El-Arini and Krithika Iyer and Kshitiz Malik and Kuenley Chiu and Kunal Bhalla and Kushal Lakhotia and Lauren Rantala-Yeary and Laurens van der Maaten and Lawrence Chen and Liang Tan and Liz Jenkins and Louis Martin and Lovish Madaan and Lubo Malo and Lukas Blecher and Lukas Landzaat and Luke de Oliveira and Madeline Muzzi and Mahesh Pasupuleti and Mannat Singh and Manohar Paluri and Marcin Kardas and Maria Tsimpoukelli and Mathew Oldham and Mathieu Rita and Maya Pavlova and Melanie Kambadur and Mike Lewis and Min Si and Mitesh Kumar Singh and Mona Hassan and Naman Goyal and Narjes Torabi and Nikolay Bashlykov and Nikolay Bogoychev and Niladri Chatterji and Ning Zhang and Olivier Duchenne and Onur Çelebi and Patrick Alrassy and Pengchuan Zhang and Pengwei Li and Petar Vasic and Peter Weng and Prajjwal Bhargava and Pratik Dubal and Praveen Krishnan and Punit Singh Koura and Puxin Xu and Qing He and Qingxiao Dong and Ragavan Srinivasan and Raj Ganapathy and Ramon Calderer and Ricardo Silveira Cabral and Robert Stojnic and Roberta Raileanu and Rohan Maheswari and Rohit Girdhar and Rohit Patel and Romain Sauvestre and Ronnie Polidoro and Roshan Sumbaly and Ross Taylor and Ruan Silva and Rui Hou and Rui Wang and Saghar Hosseini and Sahana Chennabasappa and Sanjay Singh and Sean Bell and Seohyun Sonia Kim and Sergey Edunov and Shaoliang Nie and Sharan Narang and Sharath Raparthy and Sheng Shen and Shengye Wan and Shruti Bhosale and Shun Zhang and Simon Vandenhende and Soumya Batra and Spencer Whitman and Sten Sootla and Stephane Collot and Suchin Gururangan and Sydney Borodinsky and Tamar Herman and Tara Fowler and Tarek Sheasha and Thomas Georgiou and Thomas Scialom and Tobias Speckbacher and Todor Mihaylov and Tong Xiao and Ujjwal Karn and Vedanuj Goswami and Vibhor Gupta and Vignesh Ramanathan and Viktor Kerkez and Vincent Gonguet and Virginie Do and Vish Vogeti and Vítor Albiero and Vladan Petrovic and Weiwei Chu and Wenhan Xiong and Wenyin Fu and Whitney Meers and Xavier Martinet and Xiaodong Wang and Xiaofang Wang and Xiaoqing Ellen Tan and Xide Xia and Xinfeng Xie and Xuchao Jia and Xuewei Wang and Yaelle Goldschlag and Yashesh Gaur and Yasmine Babaei and Yi Wen and Yiwen Song and Yuchen Zhang and Yue Li and Yuning Mao and Zacharie Delpierre Coudert and Zheng Yan and Zhengxing Chen and Zoe Papakipos and Aaditya Singh and Aayushi Srivastava and Abha Jain and Adam Kelsey and Adam Shajnfeld and Adithya Gangidi and Adolfo Victoria and Ahuva Goldstand and Ajay Menon and Ajay Sharma and Alex Boesenberg and Alexei Baevski and Allie Feinstein and Amanda Kallet and Amit Sangani and Amos Teo and Anam Yunus and Andrei Lupu and Andres Alvarado and Andrew Caples and Andrew Gu and Andrew Ho and Andrew Poulton and Andrew Ryan and Ankit Ramchandani and Annie Dong and Annie Franco and Anuj Goyal and Aparajita Saraf and Arkabandhu Chowdhury and Ashley Gabriel and Ashwin Bharambe and Assaf Eisenman and Azadeh Yazdan and Beau James and Ben Maurer and Benjamin Leonhardi and Bernie Huang and Beth Loyd and Beto De Paola and Bhargavi Paranjape and Bing Liu and Bo Wu and Boyu Ni and Braden Hancock and Bram Wasti and Brandon Spence and Brani Stojkovic and Brian Gamido and Britt Montalvo and Carl Parker and Carly Burton and Catalina Mejia and Ce Liu and Changhan Wang and Changkyu Kim and Chao Zhou and Chester Hu and Ching-Hsiang Chu and Chris Cai and Chris Tindal and Christoph Feichtenhofer and Cynthia Gao and Damon Civin and Dana Beaty and Daniel Kreymer and Daniel Li and David Adkins and David Xu and Davide Testuggine and Delia David and Devi Parikh and Diana Liskovich and Didem Foss and Dingkang Wang and Duc Le and Dustin Holland and Edward Dowling and Eissa Jamil and Elaine Montgomery and Eleonora Presani and Emily Hahn and Emily Wood and Eric-Tuan Le and Erik Brinkman and Esteban Arcaute and Evan Dunbar and Evan Smothers and Fei Sun and Felix Kreuk and Feng Tian and Filippos Kokkinos and Firat Ozgenel and Francesco Caggioni and Frank Kanayet and Frank Seide and Gabriela Medina Florez and Gabriella Schwarz and Gada Badeer and Georgia Swee and Gil Halpern and Grant Herman and Grigory Sizov and Guangyi and Zhang and Guna Lakshminarayanan and Hakan Inan and Hamid Shojanazeri and Han Zou and Hannah Wang and Hanwen Zha and Haroun Habeeb and Harrison Rudolph and Helen Suk and Henry Aspegren and Hunter Goldman and Hongyuan Zhan and Ibrahim Damlaj and Igor Molybog and Igor Tufanov and Ilias Leontiadis and Irina-Elena Veliche and Itai Gat and Jake Weissman and James Geboski and James Kohli and Janice Lam and Japhet Asher and Jean-Baptiste Gaya and Jeff Marcus and Jeff Tang and Jennifer Chan and Jenny Zhen and Jeremy Reizenstein and Jeremy Teboul and Jessica Zhong and Jian Jin and Jingyi Yang and Joe Cummings and Jon Carvill and Jon Shepard and Jonathan McPhie and Jonathan Torres and Josh Ginsburg and Junjie Wang and Kai Wu and Kam Hou U and Karan Saxena and Kartikay Khandelwal and Katayoun Zand and Kathy Matosich and Kaushik Veeraraghavan and Kelly Michelena and Keqian Li and Kiran Jagadeesh and Kun Huang and Kunal Chawla and Kyle Huang and Lailin Chen and Lakshya Garg and Lavender A and Leandro Silva and Lee Bell and Lei Zhang and Liangpeng Guo and Licheng Yu and Liron Moshkovich and Luca Wehrstedt and Madian Khabsa and Manav Avalani and Manish Bhatt and Martynas Mankus and Matan Hasson and Matthew Lennie and Matthias Reso and Maxim Groshev and Maxim Naumov and Maya Lathi and Meghan Keneally and Miao Liu and Michael L. Seltzer and Michal Valko and Michelle Restrepo and Mihir Patel and Mik Vyatskov and Mikayel Samvelyan and Mike Clark and Mike Macey and Mike Wang and Miquel Jubert Hermoso and Mo Metanat and Mohammad Rastegari and Munish Bansal and Nandhini Santhanam and Natascha Parks and Natasha White and Navyata Bawa and Nayan Singhal and Nick Egebo and Nicolas Usunier and Nikhil Mehta and Nikolay Pavlovich Laptev and Ning Dong and Norman Cheng and Oleg Chernoguz and Olivia Hart and Omkar Salpekar and Ozlem Kalinli and Parkin Kent and Parth Parekh and Paul Saab and Pavan Balaji and Pedro Rittner and Philip Bontrager and Pierre Roux and Piotr Dollar and Polina Zvyagina and Prashant Ratanchandani and Pritish Yuvraj and Qian Liang and Rachad Alao and Rachel Rodriguez and Rafi Ayub and Raghotham Murthy and Raghu Nayani and Rahul Mitra and Rangaprabhu Parthasarathy and Raymond Li and Rebekkah Hogan and Robin Battey and Rocky Wang and Russ Howes and Ruty Rinott and Sachin Mehta and Sachin Siby and Sai Jayesh Bondu and Samyak Datta and Sara Chugh and Sara Hunt and Sargun Dhillon and Sasha Sidorov and Satadru Pan and Saurabh Mahajan and Saurabh Verma and Seiji Yamamoto and Sharadh Ramaswamy and Shaun Lindsay and Shaun Lindsay and Sheng Feng and Shenghao Lin and Shengxin Cindy Zha and Shishir Patil and Shiva Shankar and Shuqiang Zhang and Shuqiang Zhang and Sinong Wang and Sneha Agarwal and Soji Sajuyigbe and Soumith Chintala and Stephanie Max and Stephen Chen and Steve Kehoe and Steve Satterfield and Sudarshan Govindaprasad and Sumit Gupta and Summer Deng and Sungmin Cho and Sunny Virk and Suraj Subramanian and Sy Choudhury and Sydney Goldman and Tal Remez and Tamar Glaser and Tamara Best and Thilo Koehler and Thomas Robinson and Tianhe Li and Tianjun Zhang and Tim Matthews and Timothy Chou and Tzook Shaked and Varun Vontimitta and Victoria Ajayi and Victoria Montanez and Vijai Mohan and Vinay Satish Kumar and Vishal Mangla and Vlad Ionescu and Vlad Poenaru and Vlad Tiberiu Mihailescu and Vladimir Ivanov and Wei Li and Wenchen Wang and Wenwen Jiang and Wes Bouaziz and Will Constable and Xiaocheng Tang and Xiaojian Wu and Xiaolan Wang and Xilun Wu and Xinbo Gao and Yaniv Kleinman and Yanjun Chen and Ye Hu and Ye Jia and Ye Qi and Yenda Li and Yilin Zhang and Ying Zhang and Yossi Adi and Youngjin Nam and Yu and Wang and Yu Zhao and Yuchen Hao and Yundi Qian and Yunlu Li and Yuzi He and Zach Rait and Zachary DeVito and Zef Rosnbrick and Zhaoduo Wen and Zhenyu Yang and Zhiwei Zhao and Zhiyu Ma},
      year={2024},
      eprint={2407.21783},
      archivePrefix={arXiv},
      primaryClass={cs.AI},
      url={https://arxiv.org/abs/2407.21783}, 
}

@misc{Gemma:et:al:2024,
      title={Gemma 2: Improving Open Language Models at a Practical Size}, 
      author={{Gemma Team} and Morgane Riviere and Shreya Pathak and Pier Giuseppe Sessa and Cassidy Hardin and Surya Bhupatiraju and Léonard Hussenot and Thomas Mesnard and Bobak Shahriari and Alexandre Ramé and Johan Ferret and Peter Liu and Pouya Tafti and Abe Friesen and Michelle Casbon and Sabela Ramos and Ravin Kumar and Charline Le Lan and Sammy Jerome and Anton Tsitsulin and Nino Vieillard and Piotr Stanczyk and Sertan Girgin and Nikola Momchev and Matt Hoffman and Shantanu Thakoor and Jean-Bastien Grill and Behnam Neyshabur and Olivier Bachem and Alanna Walton and Aliaksei Severyn and Alicia Parrish and Aliya Ahmad and Allen Hutchison and Alvin Abdagic and Amanda Carl and Amy Shen and Andy Brock and Andy Coenen and Anthony Laforge and Antonia Paterson and Ben Bastian and Bilal Piot and Bo Wu and Brandon Royal and Charlie Chen and Chintu Kumar and Chris Perry and Chris Welty and Christopher A. Choquette-Choo and Danila Sinopalnikov and David Weinberger and Dimple Vijaykumar and Dominika Rogozińska and Dustin Herbison and Elisa Bandy and Emma Wang and Eric Noland and Erica Moreira and Evan Senter and Evgenii Eltyshev and Francesco Visin and Gabriel Rasskin and Gary Wei and Glenn Cameron and Gus Martins and Hadi Hashemi and Hanna Klimczak-Plucińska and Harleen Batra and Harsh Dhand and Ivan Nardini and Jacinda Mein and Jack Zhou and James Svensson and Jeff Stanway and Jetha Chan and Jin Peng Zhou and Joana Carrasqueira and Joana Iljazi and Jocelyn Becker and Joe Fernandez and Joost van Amersfoort and Josh Gordon and Josh Lipschultz and Josh Newlan and Ju-yeong Ji and Kareem Mohamed and Kartikeya Badola and Kat Black and Katie Millican and Keelin McDonell and Kelvin Nguyen and Kiranbir Sodhia and Kish Greene and Lars Lowe Sjoesund and Lauren Usui and Laurent Sifre and Lena Heuermann and Leticia Lago and Lilly McNealus and Livio Baldini Soares and Logan Kilpatrick and Lucas Dixon and Luciano Martins and Machel Reid and Manvinder Singh and Mark Iverson and Martin Görner and Mat Velloso and Mateo Wirth and Matt Davidow and Matt Miller and Matthew Rahtz and Matthew Watson and Meg Risdal and Mehran Kazemi and Michael Moynihan and Ming Zhang and Minsuk Kahng and Minwoo Park and Mofi Rahman and Mohit Khatwani and Natalie Dao and Nenshad Bardoliwalla and Nesh Devanathan and Neta Dumai and Nilay Chauhan and Oscar Wahltinez and Pankil Botarda and Parker Barnes and Paul Barham and Paul Michel and Pengchong Jin and Petko Georgiev and Phil Culliton and Pradeep Kuppala and Ramona Comanescu and Ramona Merhej and Reena Jana and Reza Ardeshir Rokni and Rishabh Agarwal and Ryan Mullins and Samaneh Saadat and Sara Mc Carthy and Sarah Cogan and Sarah Perrin and Sébastien M. R. Arnold and Sebastian Krause and Shengyang Dai and Shruti Garg and Shruti Sheth and Sue Ronstrom and Susan Chan and Timothy Jordan and Ting Yu and Tom Eccles and Tom Hennigan and Tomas Kocisky and Tulsee Doshi and Vihan Jain and Vikas Yadav and Vilobh Meshram and Vishal Dharmadhikari and Warren Barkley and Wei Wei and Wenming Ye and Woohyun Han and Woosuk Kwon and Xiang Xu and Zhe Shen and Zhitao Gong and Zichuan Wei and Victor Cotruta and Phoebe Kirk and Anand Rao and Minh Giang and Ludovic Peran and Tris Warkentin and Eli Collins and Joelle Barral and Zoubin Ghahramani and Raia Hadsell and D. Sculley and Jeanine Banks and Anca Dragan and Slav Petrov and Oriol Vinyals and Jeff Dean and Demis Hassabis and Koray Kavukcuoglu and Clement Farabet and Elena Buchatskaya and Sebastian Borgeaud and Noah Fiedel and Armand Joulin and Kathleen Kenealy and Robert Dadashi and Alek Andreev},
      year={2024},
      eprint={2408.00118},
      archivePrefix={arXiv},
      primaryClass={cs.CL},
      url={https://arxiv.org/abs/2408.00118}, 
}

@misc{DeepSeek:et:al:2025a,
      title={DeepSeek-V3 Technical Report}, 
      author={DeepSeek-AI and Aixin Liu and Bei Feng and Bing Xue and Bingxuan Wang and Bochao Wu and Chengda Lu and Chenggang Zhao and Chengqi Deng and Chenyu Zhang and Chong Ruan and Damai Dai and Daya Guo and Dejian Yang and Deli Chen and Dongjie Ji and Erhang Li and Fangyun Lin and Fucong Dai and Fuli Luo and Guangbo Hao and Guanting Chen and Guowei Li and H. Zhang and Han Bao and Hanwei Xu and Haocheng Wang and Haowei Zhang and Honghui Ding and Huajian Xin and Huazuo Gao and Hui Li and Hui Qu and J. L. Cai and Jian Liang and Jianzhong Guo and Jiaqi Ni and Jiashi Li and Jiawei Wang and Jin Chen and Jingchang Chen and Jingyang Yuan and Junjie Qiu and Junlong Li and Junxiao Song and Kai Dong and Kai Hu and Kaige Gao and Kang Guan and Kexin Huang and Kuai Yu and Lean Wang and Lecong Zhang and Lei Xu and Leyi Xia and Liang Zhao and Litong Wang and Liyue Zhang and Meng Li and Miaojun Wang and Mingchuan Zhang and Minghua Zhang and Minghui Tang and Mingming Li and Ning Tian and Panpan Huang and Peiyi Wang and Peng Zhang and Qiancheng Wang and Qihao Zhu and Qinyu Chen and Qiushi Du and R. J. Chen and R. L. Jin and Ruiqi Ge and Ruisong Zhang and Ruizhe Pan and Runji Wang and Runxin Xu and Ruoyu Zhang and Ruyi Chen and S. S. Li and Shanghao Lu and Shangyan Zhou and Shanhuang Chen and Shaoqing Wu and Shengfeng Ye and Shengfeng Ye and Shirong Ma and Shiyu Wang and Shuang Zhou and Shuiping Yu and Shunfeng Zhou and Shuting Pan and T. Wang and Tao Yun and Tian Pei and Tianyu Sun and W. L. Xiao and Wangding Zeng and Wanjia Zhao and Wei An and Wen Liu and Wenfeng Liang and Wenjun Gao and Wenqin Yu and Wentao Zhang and X. Q. Li and Xiangyue Jin and Xianzu Wang and Xiao Bi and Xiaodong Liu and Xiaohan Wang and Xiaojin Shen and Xiaokang Chen and Xiaokang Zhang and Xiaosha Chen and Xiaotao Nie and Xiaowen Sun and Xiaoxiang Wang and Xin Cheng and Xin Liu and Xin Xie and Xingchao Liu and Xingkai Yu and Xinnan Song and Xinxia Shan and Xinyi Zhou and Xinyu Yang and Xinyuan Li and Xuecheng Su and Xuheng Lin and Y. K. Li and Y. Q. Wang and Y. X. Wei and Y. X. Zhu and Yang Zhang and Yanhong Xu and Yanhong Xu and Yanping Huang and Yao Li and Yao Zhao and Yaofeng Sun and Yaohui Li and Yaohui Wang and Yi Yu and Yi Zheng and Yichao Zhang and Yifan Shi and Yiliang Xiong and Ying He and Ying Tang and Yishi Piao and Yisong Wang and Yixuan Tan and Yiyang Ma and Yiyuan Liu and Yongqiang Guo and Yu Wu and Yuan Ou and Yuchen Zhu and Yuduan Wang and Yue Gong and Yuheng Zou and Yujia He and Yukun Zha and Yunfan Xiong and Yunxian Ma and Yuting Yan and Yuxiang Luo and Yuxiang You and Yuxuan Liu and Yuyang Zhou and Z. F. Wu and Z. Z. Ren and Zehui Ren and Zhangli Sha and Zhe Fu and Zhean Xu and Zhen Huang and Zhen Zhang and Zhenda Xie and Zhengyan Zhang and Zhewen Hao and Zhibin Gou and Zhicheng Ma and Zhigang Yan and Zhihong Shao and Zhipeng Xu and Zhiyu Wu and Zhongyu Zhang and Zhuoshu Li and Zihui Gu and Zijia Zhu and Zijun Liu and Zilin Li and Ziwei Xie and Ziyang Song and Ziyi Gao and Zizheng Pan},
      year={2025},
      eprint={2412.19437},
      archivePrefix={arXiv},
      primaryClass={cs.CL},
      url={https://arxiv.org/abs/2412.19437}, 
}

@Article{DeepSeek:et:al:2025b,
author={Guo, Daya
and Yang, Dejian
and Zhang, Haowei
and Song, Junxiao
and Wang, Peiyi
and Zhu, Qihao
and Xu, Runxin
and Zhang, Ruoyu
and Ma, Shirong
and Bi, Xiao
and Zhang, Xiaokang
and Yu, Xingkai
and Wu, Yu
and Wu, Z. F.
and Gou, Zhibin
and Shao, Zhihong
and Li, Zhuoshu
and Gao, Ziyi
and Liu, Aixin
and Xue, Bing
and Wang, Bingxuan
and Wu, Bochao
and Feng, Bei
and Lu, Chengda
and Zhao, Chenggang
and Deng, Chengqi
and Ruan, Chong
and Dai, Damai
and Chen, Deli
and Ji, Dongjie
and Li, Erhang
and Lin, Fangyun
and Dai, Fucong
and Luo, Fuli
and Hao, Guangbo
and Chen, Guanting
and Li, Guowei
and Zhang, H.
and Xu, Hanwei
and Ding, Honghui
and Gao, Huazuo
and Qu, Hui
and Li, Hui
and Guo, Jianzhong
and Li, Jiashi
and Chen, Jingchang
and Yuan, Jingyang
and Tu, Jinhao
and Qiu, Junjie
and Li, Junlong
and Cai, J. L.
and Ni, Jiaqi
and Liang, Jian
and Chen, Jin
and Dong, Kai
and Hu, Kai
and You, Kaichao
and Gao, Kaige
and Guan, Kang
and Huang, Kexin
and Yu, Kuai
and Wang, Lean
and Zhang, Lecong
and Zhao, Liang
and Wang, Litong
and Zhang, Liyue
and Xu, Lei
and Xia, Leyi
and Zhang, Mingchuan
and Zhang, Minghua
and Tang, Minghui
and Zhou, Mingxu
and Li, Meng
and Wang, Miaojun
and Li, Mingming
and Tian, Ning
and Huang, Panpan
and Zhang, Peng
and Wang, Qiancheng
and Chen, Qinyu
and Du, Qiushi
and Ge, Ruiqi
and Zhang, Ruisong
and Pan, Ruizhe
and Wang, Runji
and Chen, R. J.
and Jin, R. L.
and Chen, Ruyi
and Lu, Shanghao
and Zhou, Shangyan
and Chen, Shanhuang
and Ye, Shengfeng
and Wang, Shiyu
and Yu, Shuiping
and Zhou, Shunfeng
and Pan, Shuting
and Li, S. S.
and Zhou, Shuang
and Wu, Shaoqing
and Yun, Tao
and Pei, Tian
and Sun, Tianyu
and Wang, T.
and Zeng, Wangding
and Liu, Wen
and Liang, Wenfeng
and Gao, Wenjun
and Yu, Wenqin
and Zhang, Wentao
and Xiao, W. L.
and An, Wei
and Liu, Xiaodong
and Wang, Xiaohan
and Chen, Xiaokang
and Nie, Xiaotao
and Cheng, Xin
and Liu, Xin
and Xie, Xin
and Liu, Xingchao
and Yang, Xinyu
and Li, Xinyuan
and Su, Xuecheng
and Lin, Xuheng
and Li, X. Q.
and Jin, Xiangyue
and Shen, Xiaojin
and Chen, Xiaosha
and Sun, Xiaowen
and Wang, Xiaoxiang
and Song, Xinnan
and Zhou, Xinyi
and Wang, Xianzu
and Shan, Xinxia
and Li, Y. K.
and Wang, Y. Q.
and Wei, Y. X.
and Zhang, Yang
and Xu, Yanhong
and Li, Yao
and Zhao, Yao
and Sun, Yaofeng
and Wang, Yaohui
and Yu, Yi
and Zhang, Yichao
and Shi, Yifan
and Xiong, Yiliang
and He, Ying
and Piao, Yishi
and Wang, Yisong
and Tan, Yixuan
and Ma, Yiyang
and Liu, Yiyuan
and Guo, Yongqiang
and Ou, Yuan
and Wang, Yuduan
and Gong, Yue
and Zou, Yuheng
and He, Yujia
and Xiong, Yunfan
and Luo, Yuxiang
and You, Yuxiang
and Liu, Yuxuan
and Zhou, Yuyang
and Zhu, Y. X.
and Huang, Yanping
and Li, Yaohui
and Zheng, Yi
and Zhu, Yuchen
and Ma, Yunxian
and Tang, Ying
and Zha, Yukun
and Yan, Yuting
and Ren, Z. Z.
and Ren, Zehui
and Sha, Zhangli
and Fu, Zhe
and Xu, Zhean
and Xie, Zhenda
and Zhang, Zhengyan
and Hao, Zhewen
and Ma, Zhicheng
and Yan, Zhigang
and Wu, Zhiyu
and Gu, Zihui
and Zhu, Zijia
and Liu, Zijun
and Li, Zilin
and Xie, Ziwei
and Song, Ziyang
and Pan, Zizheng
and Huang, Zhen
and Xu, Zhipeng
and Zhang, Zhongyu
and Zhang, Zhen},
title={DeepSeek-R1 incentivizes reasoning in LLMs through reinforcement learning},
journal={Nature},
year={2025},
month={Sep},
day={01},
volume={645},
number={8081},
pages={633-638},
issn={1476-4687},
doi={10.1038/s41586-025-09422-z},
url={https://doi.org/10.1038/s41586-025-09422-z}
}

@misc{OpenAI:2026,
  author       = {{OpenAI}},
  title        = {Reasoning best practices},
  howpublished = {OpenAI API documentation},
  url          = {https://developers.openai.com/api/docs/guides/reasoning-best-practices},
  note         = {Accessed: 2026-02-26},
  year = {2026}
}

@ARTICLE{Molerov:et:al:2020,
    AUTHOR={Molerov, Dimitri  and Zlatkin-Troitschanskaia, Olga  and Nagel, Marie-Theres  and Brückner, Sebastian  and Schmidt, Susanne  and Shavelson, Richard J. },
    TITLE={Assessing University Students' Critical Online Reasoning Ability: A Conceptual and Assessment Framework With Preliminary Evidence},
    JOURNAL={Frontiers in Education},
    VOLUME={Volume 5 - 2020},
    YEAR={2020},
    URL={https://www.frontiersin.org/journals/education/articles/10.3389/feduc.2020.577843},
    DOI={10.3389/feduc.2020.577843},
    ISSN={2504-284X}
}

@inproceedings{Yang:et:al:2025b,
    title = "How Is {LLM} Reasoning Distracted by Irrelevant Context? An Analysis Using a Controlled Benchmark",
    author = "Yang, Minglai  and
      Huang, Ethan  and
      Zhang, Liang  and
      Surdeanu, Mihai  and
      Wang, William Yang  and
      Pan, Liangming",
    editor = "Christodoulopoulos, Christos  and
      Chakraborty, Tanmoy  and
      Rose, Carolyn  and
      Peng, Violet",
    booktitle = "Proceedings of the 2025 Conference on Empirical Methods in Natural Language Processing",
    month = nov,
    year = "2025",
    address = "Suzhou, China",
    publisher = "Association for Computational Linguistics",
    url = "https://aclanthology.org/2025.emnlp-main.674/",
    doi = "10.18653/v1/2025.emnlp-main.674",
    pages = "13329--13347",
    ISBN = "979-8-89176-332-6"
}

@inproceedings{Tavakoli:Zamani:2025,
author = {Tavakoli, Leila and Zamani, Hamed},
title = {Reliable Annotations with Less Effort: Evaluating LLM-Human Collaboration in Search Clarifications},
year = {2025},
isbn = {9798400718618},
publisher = {Association for Computing Machinery},
address = {New York, NY, USA},
url = {https://doi.org/10.1145/3731120.3744574},
doi = {10.1145/3731120.3744574},
booktitle = {Proceedings of the 2025 International ACM SIGIR Conference on Innovative Concepts and Theories in Information Retrieval (ICTIR)},
pages = {92–102},
numpages = {11},
location = {Padua, Italy},
series = {ICTIR '25}
}

@inproceedings{Guerdan:et:al:2025,
title={Validating {LLM}-as-a-Judge Systems under Rating Indeterminacy},
author={Luke Guerdan and Solon Barocas and Ken Holstein and Hanna Wallach and Steven Wu and Alexandra Chouldechova},
booktitle={The Thirty-ninth Annual Conference on Neural Information Processing Systems},
year={2025},
url={https://openreview.net/forum?id=ZwDMrArTBg}
}

@InProceedings{Deroy:et:al:2025,
  title = 	 {MIRROR: A Novel Approach for the Automated Evaluation of Open-Ended Question Generation},
  author =       {Deroy, Aniket and Maity, Subhankar and Sarkar, Sudeshna},
  booktitle = 	 {Proceedings of Large Foundation Models for Educational Assessment},
  pages = 	 {3--32},
  year = 	 {2025},
  editor = 	 {Li, Sheng and Cui, Zhongmin and Lu, Jiasen and Harris, Deborah and Jing, Shumin},
  volume = 	 {264},
  series = 	 {Proceedings of Machine Learning Research},
  month = 	 {15--16 Dec},
  publisher =    {PMLR},
  url = 	 {https://proceedings.mlr.press/v264/deroy25a.html}
}

@inproceedings{Zhou:et:al:2024,
    title = "Is {LLM} a Reliable Reviewer? A Comprehensive Evaluation of {LLM} on Automatic Paper Reviewing Tasks",
    author = "Zhou, Ruiyang  and
      Chen, Lu  and
      Yu, Kai",
    editor = "Calzolari, Nicoletta  and
      Kan, Min-Yen  and
      Hoste, Veronique  and
      Lenci, Alessandro  and
      Sakti, Sakriani  and
      Xue, Nianwen",
    booktitle = "Proceedings of the 2024 Joint International Conference on Computational Linguistics, Language Resources and Evaluation (LREC-COLING 2024)",
    month = may,
    year = "2024",
    address = "Torino, Italia",
    publisher = "ELRA and ICCL",
    url = "https://aclanthology.org/2024.lrec-main.816/",
    pages = "9340--9351"
}

@inproceedings{Lin:Chen:2023,
    title = "{LLM}-Eval: Unified Multi-Dimensional Automatic Evaluation for Open-Domain Conversations with Large Language Models",
    author = "Lin, Yen-Ting  and
      Chen, Yun-Nung",
    editor = "Chen, Yun-Nung  and
      Rastogi, Abhinav",
    booktitle = "Proceedings of the 5th Workshop on NLP for Conversational AI (NLP4ConvAI 2023)",
    month = jul,
    year = "2023",
    address = "Toronto, Canada",
    publisher = "Association for Computational Linguistics",
    url = "https://aclanthology.org/2023.nlp4convai-1.5/",
    doi = "10.18653/v1/2023.nlp4convai-1.5",
    pages = "47--58"
}

@misc{Li:et:al:2024,
      title={LLMs-as-Judges: A Comprehensive Survey on LLM-based Evaluation Methods}, 
      author={Haitao Li and Qian Dong and Junjie Chen and Huixue Su and Yujia Zhou and Qingyao Ai and Ziyi Ye and Yiqun Liu},
      year={2024},
      eprint={2412.05579},
      archivePrefix={arXiv},
      primaryClass={cs.CL},
      url={https://arxiv.org/abs/2412.05579}, 
}

@Article{Johnson:Zhang:2024,
author={Johnson, Matthew
and Zhang, Mo},
title={Examining the responsible use of zero-shot AI approaches to scoring essays},
journal={Scientific Reports},
year={2024},
month={Dec},
day={03},
volume={14},
number={1},
pages={30064},
issn={2045-2322},
doi={10.1038/s41598-024-79208-2},
url={https://doi.org/10.1038/s41598-024-79208-2}
}

@misc{Stasuik:2025,
series={Undergraduate Research},
title={Evaluating LLM Performance in Essay Assessment: A Comparative Analysis of AI Grading and Feedback Systems for University English Courses},
url={https://open.library.ubc.ca/collections/undergraduateresearch/52966/items/1.0448868},
DOI={http://dx.doi.org/10.14288/1.0448868},
author={Stasuik, Noah Carter},
year={2025},
month={Apr},
collection={Undergraduate Research}
}

@InProceedings{Yoshida:2025,
author="Yoshida, Lui",
editor="Cristea, Alexandra I.
and Walker, Erin
and Lu, Yu
and Santos, Olga C.
and Isotani, Seiji",
title="Do We Need a Detailed Rubric for Automated Essay Scoring Using Large Language Models?",
booktitle="Artificial Intelligence in Education",
year="2025",
publisher="Springer Nature Switzerland",
address="Cham",
pages="60--67",
isbn="978-3-031-98465-5"
}

@InProceedings{Frohn:et:al:2025,
author="Frohn, Scott
and Burleigh, Tyler
and Chen, Jing",
editor="Cristea, Alexandra I.
and Walker, Erin
and Lu, Yu
and Santos, Olga C.
and Isotani, Seiji",
title="Automated Scoring of Short Answer Questions with Large Language Models: Impacts of Model, Item, and Rubric Design",
booktitle="Artificial Intelligence in Education",
year="2025",
publisher="Springer Nature Switzerland",
address="Cham",
pages="44--51",
isbn="978-3-031-98465-5"
}

@misc{Borse:et:al:2025,
      title={Investigation of the Inter-Rater Reliability between Large Language Models and Human Raters in Qualitative Analysis}, 
      author={Nikhil Sanjay Borse and Ravishankar Chatta Subramaniam and N. Sanjay Rebello},
      year={2025},
      eprint={2508.14764},
      archivePrefix={arXiv},
      primaryClass={physics.ed-ph},
      url={https://arxiv.org/abs/2508.14764}, 
}

@article{Leckie:et:al:2011,
author = {Leckie, George and Baird, Jo-Anne},
title = {Rater Effects on Essay Scoring: A Multilevel Analysis of Severity Drift, Central Tendency, and Rater Experience},
journal = {Journal of Educational Measurement},
volume = {48},
number = {4},
pages = {399-418},
doi = {https://doi.org/10.1111/j.1745-3984.2011.00152.x},
url = {https://onlinelibrary.wiley.com/doi/abs/10.1111/j.1745-3984.2011.00152.x},
eprint = {https://onlinelibrary.wiley.com/doi/pdf/10.1111/j.1745-3984.2011.00152.x},
year = {2011}
}

@misc{Xu:et:al:2025,
      title={Tiny Model, Big Logic: Diversity-Driven Optimization Elicits Large-Model Reasoning Ability in VibeThinker-1.5B}, 
      author={Sen Xu and Yi Zhou and Wei Wang and Jixin Min and Zhibin Yin and Yingwei Dai and Shixi Liu and Lianyu Pang and Yirong Chen and Junlin Zhang},
      year={2025},
      eprint={2511.06221},
      archivePrefix={arXiv},
      primaryClass={cs.AI},
      url={https://arxiv.org/abs/2511.06221}, 
}

@inproceedings{Huang:et:al:2025,
    title = "Evaluating {LLM}-Based Automated Essay Scoring: Accuracy, Fairness, and Validity",
    author = "Huang, Yue  and
      Wilson, Joshua",
    editor = "Wilson, Joshua  and
      Ormerod, Christopher  and
      Beiting Parrish, Magdalen",
    booktitle = "Proceedings of the Artificial Intelligence in Measurement and Education Conference (AIME-Con): Works in Progress",
    month = oct,
    year = "2025",
    address = "Wyndham Grand Pittsburgh, Downtown, Pittsburgh, Pennsylvania, United States",
    publisher = "National Council on Measurement in Education (NCME)",
    url = "https://aclanthology.org/2025.aimecon-wip.9/",
    pages = "71--83",
    ISBN = "979-8-218-84229-1"
}

@inproceedings{Chamieh:et:al:2024,
    title = "{LLM}s in Short Answer Scoring: Limitations and Promise of Zero-Shot and Few-Shot Approaches",
    author = "Chamieh, Imran  and
      Zesch, Torsten  and
      Giebermann, Klaus",
    editor = {Kochmar, Ekaterina  and
      Bexte, Marie  and
      Burstein, Jill  and
      Horbach, Andrea  and
      Laarmann-Quante, Ronja  and
      Tack, Ana{\"i}s  and
      Yaneva, Victoria  and
      Yuan, Zheng},
    booktitle = "Proceedings of the 19th Workshop on Innovative Use of NLP for Building Educational Applications (BEA 2024)",
    month = jun,
    year = "2024",
    address = "Mexico City, Mexico",
    publisher = "Association for Computational Linguistics",
    url = "https://aclanthology.org/2024.bea-1.25/",
    pages = "309--315"
}

@article{Honnibal:et:al:2020,
author = {Honnibal, Matthew and Montani, Ines and Van Landeghem, Sofie and Boyd, Adriane},
doi = {10.5281/zenodo.1212303},
title = {{spaCy: Industrial-strength Natural Language Processing in Python}},
year = {2020}
}

@inproceedings{Barbieri:et:al:2022,
    title = "{XLM}-{T}: Multilingual Language Models in {T}witter for Sentiment Analysis and Beyond",
    author = "Barbieri, Francesco  and
      Espinosa Anke, Luis  and
      Camacho-Collados, Jose",
    booktitle = "Proceedings of the Thirteenth Language Resources and Evaluation Conference",
    month = jun,
    year = "2022",
    address = "Marseille, France",
    publisher = "European Language Resources Association",
    url = "https://aclanthology.org/2022.lrec-1.27",
    pages = "258--266"
}

@InProceedings{Guhr:et:al:2020,
  author    = {Guhr, Oliver  and  Schumann, Anne-Kathrin  and  Bahrmann, Frank  and  Böhme, Hans Joachim},
  title     = {Training a Broad-Coverage German Sentiment Classification Model for Dialog Systems},
  booktitle      = {Proceedings of The 12th Language Resources and Evaluation Conference},
  month          = {May},
  year           = {2020},
  address        = {Marseille, France},
  publisher      = {European Language Resources Association},
  pages     = {1620--1625},
  url       = {https://www.aclweb.org/anthology/2020.lrec-1.202}
}

@inproceedings{Tymann:et:al:2019,
  author    = {Tymann, Karsten Michael and Lutz, Matthias and Palsbr{\"o}ker, Patrick and Gips, Carsten},
  title     = {{GerVADER -- A German adaptation of the VADER sentiment analysis tool for social media texts}},
  booktitle = {Proceedings of the Conference ``Lernen, Wissen, Daten, Analysen'' (LWDA 2019)},
  series    = {CEUR Workshop Proceedings},
  volume    = {2454},
  year      = {2019},
  address   = {Berlin, Germany},
  month     = sep,
  url       = {http://ceur-ws.org/Vol-2454/paper_14.pdf},
  note      = {September 30--October 2, 2019. Available at: http://ceur-ws.org/Vol-2454/}
}

@article{Hutto:Gilbert:2014,
title={VADER: A Parsimonious Rule-Based Model for Sentiment Analysis of Social Media Text},
volume={8},
url={https://ojs.aaai.org/index.php/ICWSM/article/view/14550},
DOI={10.1609/icwsm.v8i1.14550},
number={1},
journal={Proceedings of the International AAAI Conference on Web and Social Media},
author={Hutto, C. and Gilbert, Eric},
year={2014},
month={May},
pages={216-225}
}

@inproceedings{Uslu:Mehler:Baumartz:2019,
  author    = {Uslu, Tolga and Mehler, Alexander and Baumartz, Daniel},
  booktitle = {{Proceedings of the 20th International Conference on Computational
               Linguistics and Intelligent Text Processing, (CICLing 2019)}},
  location  = {La Rochelle, France},
  series    = {{CICLing 2019}},
  title     = {{Computing Classifier-based Embeddings with the Help of text2ddc}},
  year      = {2019}
}

@inproceedings{Uslu:Mehler:Niekler:Baumartz:2018,
  author    = {Tolga Uslu and Alexander Mehler and Andreas Niekler and Daniel Baumartz},
  title     = {Towards a {DDC}-based Topic Network Model of Wikipedia},
  booktitle = {Proceedings of 2nd International Workshop on Modeling, Analysis,
               and Management of Social Networks and their Applications (SOCNET
               2018), February 28, 2018},
  year      = {2018}
}

@ARTICLE{Kuzman:Ljubesic:2025a,
  author={Kuzman, Taja and Ljubešić, Nikola},
  journal={IEEE Access}, 
  title={LLM Teacher-Student Framework for Text Classification With No Manually Annotated Data: A Case Study in IPTC News Topic Classification}, 
  year={2025},
  volume={},
  number={},
  pages={1-1},
  doi={10.1109/ACCESS.2025.3544814}}

@misc{Kuzman:Ljubesic:2025b,
    author       = { Kuzman, Taja and Ljube{\v s}i{\'c}, Nikola },
    title        = {{ Multilingual IPTC News Topic Classifier}},
    year         = 2025,
    url          = { https://huggingface.co/classla/multilingual-IPTC-news-topic-classifier },
    doi          = { 10.57967/hf/4709 },
    publisher    = { Hugging Face }
}

@software{Helin:2022,
  author       = {Helin, Ivan},
  title        = {{multilingual-sarcasm-detector}},
  year         = {2022},
  url          = {https://huggingface.co/helinivan/multilingual-sarcasm-detector},
  publisher    = {Hugging Face},
  note         = {Accessed: 2026-02-16},  
}

@article{Bertolini:et:al:2024,
title = {DReAMy: a library for the automatic analysis and annotation of dream reports with multilingual large language models},
journal = {Sleep Medicine},
volume = {115},
pages = {406-407},
year = {2024},
note = {Abstracts from the 17th World Sleep Congress},
issn = {1389-9457},
doi = {https://doi.org/10.1016/j.sleep.2023.11.1092},
url = {https://www.sciencedirect.com/science/article/pii/S1389945723015186},
author = {L. Bertolini and A. Michalak and J. Weeds}
}

@inproceedings{Bianchi:et:al:2022,
    title = "{XLM-EMO: Multilingual Emotion Prediction in Social Media Text}",
    author = "Bianchi, Federico and Nozza, Debora and Hovy, Dirk",
    booktitle = "Proceedings of the 12th Workshop on Computational Approaches to Subjectivity, Sentiment and Social Media Analysis",
    year = "2022",
    publisher = "Association for Computational Linguistics",
}

@software{AIstrova:2023,
  author       = {{AIstrova Technologies Inc.}},
  title        = {{One-for-all Toxicity v3}},
  year         = {2023},
  url          = {https://huggingface.co/FredZhang7/one-for-all-toxicity-v3},
  publisher    = {Hugging Face},
  note         = {Accessed: 2026-02-16},
  license      = {CC-BY-4.0}
}

@software{Stakovskii:2022,
  author       = {Stakovskii, Elisei},
  title        = {{xlm\_roberta\_base\_multilingual\_toxicity\_classifier\_plus}},
  year         = {2022},
  url          = {https://huggingface.co/EIStakovskii/xlm_roberta_base_multilingual_toxicity_classifier_plus},
  publisher    = {Hugging Face},
  note         = {Accessed: 2026-02-16},
}

@software{deepset:2020,
  author       = {{deepset}},
  title        = {{bert-base-german-cased-hatespeech-GermEval18Coarse}},
  year         = {2020},
  url          = {https://huggingface.co/deepset/bert-base-german-cased-hatespeech-GermEval18Coarse},
  publisher    = {Hugging Face},
  note         = {Accessed: 2026-02-16},
  license      = {CC-BY-4.0},
}

@misc{Aluru:et:al:2020,
      title={Deep Learning Models for Multilingual Hate Speech Detection}, 
      author={Sai Saketh Aluru and Binny Mathew and Punyajoy Saha and Animesh Mukherjee},
      year={2020},
      eprint={2004.06465},
      archivePrefix={arXiv},
      primaryClass={cs.SI},
      url={https://arxiv.org/abs/2004.06465}, 
}

@software{DiMascio:2020,
  author       = {DiMascio, Carmine},
  title        = {py-readability-metrics},
  year         = {2020},
  url          = {https://github.com/cdimascio/py-readability-metrics},
  version      = {1.4.5},
  publisher    = {GitHub},
  note         = {Accessed: 2026-02-16},
}

@inproceedings{Ye:et:al:2024,
    title = "Prompt Engineering a Prompt Engineer",
    author = "Ye, Qinyuan  and
      Ahmed, Mohamed  and
      Pryzant, Reid  and
      Khani, Fereshte",
    editor = "Ku, Lun-Wei  and
      Martins, Andre  and
      Srikumar, Vivek",
    booktitle = "Findings of the Association for Computational Linguistics: ACL 2024",
    month = aug,
    year = "2024",
    address = "Bangkok, Thailand",
    publisher = "Association for Computational Linguistics",
    url = "https://aclanthology.org/2024.findings-acl.21/",
    doi = "10.18653/v1/2024.findings-acl.21",
    pages = "355--385"
}

@misc{Du:2026,
      title={Tiered Super-Moore's Law: Price Evolution, Production Frontiers, and Market Competition in Large Language Model Inference Services}, 
      author={Mingdeng Du},
      year={2026},
      eprint={2603.28576},
      archivePrefix={arXiv},
      primaryClass={cs.CE},
      url={https://arxiv.org/abs/2603.28576}, 
}

@misc{Knoop:Holtmann:2026,
      title={Private LLM Inference on Consumer Blackwell GPUs: A Practical Guide for Cost-Effective Local Deployment in SMEs}, 
      author={Jonathan Knoop and Hendrik Holtmann},
      year={2026},
      eprint={2601.09527},
      archivePrefix={arXiv},
      primaryClass={cs.LG},
      url={https://arxiv.org/abs/2601.09527}, 
}

@article{Pedregosa:et:al:2011,
  title={Scikit-learn: Machine Learning in {P}ython},
  author={Pedregosa, F. and Varoquaux, G. and Gramfort, A. and Michel, V.
          and Thirion, B. and Grisel, O. and Blondel, M. and Prettenhofer, P.
          and Weiss, R. and Dubourg, V. and Vanderplas, J. and Passos, A. and
          Cournapeau, D. and Brucher, M. and Perrot, M. and Duchesnay, E.},
  journal={Journal of Machine Learning Research},
  volume={12},
  pages={2825--2830},
  year={2011}
}

@software{Chase:2022,
author = {Chase, Harrison},
month = oct,
title = {{LangChain}},
url = {https://github.com/langchain-ai/langchain},
year = {2022}
}

@software{Ollama:2023,
author = {Ollama},
month = jul,
title = {{Ollama}},
url = {https://github.com/ollama/ollama},
year = {2023}
}

@misc{Yao:et:al:2024,
      title={MiniCPM-V: A GPT-4V Level MLLM on Your Phone}, 
      author={Yuan Yao and Tianyu Yu and Ao Zhang and Chongyi Wang and Junbo Cui and Hongji Zhu and Tianchi Cai and Haoyu Li and Weilin Zhao and Zhihui He and Qianyu Chen and Huarong Zhou and Zhensheng Zou and Haoye Zhang and Shengding Hu and Zhi Zheng and Jie Zhou and Jie Cai and Xu Han and Guoyang Zeng and Dahai Li and Zhiyuan Liu and Maosong Sun},
      year={2024},
      eprint={2408.01800},
      archivePrefix={arXiv},
      primaryClass={cs.CV},
      url={https://arxiv.org/abs/2408.01800}, 
}

@misc{Wang:et:al:2024b,
      title={HelpSteer2-Preference: Complementing Ratings with Preferences}, 
      author={Zhilin Wang and Alexander Bukharin and Olivier Delalleau and Daniel Egert and Gerald Shen and Jiaqi Zeng and Oleksii Kuchaiev and Yi Dong},
      year={2024},
      eprint={2410.01257},
      archivePrefix={arXiv},
      primaryClass={cs.LG},
      url={https://arxiv.org/abs/2410.01257}, 
}

@article{Artstein:Poesio:2008,
    title = "Survey Article: Inter-Coder Agreement for Computational Linguistics",
    author = "Artstein, Ron  and
      Poesio, Massimo",
    journal = "Computational Linguistics",
    volume = "34",
    number = "4",
    year = "2008",
    url = "https://aclanthology.org/J08-4004/",
    doi = "10.1162/coli.07-034-R2",
    pages = "555--596"
}

@book{George:Mallery:2016,
  title     = {IBM SPSS Statistics 23 Step by Step: A Simple Guide and Reference},
  author    = {George, Darren and Mallery, Paul},
  edition   = {14},
  year      = {2016},
  publisher = {Routledge},
  address   = {New York}
}

@inproceedings{Shibata:Miyamura:2025,
    title = "{LCES}: Zero-shot Automated Essay Scoring via Pairwise Comparisons Using Large Language Models",
    author = "Shibata, Takumi  and
      Miyamura, Yuichi",
    editor = "Christodoulopoulos, Christos  and
      Chakraborty, Tanmoy  and
      Rose, Carolyn  and
      Peng, Violet",
    booktitle = "Proceedings of the 2025 Conference on Empirical Methods in Natural Language Processing",
    month = nov,
    year = "2025",
    address = "Suzhou, China",
    publisher = "Association for Computational Linguistics",
    url = "https://aclanthology.org/2025.emnlp-main.1523/",
    doi = "10.18653/v1/2025.emnlp-main.1523",
    pages = "29988--30001",
    ISBN = "979-8-89176-332-6"
}

@inproceedings{Burstein:et:al:1998,
    title = "Automated Scoring Using A Hybrid Feature Identification Technique",
    author = "Burstein, Jill  and
      Kukich, Karen  and
      Wolff, Susanne  and
      Lu, Chi  and
      Chodorow, Martin  and
      Braden-Harder, Lisa  and
      Harris, Mary Dee",
    booktitle = "{COLING} 1998 Volume 1: The 17th International Conference on Computational Linguistics",
    year = "1998",
    url = "https://aclanthology.org/C98-1032/"
}

@article{Rudner:Liang:2002,
title={Automated Essay Scoring Using Bayes’ Theorem},
volume={1},
url={https://ejournals.bc.edu/index.php/jtla/article/view/1668}, abstractNote={Two Bayesian models for text classification from the information science field were extended and applied to student produced essays. Both models were calibrated using 462 essays with two score points. The calibrated systems were applied to 80 new, pre-scored essays with 40 essays in each score group. Manipulated variables included the two models; the use of words, phrases and arguments; two approaches to trimming; stemming; and the use of stopwords. While the text classification literature suggests the need to calibrate on thousands of cases per score group, accuracy of over 80% was achieved with the sparse dataset used in this study.},
number={2},
journal={The Journal of Technology, Learning and Assessment},
author={Rudner, Lawrence M. and Liang, Tahung},
year={2002},
month={Jun.}
}

@inproceedings{Xie:et:al:2022,
    title = "Automated Essay Scoring via Pairwise Contrastive Regression",
    author = "Xie, Jiayi  and
      Cai, Kaiwei  and
      Kong, Li  and
      Zhou, Junsheng  and
      Qu, Weiguang",
    editor = "Calzolari, Nicoletta  and
      Huang, Chu-Ren  and
      Kim, Hansaem  and
      Pustejovsky, James  and
      Wanner, Leo  and
      Choi, Key-Sun  and
      Ryu, Pum-Mo  and
      Chen, Hsin-Hsi  and
      Donatelli, Lucia  and
      Ji, Heng  and
      Kurohashi, Sadao  and
      Paggio, Patrizia  and
      Xue, Nianwen  and
      Kim, Seokhwan  and
      Hahm, Younggyun  and
      He, Zhong  and
      Lee, Tony Kyungil  and
      Santus, Enrico  and
      Bond, Francis  and
      Na, Seung-Hoon",
    booktitle = "Proceedings of the 29th International Conference on Computational Linguistics",
    month = oct,
    year = "2022",
    address = "Gyeongju, Republic of Korea",
    publisher = "International Committee on Computational Linguistics",
    url = "https://aclanthology.org/2022.coling-1.240/",
    pages = "2724--2733"
}

@inproceedings{Dong:et:al:2017,
    title = "Attention-based Recurrent Convolutional Neural Network for Automatic Essay Scoring",
    author = "Dong, Fei  and
      Zhang, Yue  and
      Yang, Jie",
    editor = "Levy, Roger  and
      Specia, Lucia",
    booktitle = "Proceedings of the 21st Conference on Computational Natural Language Learning ({C}o{NLL} 2017)",
    month = aug,
    year = "2017",
    address = "Vancouver, Canada",
    publisher = "Association for Computational Linguistics",
    url = "https://aclanthology.org/K17-1017/",
    doi = "10.18653/v1/K17-1017",
    pages = "153--162"
}

@inproceedings{Yannakoudakis:et:al:2011,
    title = "A New Dataset and Method for Automatically Grading {ESOL} Texts",
    author = "Yannakoudakis, Helen  and
      Briscoe, Ted  and
      Medlock, Ben",
    editor = "Lin, Dekang  and
      Matsumoto, Yuji  and
      Mihalcea, Rada",
    booktitle = "Proceedings of the 49th Annual Meeting of the Association for Computational Linguistics: Human Language Technologies",
    month = jun,
    year = "2011",
    address = "Portland, Oregon, USA",
    publisher = "Association for Computational Linguistics",
    url = "https://aclanthology.org/P11-1019/",
    pages = "180--189"
}

@inproceedings{Calderon:et:al:2025,
    title = "The Alternative Annotator Test for {LLM}-as-a-Judge: How to Statistically Justify Replacing Human Annotators with {LLM}s",
    author = "Calderon, Nitay  and
      Reichart, Roi  and
      Dror, Rotem",
    editor = "Che, Wanxiang  and
      Nabende, Joyce  and
      Shutova, Ekaterina  and
      Pilehvar, Mohammad Taher",
    booktitle = "Proceedings of the 63rd Annual Meeting of the Association for Computational Linguistics (Volume 1: Long Papers)",
    month = jul,
    year = "2025",
    address = "Vienna, Austria",
    publisher = "Association for Computational Linguistics",
    url = "https://aclanthology.org/2025.acl-long.782/",
    doi = "10.18653/v1/2025.acl-long.782",
    pages = "16051--16081",
    ISBN = "979-8-89176-251-0"
}

@article{Vallat:2018,
    title = {Pingouin: statistics in Python},
    volume = {3},
    DOI = {10.21105/joss.01026},
    number = {31},
    journal = {Journal of Open Source Software},
    publisher = {The Open Journal},
    author = {Vallat,  Raphael},
    year = {2018},
    month = nov,
    pages = {1026}
}

@article{Tate:et:al:2024,
title = {Can AI provide useful holistic essay scoring?},
journal = {Computers and Education: Artificial Intelligence},
volume = {7},
pages = {100255},
year = {2024},
issn = {2666-920X},
doi = {https://doi.org/10.1016/j.caeai.2024.100255},
url = {https://www.sciencedirect.com/science/article/pii/S2666920X24000584},
author = {Tamara P. Tate and Jacob Steiss and Drew Bailey and Steve Graham and Youngsun Moon and Daniel Ritchie and Waverly Tseng and Mark Warschauer}
}

@article{Latif:Zhai:2024,
title = {Fine-tuning ChatGPT for automatic scoring},
journal = {Computers and Education: Artificial Intelligence},
volume = {6},
pages = {100210},
year = {2024},
issn = {2666-920X},
doi = {https://doi.org/10.1016/j.caeai.2024.100210},
url = {https://www.sciencedirect.com/science/article/pii/S2666920X24000110},
author = {Ehsan Latif and Xiaoming Zhai}
}

@Article{Cicchetti:Sparrow:1981,
author={Cicchetti, Domenic V.
and Sparrow, Sara A.},
title={Developing criteria for establishing interrater reliability of specific items: Applications to assessment of adaptive behavior.},
journal={American Journal of Mental Deficiency},
year={1981},
publisher={American Assn on Mental Retardation},
address={US},
volume={86},
number={2},
pages={127-137},
issn={0002-9351(Print)}
}

@misc{Krippendorff:2011,
  author = {Krippendorff, Klaus},
  title = {Computing Krippendorff's Alpha-Reliability},
  year = {2011},
  howpublished = {\url{https://repository.upenn.edu/handle/20.500.14332/2089}}
}

@article{Klie:et:al:2024,
    title = "Analyzing Dataset Annotation Quality Management in the Wild",
    author = "Klie, Jan-Christoph  and
      Eckart de Castilho, Richard  and
      Gurevych, Iryna",
    journal = "Computational Linguistics",
    volume = "50",
    number = "3",
    month = sep,
    year = "2024",
    address = "Cambridge, MA",
    publisher = "MIT Press",
    url = "https://aclanthology.org/2024.cl-3.1/",
    doi = "10.1162/coli_a_00516",
    pages = "817--866"
}

@Article{Shrout:Fleiss:1979,
author={Shrout, Patrick E.
and Fleiss, Joseph L.},
title={Intraclass correlations: Uses in assessing rater reliability.},
journal={Psychological Bulletin},
year={1979},
publisher={American Psychological Association},
address={US},
volume={86},
number={2},
pages={420-428},
doi={10.1037/0033-2909.86.2.420},
url={https://doi.org/10.1037/0033-2909.86.2.420}
}

@article{Cronbach:1951,
title={Coefficient Alpha and the Internal Structure of Tests},
volume={16},
DOI={10.1007/BF02310555},
number={3},
journal={Psychometrika},
author={Cronbach, Lee J.},
year={1951},
pages={297–334}
}

@article{Carletta:1996,
    title = "Assessing Agreement on Classification Tasks: The Kappa Statistic",
    author = "Carletta, Jean",
    editor = "Hirschberg, Julia",
    journal = "Computational Linguistics",
    volume = "22",
    number = "2",
    year = "1996",
    address = "Cambridge, MA",
    publisher = "MIT Press",
    url = "https://aclanthology.org/J96-2004/",
    pages = "249--254"
}

@misc{Castro:2017,
  author = {Santiago Castro},
  title = {Fast {K}rippendorff: Fast computation of {K}rippendorff's alpha agreement measure},
  year = {2017},
  publisher = {GitHub},
  journal = {GitHub repository},
  howpublished = {\url{https://github.com/pln-fing-udelar/fast-krippendorff}}
}

@misc{Zhao:et:al:2025,
      title={A Survey of Large Language Models}, 
      author={Wayne Xin Zhao and Kun Zhou and Junyi Li and Tianyi Tang and Xiaolei Wang and Yupeng Hou and Yingqian Min and Beichen Zhang and Junjie Zhang and Zican Dong and Yifan Du and Chen Yang and Yushuo Chen and Zhipeng Chen and Jinhao Jiang and Ruiyang Ren and Yifan Li and Xinyu Tang and Zikang Liu and Peiyu Liu and Jian-Yun Nie and Ji-Rong Wen},
      year={2025},
      eprint={2303.18223},
      archivePrefix={arXiv},
      primaryClass={cs.CL},
      url={https://arxiv.org/abs/2303.18223}, 
}

@Article{Guo:et:al:2025,
author={Guo, Daya
and Yang, Dejian
and Zhang, Haowei
and Song, Junxiao
and Wang, Peiyi
and Zhu, Qihao
and Xu, Runxin
and Zhang, Ruoyu
and Ma, Shirong
and Bi, Xiao
and Zhang, Xiaokang
and Yu, Xingkai
and Wu, Yu
and Wu, Z. F.
and Gou, Zhibin
and Shao, Zhihong
and Li, Zhuoshu
and Gao, Ziyi
and Liu, Aixin
and Xue, Bing
and Wang, Bingxuan
and Wu, Bochao
and Feng, Bei
and Lu, Chengda
and Zhao, Chenggang
and Deng, Chengqi
and Ruan, Chong
and Dai, Damai
and Chen, Deli
and Ji, Dongjie
and Li, Erhang
and Lin, Fangyun
and Dai, Fucong
and Luo, Fuli
and Hao, Guangbo
and Chen, Guanting
and Li, Guowei
and Zhang, H.
and Xu, Hanwei
and Ding, Honghui
and Gao, Huazuo
and Qu, Hui
and Li, Hui
and Guo, Jianzhong
and Li, Jiashi
and Chen, Jingchang
and Yuan, Jingyang
and Tu, Jinhao
and Qiu, Junjie
and Li, Junlong
and Cai, J. L.
and Ni, Jiaqi
and Liang, Jian
and Chen, Jin
and Dong, Kai
and Hu, Kai
and You, Kaichao
and Gao, Kaige
and Guan, Kang
and Huang, Kexin
and Yu, Kuai
and Wang, Lean
and Zhang, Lecong
and Zhao, Liang
and Wang, Litong
and Zhang, Liyue
and Xu, Lei
and Xia, Leyi
and Zhang, Mingchuan
and Zhang, Minghua
and Tang, Minghui
and Zhou, Mingxu
and Li, Meng
and Wang, Miaojun
and Li, Mingming
and Tian, Ning
and Huang, Panpan
and Zhang, Peng
and Wang, Qiancheng
and Chen, Qinyu
and Du, Qiushi
and Ge, Ruiqi
and Zhang, Ruisong
and Pan, Ruizhe
and Wang, Runji
and Chen, R. J.
and Jin, R. L.
and Chen, Ruyi
and Lu, Shanghao
and Zhou, Shangyan
and Chen, Shanhuang
and Ye, Shengfeng
and Wang, Shiyu
and Yu, Shuiping
and Zhou, Shunfeng
and Pan, Shuting
and Li, S. S.
and Zhou, Shuang
and Wu, Shaoqing
and Yun, Tao
and Pei, Tian
and Sun, Tianyu
and Wang, T.
and Zeng, Wangding
and Liu, Wen
and Liang, Wenfeng
and Gao, Wenjun
and Yu, Wenqin
and Zhang, Wentao
and Xiao, W. L.
and An, Wei
and Liu, Xiaodong
and Wang, Xiaohan
and Chen, Xiaokang
and Nie, Xiaotao
and Cheng, Xin
and Liu, Xin
and Xie, Xin
and Liu, Xingchao
and Yang, Xinyu
and Li, Xinyuan
and Su, Xuecheng
and Lin, Xuheng
and Li, X. Q.
and Jin, Xiangyue
and Shen, Xiaojin
and Chen, Xiaosha
and Sun, Xiaowen
and Wang, Xiaoxiang
and Song, Xinnan
and Zhou, Xinyi
and Wang, Xianzu
and Shan, Xinxia
and Li, Y. K.
and Wang, Y. Q.
and Wei, Y. X.
and Zhang, Yang
and Xu, Yanhong
and Li, Yao
and Zhao, Yao
and Sun, Yaofeng
and Wang, Yaohui
and Yu, Yi
and Zhang, Yichao
and Shi, Yifan
and Xiong, Yiliang
and He, Ying
and Piao, Yishi
and Wang, Yisong
and Tan, Yixuan
and Ma, Yiyang
and Liu, Yiyuan
and Guo, Yongqiang
and Ou, Yuan
and Wang, Yuduan
and Gong, Yue
and Zou, Yuheng
and He, Yujia
and Xiong, Yunfan
and Luo, Yuxiang
and You, Yuxiang
and Liu, Yuxuan
and Zhou, Yuyang
and Zhu, Y. X.
and Huang, Yanping
and Li, Yaohui
and Zheng, Yi
and Zhu, Yuchen
and Ma, Yunxian
and Tang, Ying
and Zha, Yukun
and Yan, Yuting
and Ren, Z. Z.
and Ren, Zehui
and Sha, Zhangli
and Fu, Zhe
and Xu, Zhean
and Xie, Zhenda
and Zhang, Zhengyan
and Hao, Zhewen
and Ma, Zhicheng
and Yan, Zhigang
and Wu, Zhiyu
and Gu, Zihui
and Zhu, Zijia
and Liu, Zijun
and Li, Zilin
and Xie, Ziwei
and Song, Ziyang
and Pan, Zizheng
and Huang, Zhen
and Xu, Zhipeng
and Zhang, Zhongyu
and Zhang, Zhen},
title={DeepSeek-R1 incentivizes reasoning in LLMs through reinforcement learning},
journal={Nature},
year={2025},
month={Sep},
day={01},
volume={645},
number={8081},
pages={633-638},
issn={1476-4687},
doi={10.1038/s41586-025-09422-z},
url={https://doi.org/10.1038/s41586-025-09422-z}
}

@misc{Zhang:et:al:2025,
      title={AFlow: Automating Agentic Workflow Generation}, 
      author={Jiayi Zhang and Jinyu Xiang and Zhaoyang Yu and Fengwei Teng and Xionghui Chen and Jiaqi Chen and Mingchen Zhuge and Xin Cheng and Sirui Hong and Jinlin Wang and Bingnan Zheng and Bang Liu and Yuyu Luo and Chenglin Wu},
      year={2025},
      eprint={2410.10762},
      archivePrefix={arXiv},
      primaryClass={cs.AI},
      url={https://arxiv.org/abs/2410.10762}, 
}

@inproceedings{Xiao:et:al:2024,
    title = "{F}low{B}ench: Revisiting and Benchmarking Workflow-Guided Planning for {LLM}-based Agents",
    author = "Xiao, Ruixuan  and
      Ma, Wentao  and
      Wang, Ke  and
      Wu, Yuchuan  and
      Zhao, Junbo  and
      Wang, Haobo  and
      Huang, Fei  and
      Li, Yongbin",
    editor = "Al-Onaizan, Yaser  and
      Bansal, Mohit  and
      Chen, Yun-Nung",
    booktitle = "Findings of the Association for Computational Linguistics: EMNLP 2024",
    month = nov,
    year = "2024",
    address = "Miami, Florida, USA",
    publisher = "Association for Computational Linguistics",
    url = "https://aclanthology.org/2024.findings-emnlp.638/",
    doi = "10.18653/v1/2024.findings-emnlp.638",
    pages = "10883--10900"
}

@article{Yin:et:al:2024,
   title={A survey on multimodal large language models},
   volume={11},
   ISSN={2053-714X},
   url={http://dx.doi.org/10.1093/nsr/nwae403},
   DOI={10.1093/nsr/nwae403},
   number={12},
   journal={National Science Review},
   publisher={Oxford University Press (OUP)},
   author={Yin, Shukang and Fu, Chaoyou and Zhao, Sirui and Li, Ke and Sun, Xing and Xu, Tong and Chen, Enhong},
   year={2024},
   month=Nov }

@inproceedings{Caffagni:et:al:2024,
    title = "The Revolution of Multimodal Large Language Models: A Survey",
    author = "Caffagni, Davide  and
      Cocchi, Federico  and
      Barsellotti, Luca  and
      Moratelli, Nicholas  and
      Sarto, Sara  and
      Baraldi, Lorenzo  and
      Baraldi, Lorenzo  and
      Cornia, Marcella  and
      Cucchiara, Rita",
    editor = "Ku, Lun-Wei  and
      Martins, Andre  and
      Srikumar, Vivek",
    booktitle = "Findings of the Association for Computational Linguistics: ACL 2024",
    month = aug,
    year = "2024",
    address = "Bangkok, Thailand",
    publisher = "Association for Computational Linguistics",
    url = "https://aclanthology.org/2024.findings-acl.807/",
    doi = "10.18653/v1/2024.findings-acl.807",
    pages = "13590--13618"
}

@inproceedings{Li:2025,
    title = "A Review of Prominent Paradigms for {LLM}-Based Agents: Tool Use, Planning (Including {RAG}), and Feedback Learning",
    author = "Li, Xinzhe",
    editor = "Rambow, Owen  and
      Wanner, Leo  and
      Apidianaki, Marianna  and
      Al-Khalifa, Hend  and
      Eugenio, Barbara Di  and
      Schockaert, Steven",
    booktitle = "Proceedings of the 31st International Conference on Computational Linguistics",
    month = jan,
    year = "2025",
    address = "Abu Dhabi, UAE",
    publisher = "Association for Computational Linguistics",
    url = "https://aclanthology.org/2025.coling-main.652/",
    pages = "9760--9779"
}

@inproceedings{Su:et:al:2025,
    title = "{E}ssay{J}udge: A Multi-Granular Benchmark for Assessing Automated Essay Scoring Capabilities of Multimodal Large Language Models",
    author = "Su, Jiamin  and
      Yan, Yibo  and
      Fu, Fangteng  and
      Han, Zhang  and
      Ye, Jingheng  and
      Liu, Xiang  and
      Huo, Jiahao  and
      Zhou, Huiyu  and
      Hu, Xuming",
    editor = "Che, Wanxiang  and
      Nabende, Joyce  and
      Shutova, Ekaterina  and
      Pilehvar, Mohammad Taher",
    booktitle = "Findings of the Association for Computational Linguistics: ACL 2025",
    month = jul,
    year = "2025",
    address = "Vienna, Austria",
    publisher = "Association for Computational Linguistics",
    url = "https://aclanthology.org/2025.findings-acl.329/",
    doi = "10.18653/v1/2025.findings-acl.329",
    pages = "6363--6389",
    ISBN = "979-8-89176-256-5"
}

@inproceedings{He:et:al:2025,
author = {He, Xinyun and Shu, Qi and Zhang, Mo and Huang, Wei and Zhao, Han and Zhu, Mengxiao},
title = {Beyond Final Products: Multi-Dimensional Essay Scoring Using Keystroke Logs and Deep Learning},
year = {2025},
isbn = {9798400707018},
publisher = {Association for Computing Machinery},
address = {New York, NY, USA},
url = {https://doi.org/10.1145/3706468.3706548},
doi = {10.1145/3706468.3706548},
booktitle = {Proceedings of the 15th International Learning Analytics and Knowledge Conference},
pages = {601–610},
numpages = {10},
location = {
},
series = {LAK '25}
}

@inproceedings{Wei:et:al:2022,
 author = {Wei, Jason and Wang, Xuezhi and Schuurmans, Dale and Bosma, Maarten and ichter, brian and Xia, Fei and Chi, Ed and Le, Quoc V and Zhou, Denny},
 booktitle = {Advances in Neural Information Processing Systems},
 editor = {S. Koyejo and S. Mohamed and A. Agarwal and D. Belgrave and K. Cho and A. Oh},
 pages = {24824--24837},
 publisher = {Curran Associates, Inc.},
 title = {Chain-of-Thought Prompting Elicits Reasoning in Large Language Models},
 url = {https://proceedings.neurips.cc/paper_files/paper/2022/file/9d5609613524ecf4f15af0f7b31abca4-Paper-Conference.pdf},
 volume = {35},
 year = {2022}
}

@inproceedings{Doewes:et:al:2023,
  author       = {Afrizal Doewes and
                  Nughthoh Arfawi Kurdhi and
                  Akrati Saxena},
  title        = {Evaluating Quadratic Weighted Kappa as the
                   Standard Performance Metric for Automated Essay
                   Scoring
                  },
  booktitle    = {Proceedings of the 16th International Conference
                   on Educational Data Mining
                  },
  year         = 2023,
  pages        = {103--113},
  publisher    = {International Educational Data Mining Society},
  month        = jul,
  venue        = {Bengaluru, India},
  doi          = {10.5281/zenodo.8115784},
  url          = {https://doi.org/10.5281/zenodo.8115784},
}

@Article{DeRaadt:et:al:2021,
author={de Raadt, Alexandra
and Warrens, Matthijs J.
and Bosker, Roel J.
and Kiers, Henk A. L.},
title={A Comparison of Reliability Coefficients for Ordinal Rating Scales},
journal={Journal of Classification},
year={2021},
month={Oct},
day={01},
volume={38},
number={3},
pages={519-543},
issn={1432-1343},
doi={10.1007/s00357-021-09386-5},
url={https://doi.org/10.1007/s00357-021-09386-5}
}

@inproceedings{Albatarni:et:al:2026,
  title = {Quadratic Weighted Kappa Is Not Enough for Evaluating Automated Essay Scoring Models},
  author = {Albatarni, Salam and Elsayed, Tamer},
  booktitle = {Proceedings of the Fifteenth Language Resources and Evaluation Conference (LREC 2026)},
  month = {May},
  year = {2026},
  pages = {4447--4456},
  address = {Palma, Mallorca, Spain},
  publisher = {European Language Resources Association (ELRA)},
  editor = {Piperidis, Stelios and Bel, Núria and van den Heuvel, Henk and Ide, Nancy and Krek, Simon and Toral, Antonio},
  doi = {10.63317/3co8wwdqqyf6}
}

@inproceedings{Brown:et:al:2020,
 author = {Brown, Tom and Mann, Benjamin and Ryder, Nick and Subbiah, Melanie and Kaplan, Jared D and Dhariwal, Prafulla and Neelakantan, Arvind and Shyam, Pranav and Sastry, Girish and Askell, Amanda and Agarwal, Sandhini and Herbert-Voss, Ariel and Krueger, Gretchen and Henighan, Tom and Child, Rewon and Ramesh, Aditya and Ziegler, Daniel and Wu, Jeffrey and Winter, Clemens and Hesse, Chris and Chen, Mark and Sigler, Eric and Litwin, Mateusz and Gray, Scott and Chess, Benjamin and Clark, Jack and Berner, Christopher and McCandlish, Sam and Radford, Alec and Sutskever, Ilya and Amodei, Dario},
 booktitle = {Advances in Neural Information Processing Systems},
 editor = {H. Larochelle and M. Ranzato and R. Hadsell and M.F. Balcan and H. Lin},
 pages = {1877--1901},
 publisher = {Curran Associates, Inc.},
 title = {Language Models are Few-Shot Learners},
 url = {https://proceedings.neurips.cc/paper_files/paper/2020/file/1457c0d6bfcb4967418bfb8ac142f64a-Paper.pdf},
 volume = {33},
 year = {2020}
}

@article{Lundberg:et:al:2020,
  title={From local explanations to global understanding with explainable AI for trees},
  author={Lundberg, Scott M. and Erion, Gabriel and Chen, Hugh and DeGrave, Alex and Prutkin, Jordan M. and Nair, Bala and Katz, Ronit and Himmelfarb, Jonathan and Bansal, Nisha and Lee, Su-In},
  journal={Nature Machine Intelligence},
  volume={2},
  number={1},
  pages={2522-5839},
  year={2020},
  publisher={Nature Publishing Group}
}

@inproceedings{Lundberg:Lee:2017,
 author = {Lundberg, Scott M and Lee, Su-In},
 booktitle = {Advances in Neural Information Processing Systems},
 editor = {I. Guyon and U. Von Luxburg and S. Bengio and H. Wallach and R. Fergus and S. Vishwanathan and R. Garnett},
 pages = {},
 publisher = {Curran Associates, Inc.},
 title = {A Unified Approach to Interpreting Model Predictions},
 url = {https://proceedings.neurips.cc/paper_files/paper/2017/file/8a20a8621978632d76c43dfd28b67767-Paper.pdf},
 volume = {30},
 year = {2017}
}

@misc{OpenAI:2026b,
  author       = {{OpenAI}},
  title        = {GPT-4.5},
  year         = {2025},
  month        = feb,
  day          = {17},
  howpublished = {\url{https://openai.com/index/gpt-4-5-system-card/}}
}

@misc{Anthropic:2026,
  author       = {{Anthropic}},
  title        = {Claude Opus 4.7},
  year         = {2026},
  month        = apr,
  day          = {16},
  howpublished = {\url{https://anthropic.com/claude-opus-4-7-system-card}}
}

@inproceedings{Amorim:et:al:2018,
    title = "Automated Essay Scoring in the Presence of Biased Ratings",
    author = "Amorim, Evelin  and
      Can{\c{c}}ado, Marcia  and
      Veloso, Adriano",
    editor = "Walker, Marilyn  and
      Ji, Heng  and
      Stent, Amanda",
    booktitle = "Proceedings of the 2018 Conference of the North {A}merican Chapter of the Association for Computational Linguistics: Human Language Technologies, Volume 1 (Long Papers)",
    month = jun,
    year = "2018",
    address = "New Orleans, Louisiana",
    publisher = "Association for Computational Linguistics",
    url = "https://aclanthology.org/N18-1021/",
    doi = "10.18653/v1/N18-1021",
    pages = "229--237"
}

@misc{Weidinger:et:al:2021,
      title={Ethical and social risks of harm from Language Models}, 
      author={Laura Weidinger and John Mellor and Maribeth Rauh and Conor Griffin and Jonathan Uesato and Po-Sen Huang and Myra Cheng and Mia Glaese and Borja Balle and Atoosa Kasirzadeh and Zac Kenton and Sasha Brown and Will Hawkins and Tom Stepleton and Courtney Biles and Abeba Birhane and Julia Haas and Laura Rimell and Lisa Anne Hendricks and William Isaac and Sean Legassick and Geoffrey Irving and Iason Gabriel},
      year={2021},
      eprint={2112.04359},
      archivePrefix={arXiv},
      primaryClass={cs.CL},
      url={https://arxiv.org/abs/2112.04359}, 
}

@inproceedings{Ouyang:et:al:2022,
 author = {Ouyang, Long and Wu, Jeffrey and Jiang, Xu and Almeida, Diogo and Wainwright, Carroll and Mishkin, Pamela and Zhang, Chong and Agarwal, Sandhini and Slama, Katarina and Ray, Alex and Schulman, John and Hilton, Jacob and Kelton, Fraser and Miller, Luke and Simens, Maddie and Askell, Amanda and Welinder, Peter and Christiano, Paul F and Leike, Jan and Lowe, Ryan},
 booktitle = {Advances in Neural Information Processing Systems},
 editor = {S. Koyejo and S. Mohamed and A. Agarwal and D. Belgrave and K. Cho and A. Oh},
 pages = {27730--27744},
 publisher = {Curran Associates, Inc.},
 title = {Training language models to follow instructions with human feedback},
 url = {https://proceedings.neurips.cc/paper_files/paper/2022/file/b1efde53be364a73914f58805a001731-Paper-Conference.pdf},
 volume = {35},
 year = {2022}
}

@Inbook{Link:Koltovskaia:2023,
author="Link, Stephanie
and Koltovskaia, Svetlana",
editor="Kruse, Otto
and Rapp, Christian
and Anson, Chris M.
and Benetos, Kalliopi
and Cotos, Elena
and Devitt, Ann
and Shibani, Antonette",
title="Automated Scoring of Writing",
bookTitle="Digital Writing Technologies in Higher Education : Theory, Research, and Practice",
year="2023",
publisher="Springer International Publishing",
address="Cham",
pages="333--345",
isbn="978-3-031-36033-6",
doi="10.1007/978-3-031-36033-6_21",
url="https://doi.org/10.1007/978-3-031-36033-6_21"
}

@Inbook{Cotos:2023,
author="Cotos, Elena",
editor="Kruse, Otto
and Rapp, Christian
and Anson, Chris M.
and Benetos, Kalliopi
and Cotos, Elena
and Devitt, Ann
and Shibani, Antonette",
title="Automated Feedback on Writing",
bookTitle="Digital Writing Technologies in Higher Education : Theory, Research, and Practice",
year="2023",
publisher="Springer International Publishing",
address="Cham",
pages="347--364",
isbn="978-3-031-36033-6",
doi="10.1007/978-3-031-36033-6_22",
url="https://doi.org/10.1007/978-3-031-36033-6_22"
}

@article{Liu:2024,
  author    = {Liu, Wen},
  title     = {A Systematic Review of Automated Writing Evaluation Feedback: Validity, Effects and Students' Engagement},
  journal   = {Language Teaching Research Quarterly},
  year      = {2024},
  volume    = {45},
  pages     = {86--105},
  doi       = {10.32038/ltrq.2024.45.05},
  url       = {https://www.eurokd.com/doi/10.32038/ltrq.2024.45.05},
  issn      = {2667-6753},
  publisher = {European Knowledge Development Institute}
}

@article{Yang:et:al:2024,
  title={Qwen2 technical report},
  author={Yang, An and Yang, Baosong and Hui, Binyuan and Zheng, Bo and Yu, Bowen and Zhou, Chang and Li, Chengpeng and Li, Chengyuan and Liu, Dayiheng and Huang, Fei and others},
  journal={arXiv preprint arXiv:2407.10671},
  year={2024}
}

@misc{Royce:2024,
  author = {Petr Royce},
  title = {Plutus-Meta-Llama-3.1-8B-Instruct-bnb-4bit},
  year = {2024},
  publisher = {Hugging Face},
  howpublished = {https://huggingface.co/0xroyce/Plutus-Meta-Llama-3.1-8B-Instruct-bnb-4bit},
}

@misc{Sellergren:et:al:2026,
      title={MedGemma 1.5 Technical Report}, 
      author={Andrew Sellergren and Chufan Gao and Fereshteh Mahvar and Timo Kohlberger and Fayaz Jamil and Madeleine Traverse and Alberto Tono and Bashir Sadjad and Lin Yang and Charles Lau and Liron Yatziv and Tiffany Chen and Bram Sterling and Kenneth Philbrick and Richa Tiwari and Yun Liu and Madhuram Jajoo and Chandrashekar Sankarapu and Swapnil Vispute and Harshad Purandare and Abhishek Bijay Mishra and Sam Schmidgall and Tao Tu and Anil Palepu and Chunjong Park and Tim Strother and Rahul Thapa and Yong Cheng and Preeti Singh and Kat Black and Yossi Matias and Katherine Chou and Avinatan Hassidim and Kavi Goel and Joelle Barral and Tris Warkentin and Shravya Shetty and Dale Webster and Sunny Virmani and David F. Steiner and Can Kirmizibayrak and Daniel Golden},
      year={2026},
      eprint={2604.05081},
      archivePrefix={arXiv},
      primaryClass={cs.AI},
      url={https://arxiv.org/abs/2604.05081}, 
}






\end{document}